\documentclass{article}

\usepackage{ifthen}
\newboolean{icml_template}
\setboolean{icml_template}{true} 
\newcommand{\icml}[2]{\ifthenelse{\boolean{icml_template}}{#1}{#2}}

\icml{

\usepackage{microtype}
\usepackage{graphicx}
\usepackage{subfigure}
\usepackage{booktabs} 

\usepackage{multirow}

\usepackage{hyperref}

\usepackage[accepted]{icml2025}

\usepackage{amsmath}
\usepackage{amssymb}
\usepackage{mathtools}
\usepackage{amsthm}

\usepackage[capitalize,noabbrev]{cleveref}

\newtheorem{thm}{Theorem}[]
\newtheorem*{thm*}{Theorem}
\newtheorem{m-thm}[thm]{Meta-Theorem}
\newtheorem*{m-thm*}{Meta-Theorem}
\newtheorem{lem}{Lemma}[]
\newtheorem{remark}{Remark}[]
\newenvironment{rem}{\begin{remark}\rm}{\end{remark}}
\newtheorem{prop}{Proposition}[]
\newtheorem*{prop*}{Proposition}
\newtheorem{Definition}{Definition}
\newenvironment{dfn}{\begin{Definition}\rm}{\end{Definition}}
\newtheorem{Corollary}[thm]{Corollary}
\newenvironment{cor}{\begin{Corollary}}{\end{Corollary}}
\newtheorem{Example}[thm]{Example}
\newenvironment{eg}{\begin{Example}\rm}{\end{Example}}
\newtheorem{algor}[thm]{Method}
\newenvironment{algo}{\begin{algor}\rm}{\end{algor}}
\newtheorem{Condition}[thm]{Condition}
\newenvironment{cond}{\begin{Condition}}{\end{Condition}}
\newtheorem{met}[thm]{Method}
\newtheorem{assp}{Assumption}[]

\newtheorem{asspp}{Assumption}[]

\renewcommand{\phi}{\varphi}

\newcommand{\X}{\mathcal{X}}

\newcommand{\E}[2][]{\mathbb{E}_{#1}\!\left[#2\right]}

\newcommand{\ord}[1]{\mathcal{O}\!\left(#1\right)}

\newcommand{\R}{\mathbb{R}}

\newcommand{\ve}{\epsilon}

\renewcommand{\S}{\mathcal{S}}

\newcommand{\sm}[1]{{\setminus #1}}

\usepackage{bm,bbm}
\usepackage{nicematrix}
\usepackage{comment}
\renewcommand{\L}{\mathcal{L}}
\newcommand{\spm}[1]{{\scriptsize$\pm$#1}}
\newcommand{\mask}{\text{\ttfamily[MASK]}}

\usepackage[textsize=tiny]{todonotes}
\usepackage{listings}

\icmltitlerunning{Distillation of
    Discrete Diffusion through
    Dimensional Correlations}

}{

\usepackage{PRIMEarxiv}
\usepackage[round,authoryear]{natbib}
\usepackage{color,xcolor}
\usepackage{tikz}
\definecolor{c_yuhta}{rgb}{0.831,0.184,0.494}
\definecolor{c_gyuhta}{rgb}{0.7,0.7,0.7}

\usepackage[utf8]{inputenc} 
\usepackage[T1]{fontenc}    
\usepackage{hyperref}       
\usepackage{url}            
\usepackage{booktabs}       
\usepackage{amsfonts}       
\usepackage{nicefrac}       
\usepackage{microtype}      
\usepackage{xcolor}         

\usepackage{amsmath,amssymb,amsthm}
\usepackage{bm,bbm}
\usepackage{booktabs}
\usepackage{nicematrix}
\usepackage{comment}
\renewcommand{\L}{\mathcal{L}}
\newcommand{\spm}[1]{{\scriptsize$\pm$#1}}
\newcommand{\mask}{\text{\ttfamily[MASK]}}
\usepackage{graphicx}

\usepackage[textsize=tiny]{todonotes}
\usepackage{listings}
\usepackage{subfigure,cleveref,wrapfig,float}

\title{Distillation of
Discrete Diffusion through\\
Dimensional Correlations}

\newtheorem{thm}{Theorem}[]
\newtheorem*{thm*}{Theorem}
\newtheorem{m-thm}[thm]{Meta-Theorem}
\newtheorem*{m-thm*}{Meta-Theorem}
\newtheorem{lem}{Lemma}[]
\newtheorem{remark}{Remark}[]

\newtheorem{prop}{Proposition}[]
\newtheorem*{prop*}{Proposition}
\newtheorem{Definition}{Definition}

\newtheorem{Corollary}[thm]{Corollary}

\newtheorem{Example}[thm]{Example}

\newtheorem{algor}[thm]{Method}

\newtheorem{Condition}[thm]{Condition}

\newtheorem{assp}{Assumption}[]

\renewcommand{\phi}{\varphi}

\newcommand{\X}{\mathcal{X}}

\newcommand{\E}[2][]{\mathbb{E}_{#1}\!\left[#2\right]}

\newcommand{\ord}[1]{\mathcal{O}\!\left(#1\right)}

\newcommand{\R}{\mathbb{R}}

\newcommand{\ve}{\epsilon}

\renewcommand{\S}{\mathcal{S}}

\newcommand{\sm}[1]{{\setminus #1}}

\author{%
  Satoshi Hayakawa$^\dagger$$^*$
  \quad
  Yuhta Takida$^\ddagger$
  \quad
  Masaaki Imaizumi$^\mathsection$
  \quad
  Hiromi Wakaki$^\dagger$
  \quad
  Yuki Mitsufuji$^\dagger$$^\ddagger$
  \\
  $^\dagger$Sony Group Corporation
  \quad
  $^\ddagger$Sony AI
  \quad
  $^\mathsection$The University of Tokyo
  \\$^*$\texttt{satoshi.a.hayakawa@sony.com}
}
}

\begin{document}

\icml{

\twocolumn[
\icmltitle{Distillation of
    Discrete Diffusion through
    Dimensional Correlations}




\begin{icmlauthorlist}
\icmlauthor{Satoshi Hayakawa}{x}
\icmlauthor{Yuhta Takida}{y}
\icmlauthor{Masaaki Imaizumi}{z}
\icmlauthor{Hiromi Wakaki}{x}
\icmlauthor{Yuki Mitsufuji}{x,y}
\end{icmlauthorlist}

\icmlaffiliation{x}{Sony Group Corporation, Tokyo, Japan}
\icmlaffiliation{y}{Sony AI, Tokyo, Japan}
\icmlaffiliation{z}{The University of Tokyo, Tokyo, Japan}

\icmlcorrespondingauthor{Satoshi Hayakawa}{satoshi.a.hayakawa@sony.com}

\icmlkeywords{Machine Learning, ICML}

\vskip 0.3in
]



\printAffiliationsAndNotice{}  
}{
    \maketitle

    \vspace{-8mm}
}

\begin{abstract}
Diffusion models have demonstrated exceptional performances in various fields of generative modeling,
but suffer from slow sampling speed due to their iterative nature.
While this issue is being addressed in continuous domains, discrete diffusion models face unique challenges, particularly in capturing dependencies between elements (e.g., pixel relationships in image, sequential dependencies in language)
mainly due to the computational cost of processing high-dimensional joint distributions.
In this paper, (i) we propose ``mixture'' models for discrete diffusion
that are capable of treating dimensional correlations while remaining scalable,
and (ii) we provide a set of loss functions for distilling the iterations of existing models.
Two primary theoretical insights underpin our approach:
First, conventional models with element-wise independence can well
approximate the data distribution,
but essentially require {\it many sampling steps}.
Second, our loss functions enable the mixture models to distill
such many-step conventional models into just a few steps by learning the dimensional correlations.
Our experimental results show the effectiveness of the proposed method in distilling pretrained discrete diffusion models
across image and language domains.
The code used in the paper is available at \url{https://github.com/sony/di4c}.
\end{abstract}

\section{Introduction}

Diffusion models~\citep{sohl2015deep,ho2020denoising,songscore} have demonstrated excellent performance in generative modeling, particularly for continuous data such as images~\citep{nichol2021glide,rombach2022high,saharia2022photorealistic}, audio~\citep{kong2020diffwave,chen2021wavegrad,evans2024fast}, and video~\citep{harvey2022flexible,ho2022video,blattmann2023stable}. Recent advancements in diffusion models often outperform traditional generative models, such as variational autoencoders~\citep[VAEs,][]{kingma2013auto,higgins2017beta,zhao2019infovae} and generative adversarial networks~\citep[GANs,][]{goodfellow2014generative}, in terms of sample quality and the controllability of the generated results. 
Furthermore, diffusion models are not limited to learning continuous data; they can also be applied to discrete or categorical data with modifications~\citep{hoogeboom2021argmax,austin2021structured}
and offer a promising approach for discrete generative modeling~\citep{gu2022vector,loudiscrete}.
Such discrete diffusion models
are the main topic of this paper.

A notable drawback of diffusion models is their slow sampling speed due to
requiring many sampling steps~\citep{xiao2021tackling,zhang2022fast}.
In continuous domains, various approaches have been proposed to reduce the number of steps,
including well-designed forward processes~\citep{songdenoising}
and fast solvers of stochastic/ordinary differential equations \citep[SDEs/ODEs,][]{lu2022dpm,lu2023dpm,zheng2023dpm}.
Another notable approach is knowledge distillation, which significantly reduces the number of sampling steps compared with earlier attempts by compressing pretrained diffusion models into single- or few-step generative models~\citep{luhman2021knowledge,salimans2022progressive,meng2023distillation,zheng2023fast}.
An emerging sub-family of distillation is the consistency-type models~\citep{song2023consistency,song2023improved,kimconsistency}, which exploit the fact that
samples generated via different paths from the same initial noise should coincide.

\begin{figure*}
    \centering
    \includegraphics[width=0.9\linewidth]{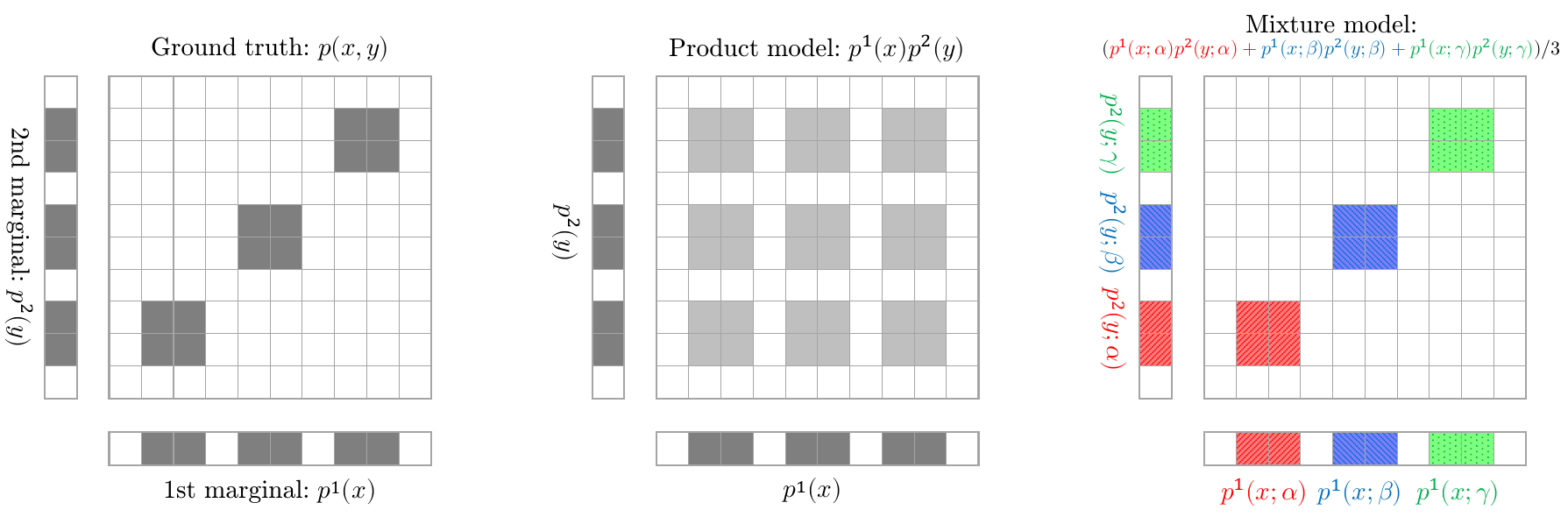}
    \vspace{-2mm}
    \caption{Illustration of dimensional correlations. ({\it Left}) Distribution $p(x, y)$ is
        two-dimensional categorical distribution. $p^1(x)$ and $p^2(y)$ are its marginals.
        ({\it Center}) Conventional denoiser in discrete diffusion uses {\it product model},
        which is simply product of marginal distributions.
        It fails to approximate ground truth distribution.
        ({\it Right}) Our {\it mixture model} is given by
        expectation of product model $p(x,y;\lambda)=p^1(x;\lambda)p^2(y;\lambda)$ for random $\lambda$.
        In figure, $\lambda$ takes $\alpha, \beta, \gamma$ in equal probabilities,
        and model reconstructs $p(x, y)$.
    }
    \label{fig:dim-correlation}
\end{figure*}

However, discrete diffusion models face a fundamental challenge when attempting to reduce the number of sampling steps.
Conventional approaches use ``product'' models
that treat each dimension independently
as sampling distributions (Figure~\ref{fig:dim-correlation}, center),
since high-dimensional joint distributions are intractable.
While this has been successful with hundreds of sampling steps,
ignoring element-wise dependencies (which we refer to as {\it dimensional correlations})
causes non-negligible approximation errors (Figure~\ref{fig:dim-correlation}).
This ignoring is also pointed out
in some concurrent works~\citep{jys,liu2024discrete,xu2024energy}.

In this paper, we propose \textbf{Di4C} (Distilling Discrete Diffusion through Dimensional Correlations)
to overcome this limitation.
Our key insight is that while individual steps in conventional models are dimensionally independent,
their composition over multiple steps can implicitly capture correlations
(Section~\ref{sec:compose-motivation}).
On the basis of this observation, we develop
(1) a ``mixture'' model that explicitly represents dimensional correlations while remaining computationally tractable
(Figure~\ref{fig:dim-correlation}, right),
and (2) novel loss functions that effectively distill the many-step denoising of a product model into fewer steps.
Our contribution is as follows:

    \textbf{Theoretical analysis:}
        In Theorem~\ref{thm:composition},
        we show that $N$-step sampling with product models
        can approximate data distributions
        in $\ord{1/N}$ total variation error.
        We also prove that this bound cannot be improved in a simple two-dimensional example.
        It underpins the empirical effectiveness of discrete diffusion models
        {\it with many steps} and, at the same time,
        shows the importance of modeling dimensional correlations to reduce the number of sampling steps.

    \textbf{Model and loss design:}
    To capture the aforementioned dimensional correlations,
    we propose
    a ``mixture'' model
    that can represent dimensional correlations
    (Section~\ref{sec:mixture}).
    To distill a many-step discrete diffusion model (teacher) into a few-step model (student),
    we also propose Di4C loss functions
    for compressing the iterative process of the teacher (Section~\ref{sec:losses}).
    In theory, we prove that the loss functions in Di4C can upper-bound the distance
    between the output distributions
    of the $N$-step teacher and the student with just one step (Theorem~\ref{thm:main}).
    In combination with Theorem~\ref{thm:composition},
    this provides an overall theoretical guarantee for Di4C.

    \textbf{Experiments:}
    Finally, we demonstrate that our approach is general and applicable
    to multiple settings.
    (1) On CIFAR-10 with a pixel-based discretized Gaussian diffusion,
    we substantially improve the sample quality metrics
    of the teacher model \citep{campbell2022continuous} in few-step sampling.
    (2) On ImageNet class-conditional generation with masked diffusion,
    our method achieves a 2x speed-up while maintaining a comparable sample quality to
    the teacher model \citep{besnier2023pytorch}.
    In addition, (3)~on masked diffusion language modeling with OpenWebText,
    we show that Di4C can further distill a well-distilled model \citep{deschenaux2024beyond}
    by capturing dimensional correlations, without much harming of sampling diversity.
    These results consistently demonstrate that Di4C can effectively compress the sampling steps of discrete diffusion models while maintaining or improving generation quality.

Finally, the remainder of this paper is organized as follows:
Section~\ref{sec:prelims} gives preliminaries on discrete diffusion models
and explains the dimensionality issue in discrete diffusion.
We then explain the central idea of Di4C in Section~\ref{sec:di4c}
and show theoretical results in Section~\ref{sec:theory},
which are partially described above as our contribution.
In Section~\ref{sec:experiments}, we also provide experimental results with image and language tasks.
After discussing related works in Section~\ref{sec:related},
we conclude the paper with discussions on its limitations and future work
in Section~\ref{sec:conclusion}.


\section{Preliminaries}
\label{sec:prelims}


\subsection{Discrete diffusion models}
Suppose we have a data distribution $q_0:=q_\mathrm{data}$
over the space $\X$.
In diffusion models \citep{sohl2015deep,ho2020denoising},
we consider a Markov process $(\bm{x}_t)_{0\le t\le T}$
with $\bm{x}_0\sim q_0$ and $\bm{x}_T\sim q_T$,
where the time $t$ can be either discrete or continuous.
In this paper, we follow the notational convention that $q_{t|s}$ and $q_{s,t}$ represent the true conditional and joint
distributions defined by this Markov process, respectively.
This process is designed so that
the terminal distribution
$q_T$ is a tractable distribution.
Following the convention,
we regard this forward process as adding noise to the data distribution.
Our aim is to generate samples approximately from the conditional distribution
$q_{0|T}(\cdot|\bm{x}_T)$
with $\bm{x}_T\sim q_T$,
which is a generative model for $q_\mathrm{data}$.
To this end, we introduce a {\it model} or {\it denoiser}, which is represented as $p_{s|t}$ (for $s<t$), to approximate $q_{s|t}$. 


Our primary interest is in the discrete diffusion models \citep{austin2021structured,campbell2022continuous},
where the space $\X$ is a finite set.
In this case, a probability distribution $p$
on $\X$ can be regarded as a function $p:\X\to\R$,
and we will sometimes abuse the notation by treating $p$ as just an ordinary function.
We are given a finite set $\S$ and consider a diffusion process over the product space $\X=\S^D$
for a large $D$.
Each state $\bm{x}\in\X$ can thus be written as
$\bm{x} = (x^d)_{d=1}^D$,
where $x^d$ indicates the entry of $\bm{x}$
in the $d$-th dimension.
Given a probability distribution $p = p(\bm{x})$ on $\X$,
let $p^d = p^d(x^d)$ be its $d$-th marginal distribution,
i.e., the distribution of $x^d$ given $\bm{x}\sim p$.
To enjoy scalability, the forward process is usually set to be factorized over dimensions,
i.e.,
$q_{t|s}(\bm{x}_t|\bm{x}_s)
= \prod_{d=1}^D q_{t|s}^d(x_t^d|x_s^d)$
holds for $s<t$
\citep{gu2022vector,campbell2022continuous}.

\subsection{Dimensional correlations in discrete diffusion}
\label{sec:detail-dim-corr}
The common practices in modeling and training discrete diffusion models lead them to ignore the dimensional correlations within a data distribution.
First, under the aforementioned problem setting,
for the sake of scalability,
the denoiser is usually defined as a {\it product model} that satisfies
\begin{equation}
    p_{s|t}(\bm{x}_s|\bm{x}_t) = \prod_{d=1}^D
    p^d_{s|t}(x_s^d|\bm{x}_t),
    \qquad s<t.
    \label{eq:product-model}
\end{equation}
Namely, the distribution $p_{s|t}(\cdot|\bm{x}_t)$ is dimensionally independent.
This product modeling
is common if not particularly highlighted \citep[Section~G]{campbell2022continuous}, due to the combinatorial explosion of the product discrete state.
Indeed, adopting a product model significantly reduces the output length
from $\mathcal{O}(D^{\lvert\S\rvert})$ to
$\ord{D\lvert\S\rvert}$
at the cost of representational capacity.
This limited expressive power can be crucial for considering few-step discrete diffusion models. As an extreme example, consider doing one-step denoising in the case of masked (absorbing-state) diffusion~\citep{austin2021structured};
there is no chance we can approximate a complex joint distribution (as in Figure~\ref{fig:dim-correlation}) in one step
when $\bm{x}_T$ is a completely masked sentence (i.e., following a delta distribution) and $p_{0|T}(\cdot|\bm{x}_T)$ is dimensionally independent.
See~Section~\ref{sec:sampling} for more examples.

Another potential factor making the learning of dimensional correlations infeasible
in discrete diffusion models is that some of the existing loss functions are not well prepared
for learning dimensional correlations.
Most notably, in the continuous-time score-based discrete diffusion,
we need only the marginal $p^d_{s|t}(\cdot|\bm{x}_t)$ or its variant
to compute the infinitesimal transition rate
(see, e.g., \citet[Proposition~3]{campbell2022continuous} or \citet[Eq.~16]{sunscore}).
Thus, learning backward transition rates does not lead to capturing dimensional correlations,
even if we use a model capable of representing them.

\section{Di4C for distilling discrete diffusion models}
\label{sec:di4c}

This section describes our proposed method,
Di4C. We first show that the {\it composition} of well-trained discrete diffusion models can represent the dimensional correlation
in Section~\ref{sec:compose-motivation},
and in the later sections we discuss how to distill the multi-step denoising of a teacher model
into a student model that requires fewer steps.
In particular, Section~\ref{sec:mixture} and Section~\ref{sec:losses}
try to solve the existing limitation described in Section~\ref{sec:detail-dim-corr}
in terms of modeling and loss design, respectively.
See Section~\ref{sec:training-di4c}
for more technical details of Di4C.

\subsection{Composition of diffusion denoisers for inducing dimensional correlations}
\label{sec:compose-motivation}
We introduce the notion of composition,
which plays a significant role in representing the dimensional correlations to be learned.
Consider two general conditional distributions
$p(\bm{x}|\bm{y})$ and $\tilde{p}(\bm{y}|\bm{z})$
over finite sets.
We define their composition as
\[
    p\circ \tilde{p}(\bm{x}|\bm{z}):=
    \E[\bm{y}\sim \tilde{p}(\cdot|\bm{z})]{p(\bm{x}|\bm{y})}
    = \sum_{\bm{y}}p(\bm{x}|\bm{y})\tilde{p}(\bm{y}|\bm{z}),
\]
where this definition can be extended to the continuous case in a straightforward way. Although this is just a convolution of two functions, it can be viewed as a composition of denoising operators in the context of diffusion models.
Specifically, given a single-step denoiser $p_{s|t}$
and the finite timesteps $0=t_0<t_1<\cdots<t_N=T$,
we typically use
$p_{t_0|t_1}\circ \cdots \circ p_{t_{N-1}|t_N}
(\cdot|\bm{x}_T)$
with the terminal noise $\bm{x}_T\sim q_T$
as a generative sampler.

Notably, the composition can serve as a source of dimensional correlation in discrete diffusion models.
Even if one-step denoisers,
$p_{s|u}$ and $p_{u|t}$ ($s<u<t$),
are dimensionally independent,
their composition is generally not.
Indeed, dimensionally independent denoisers
are successful given hundreds of sampling steps~\citep{austin2021structured,gu2022vector}.
Our method aims at compressing the composition of well-trained denoisers into few-step sampling by learning dimensional correlations.

Let $p^\psi$ be a pretrained teacher model with a product structure
and $p^\psi_{0|t_1}\circ \cdots\circ p^\psi_{t_{N-1}|t_N}$
be a sufficiently good approximation of $q_{0|T}$,
where $0<t_1<\cdots<t_N=T$ are timesteps.
In our distillation,
we would like the student model $p^\theta$
to approximate the teacher compositions as
\begin{equation}
    p^\theta_{0|t_n} \approx p^\psi_{0|t_1}\circ\cdots\circ p^\psi_{t_{n-1}|t_n},
    \qquad
    n=1,\ldots,N.
    \label{eq:student-teacher-target}
\end{equation}
To achieve this,
we provide a way of modeling $p^\theta$ that is capable of representing dimensional correlations
in Section~\ref{sec:mixture},
and we propose a set of loss functions to distill dimensional correlation represented by the compositions of a teacher model in Section~\ref{sec:losses}.

\subsection{Mixture models to capture dimensional correlations}
\label{sec:mixture}

As an effective instance to represent correlated multivariate categorical distributions, we propose a {\it mixture model}. We define it as a family of conditional probability distributions
that have the following representation for $s<t$:
\icml{
    \begin{align}
        &p_{s|t}^\theta(\bm{x}_s|\bm{x}_t)
        = \E[\lambda]{p_{s|t}^{\theta}(\bm{x}_s|\bm{x}_t; \lambda)},
        \label{eq:mixture-model}
        \\
        &\text{where}\quad p_{s|t}^{\theta}(\bm{x}_s|\bm{x}_t;\lambda)
        = \prod_{d=1}^D p_{s|t}^{\theta,d}(x_s^d|\bm{x}_t;\lambda).
        \nonumber
    \end{align}
}{
    \begin{equation}
        p_{s|t}^\theta(\bm{x}_s|\bm{x}_t)
        = \E[\lambda]{p_{s|t}^{\theta}(\bm{x}_s|\bm{x}_t; \lambda)},
        \qquad\text{where}\ \  p_{s|t}^{\theta}(\bm{x}_s|\bm{x}_t;\lambda)
        = \prod_{d=1}^D p_{s|t}^{\theta,d}(x_s^d|\bm{x}_t;\lambda).
        \label{eq:mixture-model}
    \end{equation}
}
Here, $\lambda$ is a random variable with an arbitrary distribution.
This distribution can be viewed as a convex mixture of product models indexed by $\lambda$.
See Figure~\ref{fig:dim-correlation} ({\it right}) for an intuitive illustration.
Despite the fact that $p_{0|t}^{\theta}(\bm{x}_0|\bm{x}_t;\lambda)$
is dimensionally independent for each given point $\lambda$, this mixture representation is universal in the following sense:
\begin{prop}
    For any probability distribution $p$ over $\S^D$,
    there exist a probability distribution
    $\pi$
    and a family of product distributions $p(\bm{x};\lambda) = \prod_{d=1}^Dp^d(x^d;\lambda)$ indexed by $\lambda$
    satisfying
    $p(\bm{x}) = \E[\lambda\sim\pi]{p(\bm{x};\lambda)}$
    for all $\bm{x}\in\S^D$.
\end{prop}
Indeed, we have
$p(\bm{x}) = \E[\bm{z}\sim p]{\delta_{\bm{z}}(\bm{x})}$,
where $\delta_{\bm{z}}$ is the delta distribution at $\bm{z}$,
which is certainly a product distribution.
Although the proof is not very informative,
the assertion itself implies that the mixture model has sufficient expressive power to capture dimensional correlations. 
It should also be noted that sampling from this mixture model during the inference has almost no extra computational overhead compared with the conventional product model,
since it just requires
sampling and insertion of $\lambda$
(see Section~\ref{sec:latency}). 

\subsection{Consistency for distilling dimensional correlations}\label{sec:losses}
\icml{}{
    \begin{figure}
        \centering
        \includegraphics[width=0.6\linewidth]{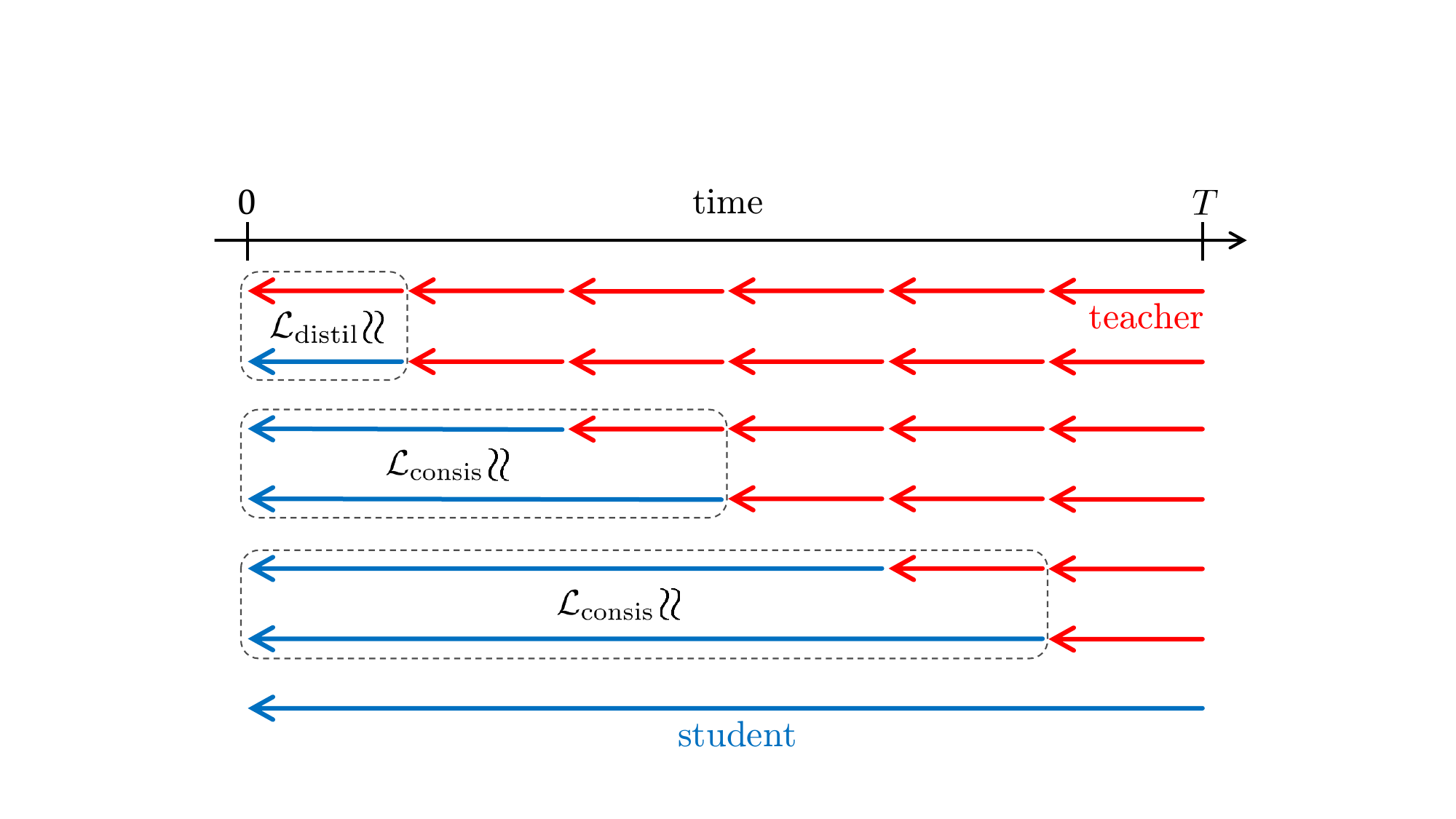}
        \caption{An illustration of how our loss functions work.
        Through $\L_\mathrm{distil}$ and $\L_\mathrm{consis}$,
        we distill the multiple teacher denoising steps into
        fewer steps of the student denoiser.}
        \label{fig:loss-illustration}
    \end{figure}
}
We present a set of (two) loss functions that take dimensional correlation into account. Consider we are given a product teacher model, which is denoted as $p^\psi$. Let $p^\theta$ be a general student model (with enough expressive power; an example is given in Section~\ref{sec:mixture}) that we want to train based on $p^\psi$.

\paragraph{Distillation loss.}
We first introduce a {\it distillation loss},
which forces the student model to approximate the teacher model at time $\delta$ $(\ll T)$:
\begin{align*}
    \mathcal{L}_\mathrm{distil}(\theta;\psi,r_\delta,\delta)
    &:=
    \E[\bm{x}_{\delta}\sim r_\delta]{
        D_{\mathrm{KL}}(
        p_{0|\delta}^{\psi}(\cdot|\bm{x}_\delta)
        \,\Vert\,
        p_{0|\delta}^{\theta}(\cdot|\bm{x}_{\delta})
        )},
\end{align*}
where 
$r_\delta$ ($\approx q_\delta$)
is a reference distribution over $\X$ at time $\delta$
and $D_\mathrm{KL}$ is the Kullback--Leibler~(KL) divergence.
We expect that a single teacher denoising step is enough for a small $\delta$; the dimensional correlation is mainly incorporated
in the following consistency loss (see also Section~\ref{sec:distil-loss}).

\paragraph{Consistency loss.}
We then propose a {\it consistency loss},
which allows the student model
to learn the dimensional correlation represented by the teacher denoiser compositions:
\icml{
    \begin{align*}
        &\mathcal{L}_\mathrm{consis}(\theta;\psi,r_t,s,u,t)\\
        &:= \E[\bm{x}_t\sim r_t]{D_\mathrm{KL}(
            p_{s|u}^\theta\circ p_{u|t}^\psi(\cdot|\bm{x}_t)
            \,\Vert\,
            p_{s|t}^\theta(\cdot|\bm{x}_t)
            )},
    \end{align*}
}{
    \[
        \mathcal{L}_\mathrm{consis}(\theta;\psi,r_t,s,u,t)
        := \E[\bm{x}_t\sim r_t]{D_\mathrm{KL}(
            p_{s|u}^\theta\circ p_{u|t}^\psi(\cdot|\bm{x}_t)
            \,\Vert\,
            p_{s|t}^\theta(\cdot|\bm{x}_t)
            )},
    \]
}
where $r_t$ is a reference distribution over
$\X$ at time $t$ approximating $q_t$.
While this loss is not straightforward to compute, we discuss how to approximate it in practice with Monte Carlo or control variates in Section~\ref{sec:consis-loss}.
Note that the idea of mixing the teacher denoiser and student denoiser in $\L_\mathrm{consis}$ can also be found in the continuous-state setting
regarding ODE trajectories \citep[Fig.~3]{kimconsistency}, but our loss is different in that
we work on the compositions of conditional probabilities as in \eqref{eq:student-teacher-target}.

Figure~\ref{fig:loss-illustration} shows the intuition behind our loss functions.
As reference distributions $r_\delta$ and $r_t$,
we can either use $q_t$ generated from data
or the distribution 
obtained by applying multiple teacher denoising steps.
See Section~\ref{sec:theory} for their roles and theoretical guarantees on $\L_\mathrm{distil}$ and $\L_\mathrm{consis}$.

\icml{
    \begin{figure}
        \centering
        \includegraphics[width=0.95\linewidth]{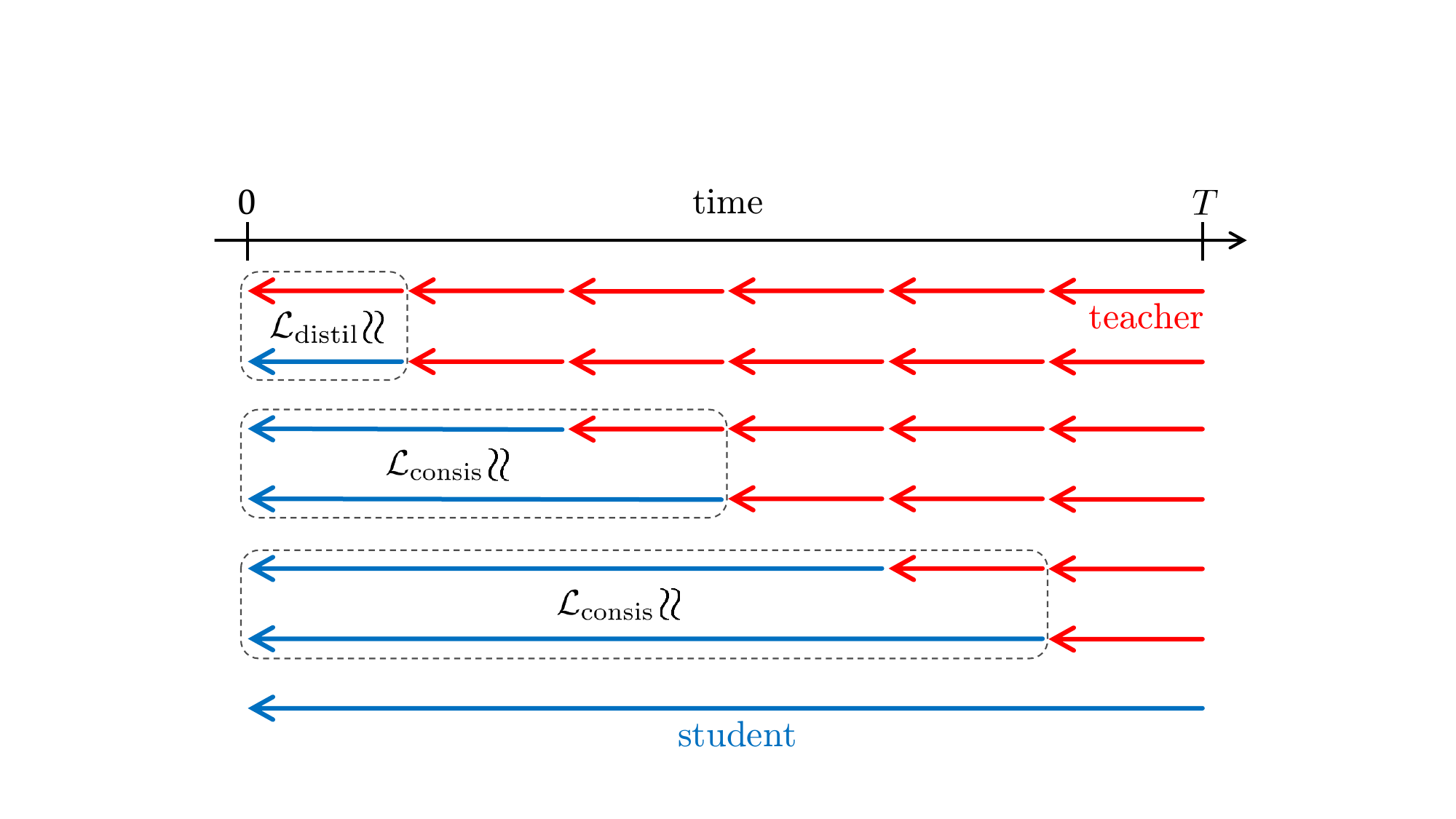}
        \vspace{-3mm}
        \caption{Illustration of how our loss functions work.
        Through $\L_\mathrm{distil}$ and $\L_\mathrm{consis}$,
        we distill multiple teacher denoising steps into
        fewer steps of student denoiser.}
        \label{fig:loss-illustration}
    \end{figure}
}{
}

\section{Theoretical analysis}\label{sec:theory}
In this section, we present an overall theoretical analysis on
our distillation method. 
In Section~\ref{sec:compose-theory}, we show that the conventional product model \eqref{eq:product-model} can approximate a data distribution if the model's marginal is perfectly trained and {\it given many steps} ($\ord{1/N}$ upper bound).
We also show that its $N$-step total variation error can be lower bounded by $\Omega(1/N)$ even for a simple two-dimensional example.
Both bounds support the empirical evidences of existing models that work (only) under many steps.
In Section~\ref{sec:distil-theory}, we prove that the proposed objective functions enable the many-step denoising with a teacher model to be distilled into a few-step student model, provided that the student model has enough expressive power.

By combining the upper bounds in these results,
we informally obtain the following estimate
when distilling an $N$-step sampling process of a teacher model that learns the marginals perfectly:
\icml{
    \begin{align*}
        &d_\mathrm{TV}(p(\text{$1$-step student}), q_0)
        \tag{\raisebox{1pt}{\tiny$\swarrow$} Theorem~\ref{thm:composition}, upper bound}\\
        &\le d_\mathrm{TV}(p(\text{$1$-step student}), p(\text{$N$-step teacher})) + \ord{1/N}\\
        &\le (\text{Di4C losses in \eqref{eq:main-ineq}}) + \ord{1/N}.
        \tag{Theorem~\ref{thm:main}}
    \end{align*}
}{
    \begin{align*}
        d_\mathrm{TV}(p(\text{$1$-step student}), q_0)
        &\le d_\mathrm{TV}(p(\text{$1$-step student}), p(\text{$N$-step teacher})) + \ord{1/N}
        \tag{Theorem~\ref{thm:composition}, upper bound}\\
        &\le (\text{Di4C losses in \eqref{eq:main-ineq}}) + \ord{1/N}.
        \tag{Theorem~\ref{thm:main}}
    \end{align*}
}
Here, $p(\text{$n$-step model})$ represents the resulting distribution of
$n$-step sampling with the model starting from $q_T$,
and $d_\mathrm{TV}$ denotes the total variation distance.

\subsection{
Product models with multi-step sampling
can approximate data distribution}
\label{sec:compose-theory}
We first show that dimensionally independent denoisers with many steps are capable of approximately recovering a data distribution,
which has already been empirically observed in existing studies.
To consider varying the number of denoising steps,
let us work on the continuous-time setting.
Let $(\bm{x}_t)_{0\le t\le T}$ follow a continuous-time Markov chain
over $[0, T]$ and the space $\X=\S^D$ with a factorized
forward process,
i.e., $q_{t|s}(\bm{x}_t|\bm{x}_s)
=\prod_{d=1}^Dq_{t|s}^d(x_t^d|x_s^d)$
for $s<t$.
See Section~\ref{sec:ctmc} for more details.

Theorem~\ref{thm:composition} shows the
capability and limitation of a dimensionally independent sampling scheme called {\it analytical sampling} \citep{sunscore} or Tweedie $\tau$-leaping \citep{loudiscrete,ou2024your},
where
we use a product-model denoiser
$p_{s|t}(\bm{x}_s|\bm{x}_t)=\prod_{d=1}^Dp_{s|t}^d(x_s^d|\bm{x}_t)$
approximating the true
marginal as $p_{s|t}^d(x_s^d|\bm{x}_t) \approx q_{s|t}^d(x_s^d|\bm{x}_t)$.
Although commonly used,
there has been only empirical evidence for the overall efficiency of this dimensionally independent method.
Note that
\citet{campbell2022continuous}
provides a guarantee for another
dimensionally independent method called $\tau$-leaping.

\begin{thm}[$N$-step analytical sampling approximates data, informal]\label{thm:composition}
    Let $q_{t|s}$ be forward transition probabilities
    that factorize as above
    and
    $p_{s|t}$ be a product model with the correct
    marginals, i.e.,
    $p_{s|t}(\bm{x}_s|\bm{x}_t)=\prod_{d=1}^Dq_{s|t}^d(x_s^d|\bm{x}_t)$ for $s<t$.
    Under regularity conditions,
    given timesteps $t_i=iT/N$ for $i=0,\ldots,N$,
    we have, as $N\to\infty$,
    \begin{equation}
        d_\mathrm{TV}\!\left(q_0, \E[\bm{x}_T\sim q_T]{
            p_{t_0|t_1} 
            \circ\cdots\circ
            p_{t_{N-1}|t_N}
            (\cdot|\bm{x}_T)
        }\right)
        = \ord{1/N}.
        \label{eq:informal-main}
    \end{equation}
    
    Moreover, there is an example with $\lvert\S\rvert = D = 2$ such that
    the left-hand side of \eqref{eq:informal-main}
    is lower-bounded by $c/N$ with some constant $c>0$
    for sufficiently large $N$.
\end{thm}
\begin{proof}[Proof (sketch)]
    We first prove
    the following estimate
    for $0\le t-\ve < t\le T$
    and $\bm{x}\in\S^D$
    (Lemma~\ref{lem:lem-comp-formal}):
    \begin{equation}
        d_\mathrm{TV}(q_{t-\ve|t}(\cdot|\bm{x}),
        p_{t-\ve|t}(\cdot|\bm{x})) =\ord{\ve^2},
        \label{eq:proof-lem-recall}
    \end{equation}
    as $\ve \to 0$.
    The proof
    exploits the factorization
    $q_{t|s}(\bm{x}_t|\bm{x}_s)
    =\prod_{d=1}^Dq_{t|s}^d(x_t^d|x_s^d)$
    and is based on additional continuity assumptions.
    We then decompose the left-hand side of
    \eqref{eq:informal-main} into $N$ terms
    by using a triangle-like inequality
    on $d_\mathrm{TV}$
    between compositional distributions
    (Proposition~\ref{prop:tv-triangle}).
    The $i$-th term essentially
    measures
    the distance between
    $q_{t_{i-1}|t_{i}}$
    and $p_{t_{i-1}|t_{i}}$
    and thus is bounded by $\ord{1/N^2}$
    from \eqref{eq:proof-lem-recall}
    with $\ve=T/N$.
    By summing up the $N$ terms,
    we obtain the desired estimate for the first part.
    For the second part, we actually construct a concrete example
    in Section~\ref{sec:lower-bound}.
\end{proof}
See Theorem~\ref{thm:comp-formal}
for a formal version.
Theorem~\ref{thm:composition} is important as
it underpins the use of dimensionally parallel denoising given sufficient steps,
which has been claimed as an advantage of discrete diffusion
over autoregressive models whose sampling is sequential \citep{loudiscrete}.
However, it still requires $\Omega(1/\ve)$
steps in order to have a uniform error bound $\ve$,
according to the latter half of the assertion.
We show next that we can further reduce the number of steps
with our loss functions,
by distilling the distribution
of an $N$-step teacher model into a few-step student model
by learning dimensional correlations.

\subsection{Our losses can
distill multi-step denoising models}
\label{sec:distil-theory}
Let $p^\psi$ and $p^\theta$ respectively be the teacher and student models given in Sections
\ref{sec:compose-motivation}~\&~\ref{sec:losses}. 
The following statement gives a theoretical guarantee
for using the proposed loss functions at the appropriate time and distribution settings.
\begin{thm}[Di4C student approximates $N$-step teacher]\label{thm:main}
    Let $0=t_0<\cdots<t_N=T$
    be timesteps and $r_T$
    be a probability distribution on $\X$.
    For each $n$, let $r_{t_n} = \E[\bm{x}_T\sim r_T]{p^{\psi}_{t_n|t_{n+1}}
        \circ\cdots\circ p^{\psi}_{t_{N-1}|t_N}(\cdot|\bm{x}_T)}$.
    Then, we have
    \icml{
        \begin{align}
            &d_\mathrm{TV}\!\left(r_0,
            \E[\bm{x}_T\sim r_T]{p^\theta_{0|T}(\cdot|\bm{x}_T)}\right)
            \nonumber\\
            &\le
            \frac1{\sqrt{2}}\Bigl(
                \mathcal{L}_\mathrm{distil}(\theta;\psi,r_{t_1},t_1)^{1/2}
            \nonumber\\
            &\quad\qquad 
            +\sum_{n=1}^{N-1}
                \mathcal{L}_\mathrm{consis}
                (\theta;\psi,r_{t_{n+1}},0,t_n,t_{n+1})^{1/2}
            \Bigr).
            \label{eq:main-ineq}
        \end{align}
    }{
        \begin{equation}
            d_\mathrm{TV}\!\left(r_0,
            \E[\bm{x}_T\sim r_T]{p^\theta_{0|T}(\cdot|\bm{x}_T)}\right)
            \le
            \frac1{\sqrt{2}}\left(
                \mathcal{L}_\mathrm{distil}(\theta;\psi,r_{t_1},t_1)^{1/2}
            +\sum_{n=1}^{N-1}
                \mathcal{L}_\mathrm{consis}
                (\theta;\psi,r_{t_{n+1}},0,t_n,t_{n+1})^{1/2}
            \right).
            \label{eq:main-ineq}
        \end{equation}
    }
\end{thm}
We can prove this by formalizing
the intuition behind Figure~\ref{fig:loss-illustration} (see Section~\ref{sec:proof-main}).
Note that
the right-hand side of inequality \eqref{eq:main-ineq}
becomes zero (so does the left-hand side)
if the student model perfectly learns the composition of the teacher
as in \eqref{eq:student-teacher-target},
so learning with these loss functions is feasible in theory
if the student model has enough expressive power.
Existing theoretical guarantees in consistency-based distillation
of continuous-state diffusions typically discuss the case
when consistency losses are exactly zero \citep{song2023consistency,daras2024consistent,lai2023equivalence},
so our guarantee would be interesting in that
it explicitly shows
the relationships between the magnitude of loss functions
and the total variation bound between the teacher and student.

Regarding the choice of $r_t$,
we should take $r_T=q_T$ if we would like to combine Theorem~\ref{thm:main} with Theorem~\ref{thm:composition} to evaluate Di4C's overall performance against the data distribution.
For $r_t$ with $t<T$,
though we can generate samples $\bm{x}_t\sim r_t$ by using
the teacher model, it might be expensive due to the multi-step inference required.
Instead, we can use $q_t$ if we have access to data,
which is given by just one-step
forward sampling from $q_{t|0}(\cdot|\bm{x}_0)$
with the data $\bm{x}_0\sim q_0$.
Since $r_t$ is an approximation of $q_t$ (Theorem~\ref{thm:composition}),
it would not harm the training quality
as long as the teacher model is well-trained.

\section{Experimental results}\label{sec:experiments}
We evaluated our Di4C method in three different diffusion settings,
each with distinct teacher models and state spaces.
First, we examined continuous-time discrete-state diffusion with pixel-space representations
using CIFAR-10, where we distilled from a well-trained U-Net teacher model (Section~\ref{sec:tldr}).
Second, we explored masked diffusion on vector-quantized (VQ) space using ImageNet,
working with a transformer-based teacher model designed for masked image generation
(Section~\ref{sec:ex-maskgit}).
Third, we tested our approach on masked diffusion language models trained on OpenWebText,
demonstrating Di4C's effectiveness in distilling transformer-based diffusion language models
(Section~\ref{sec:mdlm}).
Finally, we demonstrate that our mixture modeling
causes minimal latency overhead in all the above experiments (\Cref{sec:latency}).
These experimental results, spanning different domains, architectures, and diffusion processes,
showcase the broad applicability of our method while highlighting
its consistent ability to achieve faster sampling.

\subsection{Discretized Gaussian diffusion on pixel space}
\label{sec:tldr}
In our first experiment, we adopted the same setting as \citet{campbell2022continuous}:
a continuous-time discrete-state Markov process (of discretized Gaussian transition) with the CIFAR-10 image dataset~\citep{krizhevsky2009learning}.
We used the well-trained model checkpoint provided by \citet{campbell2022continuous} as our product teacher model ($p^\psi$).
This model outperforms previous discrete-time discrete-state models such as  \citet{austin2021structured}.
As in the original paper,
we worked directly with the discrete pixel channel values (0 to 255)
on $32\times 32 \times 3$ entries
(i.e., $\lvert\S\rvert= 256$, $D=3072$).
The teacher model $p^\psi$ has a U-net architecture \citep{ho2020denoising} tailored for discrete diffusion, which is fed a time feature at each upsampling/downsampling stage.

To obtain an architecture for our student mixture model \eqref{eq:mixture-model},
we slightly extended the teacher's architecture
so that it accepts a conditioning with $\lambda \sim\mathrm{Unif}([0, 1])$
(uniform distribution over $[0,1]$) in this experiment,
by following the original implementation of time conditioning.
In training, the student model was initialized by the teacher network parameters
with additional zero-initialized subnetworks concerning $\lambda$.
See Section~\ref{sec:implementation} for implementation details.

\begin{table}[h]
    \centering
    \caption{Comparison of models on CIFAR-10 dataset.
    Fr\'echet inception distance (FID~$\downarrow$) against training dataset
    and inception score (IS~$\uparrow$) measured with 50,000 samples are shown in this order
    (FID / IS).}
    \label{tab:tldr}
    \vspace{2mm}
    \begin{tabular}{cccc}
        \toprule
                & 10 steps & 20 steps & 40 steps  \\
        \midrule
        teacher & 32.61 / 7.59 & 12.36 / 8.55 & {\bf8.01} / {\bf8.77}\\
        \midrule
        student & {\bf20.64} / {\bf8.29} & 9.77 / 8.52 & 9.66 / 8.28 \\
        hybrid & 25.54 / 8.00 & {\bf9.47} / {\bf8.56} & 8.02 / 8.43 \\
        \bottomrule
    \end{tabular}
\end{table}

Table~\ref{tab:tldr} shows the results.
The ``hybrid'' model used the student model for the first half of the denoising process
and then switched to the teacher model for the remaining steps during inference.
The student substantially improved the metrics compared with the teacher
in 10-step sampling,
while the benefits of our method diminished as the number of steps grew.
In contrast, the hybrid model
was the best at 20 steps
and on par with the teacher in 40-step FID.
We hypothesize that Di4C is particularly effective when using fewer sampling steps,
as this is where capturing complex dimensional correlations is crucial (Theorem~\ref{thm:composition}).
We also give additional results for different sampling steps in \Cref{tab:more-tldr}.

\begin{table}[h]
    \centering
    \caption{Comparison of teacher and student with PC steps. FID against CIFAR-10 training dataset measured with 10,000 generated samples are displayed (so there is discrepancy from \Cref{tab:tldr}).
    Number of function evaluations (NFEs) is adjusted to 20, with varying budget allocations between denoising and PC steps.}
    \label{tab:pc}
    \vspace{2mm}
    \begin{tabular}{ccccc}
        \toprule
            NFE & 6{\scriptsize +7+7} & 10{\scriptsize +5+5} & 14{\scriptsize +3+3} & 20{\scriptsize +0+0}  \\
        \midrule
        teacher & 57.57 & 32.74 & 21.32 & 14.42 \\
        student & 44.57 & 22.63 & 14.25 & 11.81 \\
        \bottomrule
    \end{tabular}
\end{table}
In \Cref{tab:pc}, we also compare the teacher and student models in combination with
the predictor-corrector \citep[PC,][]{campbell2022continuous} steps.
The notation $m${\scriptsize+$n$+$n$} indicates that we used $n$ corrector steps before each of the final two out of $m$ denoising steps,
imitating the approach in \citet{campbell2022continuous}.
For each setting, the best PC step-size among $\{10^{-3}, 10^{-4}, 10^{-5}\}$ was chosen.
Since a PC step requires an additional network evaluation (as expensive as one step of denoising),
we adjusted the total NFE in the table.
While adding PC steps shows improvement upon the vanilla $m$-step denoising (see also \Cref{tab:more-tldr})
in both teacher and student models,
allocating all the NFE budget to denoising steps performs better in this small NFE regime;
this observation aligns with \citet[Fig.~4]{campbell2022continuous}.

\subsection{Masked image modeling on VQ space}\label{sec:ex-maskgit}

Next, we evaluated our method by applying it to a larger-scale image generation model. For this purpose, we adopted the framework of MaskGIT~\citep{chang2022maskgit} and worked on the ImageNet dataset~\citep{deng2009imagenet} at $256\times256$ resolution.
MaskGIT is one of the state-of-the-art image generation methods, based on masked diffusion modeling.
Its generative (backward) process relies on heuristics including confidence-based sampling (Section~\ref{sec:mg-sampling}).
The variant we used also uses discrete classifier-free guidance (CFG; Section~\ref{sec:cfg}).
We demonstrate that Di4C can enhance image generation even in combination with such heuristics.

In our setting, the model comprised two main components:
a VQGAN~\citep{esser2021taming} pre-trained on the ImageNet dataset
and a masked diffusion model trained on the VQ space.
The VQGAN encodes a $256\times256$ resolution image into $16\times16=256$ tokens,
each drawn from a shared codebook $\S^*$ of size $1024$.
The forward process of our diffusion modeling,
denoted as $q_{t|0}(\cdot|\bm{x}_0)$,
independently replaces $x_0^d$ with $\text{\ttfamily[MASK]}$ at a certain probability $m_t$ for each $d$ (i.e., $\S:=\S^* \cup \{\!\text{\ttfamily [MASK]}\!\}$, $|\S^*|=1025$, and $D=256$).
The masking probability $m_t$ increases monotonically from $m_0=0$ to $m_1=1$, resulting in completely masked sequences at $t=1$ regardless of $\bm{x}_0$.

As the teacher model,
we used the PyTorch-based implementation by \citet{besnier2023pytorch},
which replicates the performance of the original MaskGIT.
It uses a bidirectional transformer $p^\psi$
that estimates the distribution of each token
given an ImageNet label $c$ and a partially masked sequence $\bm{x}_t \in \S^D$:
$p_{0|t}^{\psi,d}(\cdot|\bm{x}_t, c)
\approx q_{0|t}^d (\cdot|\bm{x}_t, c)$.
The input sequence length is $257$ including the embedding of $c$.
Additionally, the model supports unconditional generation,
enabling discrete CFG to match the original MaskGIT's performance.
In this configuration, the teacher model generates high-quality samples with only $8$ steps,
which is significantly faster than typical diffusion-based generative models
\citep[see, e.g.,][Table~1]{chang2022maskgit}.
In the implementation of our mixture (student) model,
we simply added a single token embedding coming from $\lambda\sim\mathrm{Unif}([0,1])$,
so the input sequence length is $258$ instead of $257$ in the student models.

\icml{
    \begin{figure}[h]
        \centering
        \includegraphics[width=\linewidth]{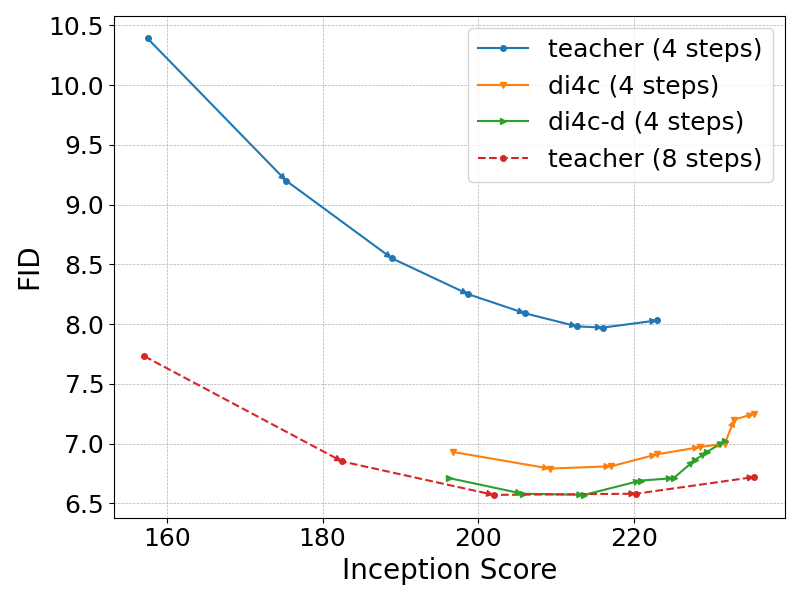}
        \vspace{-7.5mm}
        \caption{FID-IS curves of $4/8$-step teacher and $4$-step Di4C models
        on ImageNet~$256\times256$
        when varying CFG coefficients.
        Arrows connect experimental results (dots) at each CFG coefficient in ascending order.}
        \label{fig:maskgit}
    \end{figure}

    \begin{figure*}[!t]
        \centering
        \subfigure[Gen.~PPL vs Num.~Steps in unconditional generation]{
            \includegraphics[width=0.45\textwidth]{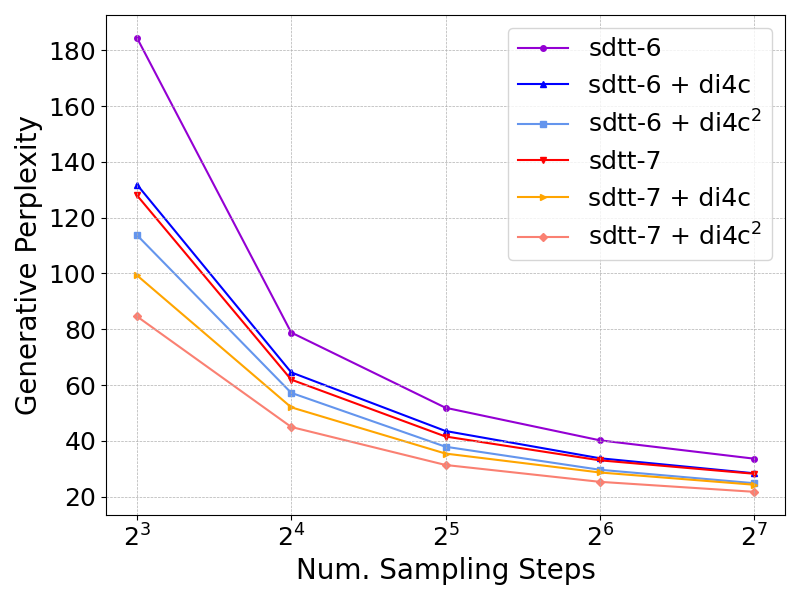}
            \label{fig:uncond-ppl}
        }
        \subfigure[Gen.~PPL vs Self-BLEU in conditional generation]{
            \includegraphics[width=0.45\textwidth]{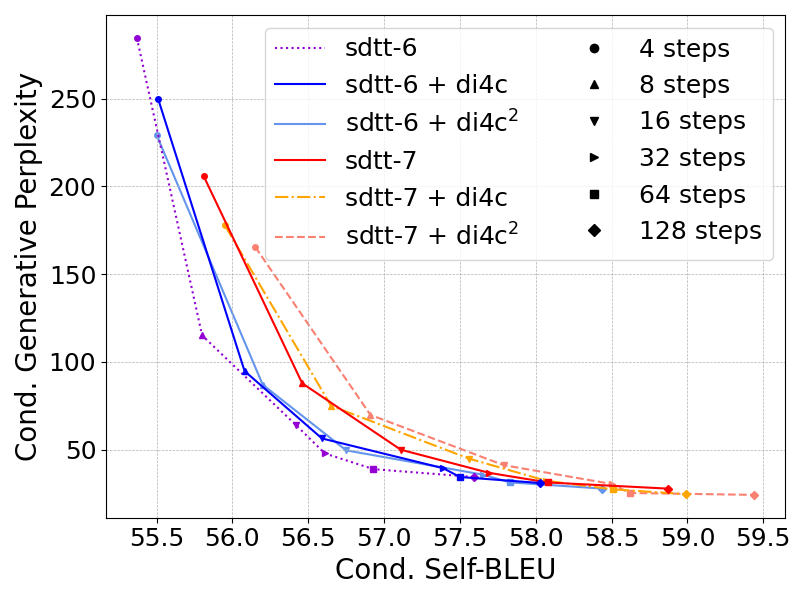}
            \label{fig:ppl-sbleu}
        }
        \vspace{-2mm}
        \caption{Comparison of SDTT checkpoints \citep{deschenaux2024beyond} and their Di4C distillations.}
        \label{fig:language}
    \end{figure*}
}{
    \begin{figure}[h]
        \centering
        \includegraphics[width=0.6\linewidth]{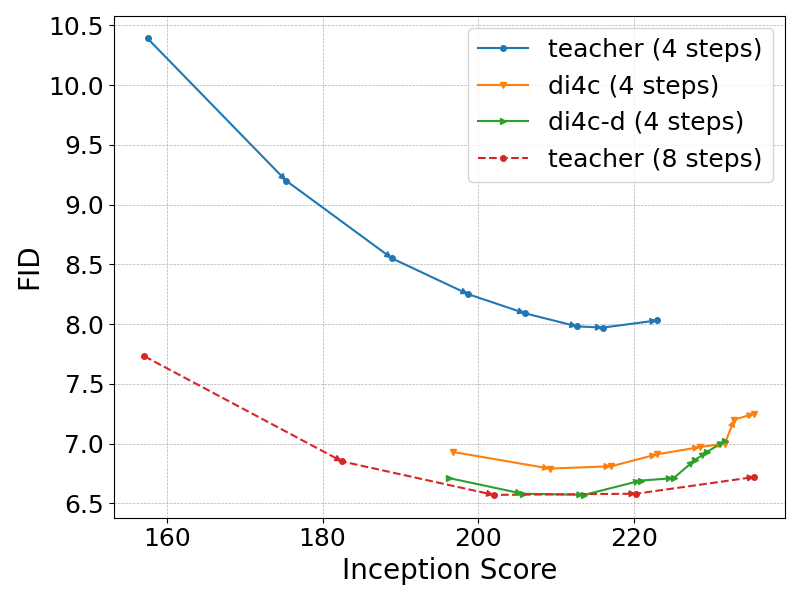}
        \caption{FID-IS curves of the $4/8$-step teacher and $4$-step Di4C models
        on ImageNet~$256\times256$
        when varying CFG coefficients.
        The arrows connect the experimental results (dots) at each CFG coefficient in ascending order.}
        \label{fig:maskgit}
    \end{figure}
}

Figure~\ref{fig:maskgit} shows
FID-IS curves of 4-step sampling with various CFG coefficients:
$w_\mathrm{cfg}\in\{5,6,7,8,9,10,11,12\}$ for 4 steps
and $w_\mathrm{cfg}\in\{2, 2.5, 3, 3.5, 4\}$ for 8 steps (teacher only).
In the figure, {\bf di4c} represents a model trained using the standard Di4C method,
while \textbf{\mbox{di4c-d}} represents a model trained with an additional datapoint loss (see Section~\ref{sec:mg-training} for details).
As shown in the figure, the FID-IS performance of the 4-step Di4C models closely matched that of the 8-step sampling of the teacher model.
This result indicates that Di4C can achieve an approximate 2x speed-up in the sampling process.

See Section~\ref{sec:mg} for further details of this experiment.
In particular, \Cref{tab:cfg-choice} shows the detailed quantitative performance (including precision and recall) of each model at its best CFG coefficient.

\subsection{Masked diffusion language models}\label{sec:mdlm}
Finally, we examined Di4C in language modeling.
As teacher models,
we adopted two versions of pretrained SDTT models~\citep[with a KL target]{deschenaux2024beyond},
which we refer to as \textbf{sdtt-6} and \textbf{sdtt-7}.
They were obtained after respectively 6 and 7 rounds of distillation
of a masked diffusion language model
(MDLM, \citealp{sahoo2024simple}. See also its concurrent works: \citealt{shi2024simplified}; \citealt{ou2024your}) trained
on the OpenWebText dataset~\citep{gokaslan2019openwebtext}.
As a forward process, they use a masked diffusion with a GPT-2 tokenizer~\citep{radford2019language},
which is essentially the same as the one explained in Section~\ref{sec:ex-maskgit},
while we had $D=1024$ and $\lvert \S^*\rvert=50257$ this time.
In the mixture modeling for Di4C,
we added one token from $\lambda\sim\mathrm{Unif}([0, 1])$
to the transformer,
similarly to the previous section.
See Section~\ref{sec:app-mdlm} for details.

\icml{}{
    \begin{figure}[h]
        \centering
        \subfigure[Gen.~PPL vs Num.~Steps in unconditional generation]{
            \includegraphics[width=0.45\textwidth]{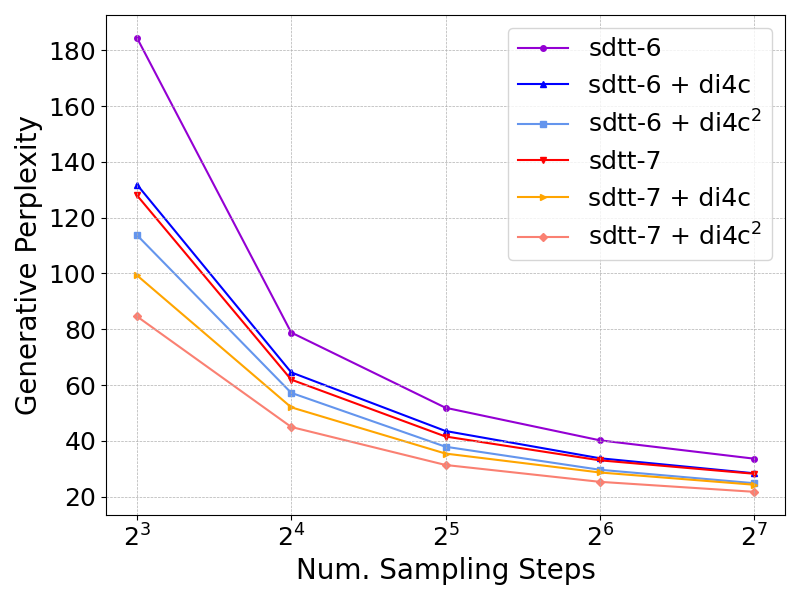}
            \label{fig:uncond-ppl}
        }
        \subfigure[Gen.~PPL vs Self-BLEU in conditional generation]{
            \includegraphics[width=0.45\textwidth]{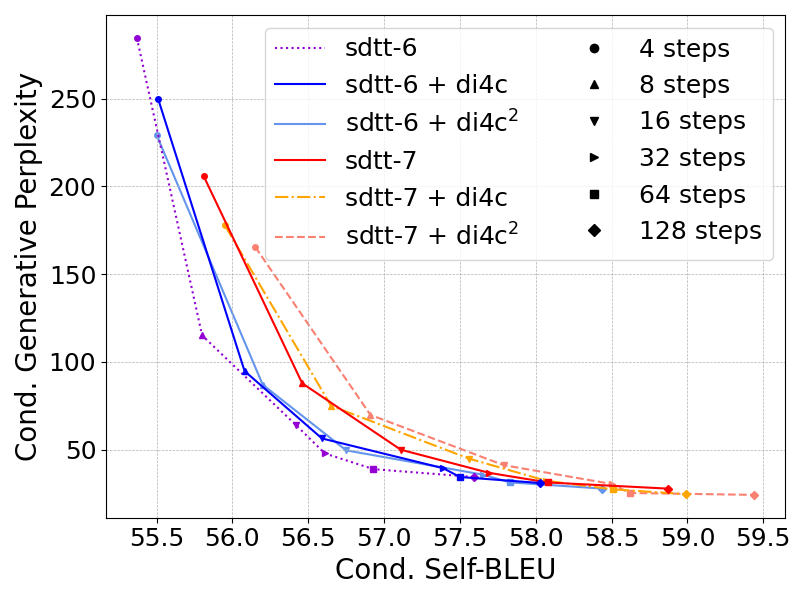}
            \label{fig:ppl-sbleu}
        }
        \vspace{-2mm}
        \caption{Comparison of SDTT checkpoints \citep{deschenaux2024beyond} and their Di4C distillations.}
        \label{fig:language}
    \end{figure}
}

Figure~\ref{fig:language} shows the results of applying Di4C to the SDTT models.
For $n = 6,7$,
\textbf{sdtt-$\boldsymbol{n}$ + di4c} represents the model obtained by fine-tuning
\textbf{sdtt-$\boldsymbol{n}$} using Di4C.
Similarly, \textbf{sdtt-$\boldsymbol{n}$ + di4c$^{\boldsymbol{2}}$}
represents the model obtained by applying another round of
Di4C to \textbf{sdtt-$\boldsymbol{n}$ + di4c}
(see Section~\ref{sec:iterative-di4c}).

\paragraph{Unconditional generation.}
Figure~\ref{fig:uncond-ppl} shows a performance comparison for unconditional generation.
Each model generated 1024 samples of 1024 tokens, with the number of sampling steps in $\{8, 16, 32, 64, 128\}$.
Following the previous work~\citep{loudiscrete,sahoo2024simple,deschenaux2024beyond},
we used the GPT-2-large model~\citep{radford2019language} to evaluate the generative perplexity.
The results show that applying one round of Di4C lead
to a similar level of improvement as applying one round of SDTT
(compare \textbf{sdtt-6 + di4c} with \textbf{sdtt-7}).
Also, two rounds of Di4C (\textbf{+ di4c$^{\boldsymbol{2}}$})
showed speed-ups of more than 2x to match the teacher's performance of 64 or 128 steps.

\paragraph{Conditional generation.}
We also tested the
quality-diversity tradeoff in conditional generation.
We used 256 samples from the WebText dataset~\citep{webtext},
and each model, on the basis of the first 50 tokens of the sample,
generated 5 continuations of 50 tokens,
following \citet{deschenaux2024beyond}.
We computed the generative perplexity of the generated continuations
and the Self-BLEU score~\citep{zhu2018texygen}
of the 5 completed samples starting from the same prompt,
and averaged them over the 256 prompts.
Self-BLEU was computed by GPT-2 tokenizer with equal weights on $n$-gram
for $n=1,2,3,4$
and scaled from 0 to 100 (lower Self-BLEU indicates higher diversity).
Note that (Self-)BLEU has several parameters including tokenization
and maximum length for $n$-grams
\citep{post2018call},
so our numbers are not directly comparable to those from an existing work
such as \citet{agarwal2024policy}.

Figure~\ref{fig:ppl-sbleu} shows the quality-diversity tradeoff curves of each model
with various numbers of sampling steps.
As shown with solid lines,
both \textbf{sdtt-6 + di4c} and \textbf{sdtt-6 + di4c$^{\boldsymbol{2}}$}
consistently achieved higher diversity than \textbf{sdtt-7}
with the same number of sampling steps
while maintaining comparable or better generative perplexity.

\subsection{Latency overhead of mixture modeling}
\label{sec:latency}
In this section, we demonstrate that the computational overhead in introducing mixture modeling
is negligible compared with its performance gain.
\Cref{tab:latency} shows the quantitative results on the latency.
The experiments listed in Table~\ref{tab:latency} correspond to those described in
Sections~\ref{sec:tldr}--\ref{sec:mdlm},
where the chosen student models are the best models
(\textbf{\mbox{di4c-d}} in ImageNet and \textbf{sdtt-7 + di4c$^{\boldsymbol{2}}$} in OpenWebText).
The table shows the average runtime and standard deviation,
calculated over 10 batches with batch sizes of 50, 64, and 16, respectively.
We tested unconditional generation for CIFAR-10 and OpenWebText,
and a class-conditional generation with a uniform random class for ImageNet
with a classifier-free guidance.
\begin{table}[h]
    \centering
    \caption{Latency comparison between teacher (product)
    and student (mixture) models.}
    \label{tab:latency}
    \vspace{2mm}
    \begin{tabular}{cc|cc}
        \toprule
        Experiment & \# steps& teacher (sec.) & student (sec.) \\
        \midrule
        CIFAR-10 & 10 & 0.5515\spm{0.0024} & 0.5786\spm{0.0017}\\
        ImageNet & 4 & 2.0741\spm{0.0035} & 2.0734\spm{0.0043}\\
        OpenWebText & 16 & 3.3409\spm{0.0417} & 3.4817\spm{0.0730}\\
        \bottomrule
    \end{tabular}
\end{table}

In the CIFAR-10 and OpenWebText experiments,
the overhead from using mixture models was up to 5\%,
which is negligible compared with their performance gain.
In the ImageNet experiment,
the difference between the student and teacher was within the statistical noise.
This is likely because the computational bottleneck in this case was not the logit inference itself
but rather the sampling process that followed it.

\section{Related work}\label{sec:related}
\paragraph{Speeding up continuous diffusion models.}
In continuous diffusion models,
knowledge distillation~\citep{luhman2021knowledge,salimans2022progressive,meng2023distillation,zheng2023fast}
and consistency-type techniques~\citep{song2023consistency,song2023improved,kimconsistency}
enable single- or few-step sampling,
most of which are tailored for probability flow ODEs.
Among the studies on continuous diffusion,
the work by \citet{lisoft} is particularly relevant to our mixture modeling approach.
They highlight the limited expressive power of unimodal Gaussian distributions
in denoising continuous diffusions and demonstrate that using Gaussian mixtures
can substantially reduce the number of sampling steps required in non-ODE diffusions.

\paragraph{Speeding up discrete diffusion models.}
For faster sample generation in discrete diffusion models,
\citet{jys} propose
a post-hoc optimization of sampling schedules
and outperforms the uniform partitioning of $[0, T]$.
To speed up MDLMs \citep{sahoo2024simple}; 
SDTT \citep{deschenaux2024beyond} gathers the logits at unmasked tokens
from a pretrained teacher MDLM throughout the denoising process,
achieving a speed-up of approximately 32x compared with the teacher model.
In addition, the confidence-based sampling by MaskGIT~\citep{chang2022maskgit}
is essentially a heuristic to enable faster sampling in masked diffusions.
These approaches can be combined with Di4C;
we indeed have demonstrated it for the latter two.

Two concurrent works are particularly relevant to our research, both of which attempt to incorporate dimensional correlations into discrete diffusion, specifically within the MDLM context. 
\citet{liu2024discrete} combine pretrained autoregressive and masked diffusion models,
achieving superior performance compared with using either model alone.
Meanwhile, \citet{xu2024energy} propose an energy-based model to
modify the dimensionally independent sampler.
While both approaches improve upon the vanilla MDLM,
their sampling processes can incur some time and memory overhead.
This is due to the use of an additional non-diffusion model in both methods,
and the reliance on importance sampling in the latter.
In this regard,
our mixture modeling allows for a simpler sampling process with minimal modification to the original diffusion.
However, since the Di4C loss functions are model-agnostic,
we should explore the combination of Di4C losses and other proposed models capturing dimensional correlations
in future research.

\section{Conclusion}\label{sec:conclusion}
In this paper,
as current discrete diffusion models
ignore the dimensional correlations that need to be incorporated
to realize few-step models,
we proposed Di4C, a method for distilling pretrained discrete diffusion models.
Di4C provides a set of loss functions for
models that can capture dimensional correlations,
an example of which is the mixture model.
As a theoretical contribution,
we proved that the existing discrete diffusion models with many steps
can indeed recover a data distribution,
even without modeling dimensional correlations.
We also proved that such many-step models can be distilled into few-step ones,
if we use the Di4C loss functions with a model that has enough expressive power, such as a mixture model.
In numerical experiments, we confirmed the efficiency of our framework upon teacher models across multiple domains:
improving sample quality in 10-step sampling on CIFAR-10 with a discretized Gaussian diffusion,
achieving a 2x speed-up in ImageNet $256\times256$ generation with masked image modeling,
and successfully distilling already-distilled masked diffusion language models on OpenWebText while maintaining generation diversity.

However, there are still problems to be solved.
For example, although we can distill many-step models
into one-step ones in theory (Theorem~\ref{thm:main}),
our empirical results show only 2x fewer steps or so.
To address this point, we need to further optimize the architecture (mainly concerning $\lambda$)
and training hyperparameters,
while the iterated Di4C training (Section~\ref{sec:iterative-di4c})
can be a promising candidate.

\icml{
    \section*{Impact Statement}
    This paper presents a novel method for reducing the number of sampling steps
    required for high-quality generation in discrete data domains.
    While there are many potential social impacts as usual in the generative AI field,
    we do not foresee any specific problems particularly caused by this (mostly theoretical) work.
}{}

\section*{Acknowledgments}
We would like to thank Masato Ishii, Chieh-Hsin Lai, Naoki Murata, Bac Nguyen, and Toshimitsu Uesaka
for their constructive feedback.
We also thank the anonymous reviewers for their valuable comments and suggestions.

\bibliography{cite}
\bibliographystyle{icml2025}


\newpage
\appendix
\onecolumn

\section{Training techniques for Di4C}
\label{sec:training-di4c}
In this section,
we review the novel loss functions of Di4C
and the mixture model given in Section~\ref{sec:losses} from an algorithmic perspective, and we provide a set of techniques to stably train it. Specifically, we introduce techniques to make the computation of the loss functions scalable through Monte Carlo integration and control variate methods.

Before going into the details of the training techniques,
we introduce two auxiliary loss functions, which we can use in addition to $\L_\mathrm{distil}$ and $\L_\mathrm{consis}$
for practical improvements.
One is the
{\it datapoint loss}
that directly computes
the negative log-likelihood with respect
to the data distribution
\citep[e.g.,][Eq.~5]{austin2021structured},
which we can use
when we have access to data $q_0$:
\begin{equation}
    \L_\mathrm{data}(\theta; t):=
    \E[(\bm{x}_0,\bm{x}_t)\sim q_{0,t}]{-\log p^\theta_{0|t}(\bm{x}_0|\bm{x}_t)}.
    \label{eq:data-loss}
\end{equation}
The other is the following {\it marginal loss},
which is easier to compute,
under the assumption that the teacher model
sufficiently learns the true marginal,
i.e., $p^{\psi,d}_{0|t}\approx q^d_{0|t}$:
\begin{equation}
    \L_\mathrm{marginal}(\theta; \psi, r_t, t)
    := \E[\bm{x}_t\sim r_t]{
        \sum_{d=1}^D
        D_\mathrm{KL}(
            p_{0|t}^{\psi,d}(\cdot|\bm{x}_t)
            \,\Vert\,
            p_{0|t}^{\theta,d}(\cdot|\bm{x}_t)
        )}.
    \label{eq:marginal-loss}
\end{equation}

\subsection{Surrogate of distillation loss}\label{sec:distil-loss}
Since the exact evaluation of $\L_\mathrm{distil}$ with a mixture model
seems intractable, we consider an upper bound of $\tilde{\L}_\mathrm{distil}$ as a practical alternative:
\begin{align*}
    \mathcal{L}_\mathrm{distil}(\theta;\psi,r_\delta,\delta)
    &=\E[\bm{x}_{\delta}\sim r_\delta]{
        D_{\mathrm{KL}}(
        p_{0|\delta}^{\psi}(\cdot|\bm{x}_\delta)\,\Vert\,
        \mathbb{E}_\lambda[p_{0|\delta}^{\theta}(\cdot|\bm{x}_{\delta};\lambda)])
        }
    \\
    &\le
    \mathbb{E}_{\bm{x}_{\delta}\sim r_\delta}\E[\lambda]{
        D_{\mathrm{KL}}(
        p_{0|\delta}^{\psi}(\cdot|\bm{x}_\delta))\,\Vert\,
        p_{0|\delta}^{\theta}(\cdot|\bm{x}_{\delta};\lambda)
        } \\
    &\le
    \E[\lambda, \bm{x}_\delta\sim r_\delta]{
        \sum_{d=1}^D
        D_{\mathrm{KL}}(
        p_{0|\delta}^{\psi, d}(\cdot|\bm{x}_\delta)
        \,\Vert\,
        p_{0|\delta}^{\theta, d}(\cdot|\bm{x}_\delta;\lambda)
        )}
        =:\tilde{\mathcal{L}}_\mathrm{distil}(\theta;\psi,r_\delta,\delta).
\end{align*}
Here, the inequality is given by the convexity of KL divergence
(see Proposition~\ref{prop:kl}).
The upper bound $\tilde{\L}_\mathrm{distil}$ (and then $\L_\mathrm{distil}$) becomes zero
if the student denoiser coincides with the teacher for the time interval $[0,\delta]$, regardless of $\lambda$.
Therefore, the use of this upper bound is feasible
if $p^\theta$ has enough expressive power.

\subsection{Surrogate of consistency loss}\label{sec:consis-loss}
We consider $\L_\mathrm{consis}$ in this section.
As $p_{s|u}^\theta$ is more ``reliable'' than $p_{s|t}^\theta$
(since $s < u < t$),
we consider only the gradient of $\L_\mathrm{consis}$
concerning $p_{s|t}^\theta$
and ignore the gradient coming from $p_{s|u}^\theta$.
Therefore,
we conduct stochastic gradient descent on $\theta$ with the loss
\begin{equation}
    D_\mathrm{KL}(p_{s|u}^{\mathrm{sg}(\theta)}\circ p_{u|t}^{\psi}(\cdot|\bm{x}_t)
    \,\Vert\,
    p_{s|t}^\theta(\cdot|\bm{x}_t))= 
    H(p_{s|u}^{\mathrm{sg}(\theta)}\circ p_{u|t}^\psi(\cdot|\bm{x}_t), p_{s|t}^{\theta}(\cdot|\bm{x}_t))
    + const.,
    \label{eq:kl-cchain}
\end{equation}
where
$\mathrm{sg}(\cdot)$ is the stop-gradient operator \citep{van2017neural}
and $H(p, q)=\E[\bm{x}\sim p]{-\log q(\bm{x})}$ is the cross entropy between $p$ and $q$.
We hereby ignore the constant term in \eqref{eq:kl-cchain}
and consider how to efficiently compute the cross entropy term.

Most naively, by using finite samples $\bm{x}_s^{(1)},\ldots, \bm{x}_s^{(M)}
\sim_\mathrm{iid} p_{s|u}^{\mathrm{sg}(\theta)}\circ p_{u|t}^{\psi}(\cdot|\bm{x}_t)$
and $\lambda_1,\ldots,\lambda_N\sim_\mathrm{iid}\lambda$,
we can approximate this cross entropy by
two-fold Monte Carlo:
\begin{align}
    &H(p_{s|u}^{\mathrm{sg}(\theta)}\circ p_{u|t}^\psi(\cdot|\bm{x}_t), p_{s|t}^{\theta}(\cdot|\bm{x}_t))\nonumber\\
    &\approx
    -\frac{1}{M}\sum_{j=1}^M
    \log p_{s|t}^{\theta}(\bm{x}_s^{(j)}|\bm{x}_t)
    \approx -
    \frac1{M}\sum_{j=1}^M
    \log\left(\frac1N\sum_{i=1}^N p_{s|t}^{\theta}(\bm{x}_s^{(j)}|\bm{x}_t;\lambda_i)\right).
    \label{eq:monte-carlo}
\end{align}
Although the value of each $p_{s|t}^{\theta}(\bm{x}_s^{(j)}|\bm{x}_u;\lambda_i)
= \prod_{d=1}^D p_{s|t}^{\theta,d}(x_s^{(j),d}|\bm{x}_u;\lambda_i)$
can be extremely small due to the $D$-fold product,
we can exploit the log-sum-exp structure:
\[
    \log\left(\sum_{i=1}^N p_{s|t}^{\theta}(\bm{x}_s^{(j)}|\bm{x}_t;\lambda_i)\right)
    = \underbrace{\log\Biggl(\sum_{i=1}^N
    \exp}_{\text{log-sum-exp}} \left(\sum_{d=1}^D \log p_{s|t}^{\theta,d}(x_s^{(j), d}|\bm{x}_t;\lambda_i)\right)\Biggr),
\]
which is implemented as a function
with additional stabilization to avoid under/overflows
in some of the common numerical packages including PyTorch.
See \citep{blanchard2021accurately}
for details of numerical properties associated with the
log-sum-exp structure.

\paragraph{Dimensionally independent control variate.}
Although naive Monte Carlo sampling with a sufficiently large sample size can approximate the left-hand side of Eq.~\eqref{eq:monte-carlo} well, a small batch can cause high variance in the evaluation of the expected values.
An established way of stabilizing Monte Carlo integration
is to use so-called {\it control variates}
\citep{glasserman2004monte,oates2017control},
also known as a {\it baseline} in reinforcement learning \citep{williams1992simple}.
To estimate an expectation $\E{f}$,
we can subtract another function/random variable $g$,
called a {\it control variate},
whose integral we know or can compute more precisely than Monte Carlo,
and execute the Monte Carlo for $f-g$
by using the decomposition $\E{f} = \E{f-g} + \E{g}$.
See Section~\ref{sec:cv} for a more detailed explanation.
As a concrete application of this technique,
we below propose the use of a dimensionally independent control variate.

We first exploit
the compositional form of $p_{s|u}^{\mathrm{sg}(\theta)}\circ p_{u|t}^\psi(\cdot|\bm{x}_t)$,
which is more informative than $\bm{x}_s^{(j)}$,
the pure samples in the Monte Carlo approach.
We can write it in an expectation as follows:
\begin{align}
    p_{s|u}^{\mathrm{sg}(\theta)}\circ p_{u|t}^\psi(\cdot|\bm{x}_t)
    = \E[\lambda,\bm{x}_u\sim p_{u|t}^\psi(\cdot|\bm{x}_t)]{
        p_{s|u}^{\mathrm{sg}(\theta)}(\cdot|\bm{x}_u;\lambda)
    }.\label{eq:product_ts}
\end{align}
To simplify \eqref{eq:product_ts},
let us denote $q^\eta:=p_{s|u}^{\mathrm{sg}(\theta)}(\cdot|\bm{x}_u;\lambda)$ and
$q:=\E[\eta]{q^\eta}$
with $\eta=(\bm{x}_u, \lambda)$.
To construct an efficient control variate
given $q$,
we need a function $g$ such that
(i) it reasonably approximates $p_{s|t}^\theta(\cdot|\bm{x}_t)$
and (ii) $\E[\bm{x}\sim q]{g(\bm{x})}$ is easy to compute/approximate.
One such example is the product model defined as
\begin{equation}
    \overline{p}_{s|t}^{\theta}(\cdot|\bm{x}_t)
    := \prod_{d=1}^D \overline{p}^{\theta,d}_{s|t}(\cdot|\bm{x}_t),
    \qquad
    \overline{p}^{\theta,d}_{s|t}(\cdot|\bm{x}_t):=
    p^{\theta,d}_{s|t}(\cdot|\bm{x}_t)=
    \E[\lambda]{p^{\theta,d}_{s|t}(\cdot|\bm{x}_t;\lambda)}.
    \label{eq:marginal-product}
\end{equation}
We defer the explanation
of how (i) and (ii) are satisfied
to Section~\ref{app:cv-explanation}.
Given a control variate $\overline{p}_{s|t}^\theta(\cdot|\bm{x}_t)$,
we can decompose the loss computation:
\begin{equation}
     H(q, p_{s|t}^{\theta}(\cdot|\bm{x}_t))
     =
    \underbrace{\E[\bm{x}_s\sim q]{-
    \log p_{s|t}^{\theta}(\bm{x}_s|\bm{x}_t)
        + \log \overline{p}^{\theta}_{s|t}(\bm{x}_s|\bm{x}_t)
    }}_{\text{Monte Carlo by sampling $\bm{x}_s$}}
    +\underbrace{
        \E[\eta]{H(q^\eta, \overline{p}^\theta_{s|t}(\cdot|\bm{x}_t))}
    }_{\text{Monte Carlo by sampling $\eta$}}.
    \label{eq:consis-cv}
\end{equation}
Here, the first term can be treated similarly to \eqref{eq:monte-carlo},
and we approximately compute the second term
by sampling $\eta$
and using the identity
$H(q^\eta, \overline{p}^\theta_{s|t}(\cdot|\bm{x}_t))
= \sum_{d=1}^DH(q^{\eta,d}, \overline{p}^{\theta,d}_{s|t}(\cdot|\bm{x}_t))$
(see \eqref{eq:cv-computation} in Section~\ref{app:cv-explanation}).
In this decomposition,
we expect the mixture model to explicitly
learn the dimensional correlation with the first term,
while the second term stabilizes the overall approximation,
as we use more detailed information on $q$ than just its samples.
See also Section~\ref{sec:cv-derivation} for more background on how we derive $\overline{p}^\theta$
and another possible choice of control variate.


\subsection{Auxiliary losses}\label{sec:app-aux}
While we can use a similar Monte Carlo
estimate for $\L_\mathrm{data}$ (with random samples of $\bm{x}_0,\bm{x}_t,\lambda$),
we can regard $\L_\mathrm{marginal}$ as a possible control variate for it.
Indeed, if the teacher network is well-trained,
we can expect that its marginal approximates the true marginal
as $p^{\psi,d}\approx q^d$.
Thus, for the marginal-matching product model $\overline{p}^\theta$ given in Eq.~\eqref{eq:marginal-product},
we have
\begin{align}
    &\E[\bm{x}_t\sim q_t]{H(q_{0|t}(\cdot|\bm{x}_t),
        \overline{p}_{0|t}^\theta(\cdot|\bm{x}_t))}
        \approx \L_\mathrm{marginal}(\theta;\psi,q_t, t) + const.,
    \label{eq:cv-data}
\end{align}
where the constant term is independent of $\theta$.
We give the derivation of \eqref{eq:cv-data}
in Appendix~\ref{app:derivation-marginal}.
We then obtain a decomposed formulation of $\L_\mathrm{data}$ for given $\bm{x}_t\sim q_t$
as follows,
by letting $q=q_{0|t}(\cdot|\bm{x}_t)$ and $s=0$ in Eq.~\eqref{eq:consis-cv} and then using approximation~\eqref{eq:cv-data}:
\begin{align*}
    \L_\mathrm{data}(\theta;t)
    &\approx
     \L_\mathrm{corr}(\theta; t)     +\L_\mathrm{marginal}(\theta;\psi,q_t,t)
     +const.,
     \\
    &\L_\mathrm{corr}(\theta; t):=
    \E[(\bm{x}_0,\bm{x}_t)\sim q_{0,t}]{
        - \log p_{0|t}^{\theta}(\bm{x}_0|\bm{x}_t)
        + \log \overline{p}^{\theta}_{0|t}(\bm{x}_0|\bm{x}_t)}.
\end{align*}
Here, $\L_\mathrm{corr}$ measures
the difference between $p^\theta$ and $\overline{p}^\theta$
and thus represents
the dimensional correlation learned
by the model $p^\theta$.
In the actual implementation for the first term $\L_\mathrm{corr}$,
we generate $\bm{x}_0\sim q_0$ and then $\bm{x}_t\sim q_{t|0}(\cdot|\bm{x}_0)$,
and regard them
as samples from
$(\bm{x}_0, \bm{x}_t)\sim q_{0,t}$,
which are required for conducting
Monte Carlo.
When combining $\L_\mathrm{data}$ and $\L_\mathrm{marginal}$ (both as a loss and control variate),
we empirically find that
mixing them as 
$\alpha_t\L_\mathrm{corr}(\theta; t)+ \L_\mathrm{marginal}(\theta; \psi,q_t,t)$
with some $\alpha_t\in [0, 1]$
depending on $t$
is more efficient than just using constant
$\alpha_t=0$ (pure marginal loss)
or
$\alpha_t=1$ (pure data loss).
See Section~\ref{sec:expelimental-details} for details in this regard.

\subsection{Iterated Di4C training}\label{sec:iterative-di4c}

While we generally assume that the teacher model is given by a product model,
the Di4C loss functions can also treat mixture teacher models.
The only exception is the marginal loss (Section~\ref{sec:app-aux}),
but we can just replace $q_{0|t}$ with $\overline{q}_{0|t}$ in \eqref{eq:cv-data}
and conduct Monte Carlo estimates.
Through this generalization,
we can run multiple rounds of Di4C in a similar spirit as the multi-round SDTT~\citep{deschenaux2024beyond}.

\section{Kullback--Leibler
divergence and total variation distance}
\label{sec:KL-TV}
Let $p$ and $q$ be probability distributions
on the same finite set $\X$.
The KL divergence $D_\mathrm{KL}$
and the total variation distance $d_\mathrm{TV}$
are defined as follows:
\[
    D_\mathrm{KL}(p\,\Vert\,q):=
    \sum_{x\in\X} p(x)\log\frac{p(x)}{q(x)},
    \quad
    d_\mathrm{TV}(p, q)
    :=\sup_{A\subset\X}\lvert p(A) - q(A)\rvert
    =\frac12\sum_{x\in\X}\lvert p(x) - q(x)\rvert.
\]
Here, in the computation of $D_\mathrm{KL}$,
we ignore the term with $p(x)=0$ and,
if there is an $x$ with $p(x)>0$ and $q(x)=0$,
we then define $D_\mathrm{KL}(p\,\Vert\,q)=0$.
These two error criteria between distributions are bridged by the following inequality
(see, e.g., \citet{canonne2022short}).
\begin{prop}[Pinsker's inequality]\label{prop:pinsker}
    For probability distributions $p$ and $q$ on $\X$,
    we have
    \[
        d_\mathrm{TV}(p, q) \le \sqrt{\frac12
            D_\mathrm{KL}(p\,\Vert\,q)}.
    \]
\end{prop}

The convexity of KL divergence in the following
plays a role in the main body of the paper.
\begin{prop}[{\citealp[][Theorem 2.7.2]{cover2006elements}}]
\label{prop:kl}
    $D_\mathrm{KL}(p\,\Vert\,q)$ is convex
        with respect to the pair $(p, q)$.
        Namely, for $t\in[0, 1]$ and probability distributions
        $p_1, p_2, q_1, q_2$ on the same domain,
        we have
        \[
            D_\mathrm{KL}(tp_1+(1-t)p_2\,\Vert\,tq_1+(1-t)q_2)
            \le t D_\mathrm{KL}(p_1\,\Vert\,q_1) + (1-t)D_\mathrm{KL}(p_2\,\Vert\,q_2).
        \]
\end{prop}

We also use the following triangle-like inequality for
the total variation distance of compositions.
\begin{prop}\label{prop:tv-triangle}
    For probability distributions $p_1(\cdot|y), p_2(\cdot|y)$
    over $\X$ conditioned on $y\in\mathcal{Y}$
    and $q_1, q_2$ over $\mathcal{Y}$,
    we have
    \[
        d_\mathrm{TV}(\E[y\sim q_1]{p_1(\cdot|{y})}, \E[{y}\sim q_2]{p_2(\cdot|{y})})
        \le \E[{y}\sim q_1]{d_\mathrm{TV}(p_1(\cdot|{y}), p_2(\cdot|{y}))}
        + d_\mathrm{TV}(q_1, q_2).
    \]
\end{prop}
We give its proof in Section~\ref{sec:proof-tv-triangle}.

\section{Continuous-time Markov chains and Kolmogorov equations}
\label{sec:ctmc}
Let us discuss the Kolmogorov forward/backward equations
associated with continuous-time Markov chains.
While the arguments below are mostly a reorganization
of those given in previous studies \citep{campbell2022continuous,sunscore},
we explicitly track the continuity/nonzero assumptions
used in their derivations.
\subsection{Kolmogorov equations
in the general case}\label{sec:kolmogorov-general}
Let us consider a general Markov process over the continuous time interval $[0,T]$
and a discrete (finite) state space $\X$,
which is called a continuous-time Markov chain \citep{anderson2012continuous,campbell2022continuous}.
The starting block is the forward transition rate
in a short-time interval.
For $t < t +\ve$, assume the following equation
for the infinitesimal forward transition:
\begin{equation}
    q_{t+\ve|t}(y|x)
    = \delta_{y,x} + \ve Q_t(y,x) + o(\ve),
    \qquad \ve>0,
    \label{eq:transition-forward}
\end{equation}
where $\delta_{y,x}$ is the Kronecker delta and
$Q_t$ is a function
$\X\times\X\to\R$ called the transition rate.
Here,
for $s\le t<t+\ve$,
we have
\begin{align*}
    q_{t+\ve|s}(y|x)
    = \sum_{z}q_{t+\ve|t}(y|z)q_{t|s}(z|x)
    &= \sum_{z}(\delta_{y,z}+Q_t(y,z)\ve)q_{t|s}(z|x) + o(\ve)\\
    &= q_{t|s}(y|x) + \ve\sum_z Q_t(y,z)q_{t|s}(z|x) + o(\ve).
\end{align*}
This means that we have
$\partial_t^+ q_{t|s}(y|x) = \sum_z Q_t(y,z)q_{t|s}(z|x)$,
where $\partial_t^+$ is a right-derivative regarding $t$.
Under the condition that $Q_t$ is continuous over $[0, T]$
(assume it is continuously extended to $t=T$,
though it is not necessary right now)
and $q_{t|s}$ is continuous over $t\in[s, T]$,
$q_{t|s}$ becomes differentiable
over the open interval (from a general fact in analysis \citep{975094})
and we have the Kolmogorov forward equation
for $t\in (s, T)$:
\begin{equation}
    \partial_t q_{t|s}(y|x) = \sum_z Q_t(y,z)q_{t|s}(z|x).
    \label{eq:forward-kolmogorov}
\end{equation}
Now, let us derive the backward equation.
For $s<s+\ve\le t$, by using \eqref{eq:transition-forward}, we have
\begin{align*}
    q_{t|s}(y|x) = \sum_{z}q_{t|s+\ve}(y|z)q_{s+\ve|s}(z|x)
    &= \sum_z q_{t|s+\ve}(y|z)(\delta_{z,x}+\ve Q_s(z, x))
    + o(\ve)\\
    &= q_{t|s+\ve}(y|x) + \ve \sum_z q_{t|s+\ve}(y|z) Q_s(z, x) + o(\ve).
\end{align*}
Thus, by additionally
assuming the continuity of $q_{t|s}$ for $s\in [0, T]$,
we obtain the one-sided derivative
$\partial_s^+ q_{t|s}(y|x) =
- \sum_z q_{t|s}(y|z)Q_s(z, x)$.
When combined with the continuity of $Q_s$ similarly to the above argument on the forward equation,
it leads to the backward Kolmogorov equation for $s\in (0, t)$:
\begin{equation}
    \partial_sq_{t|s}(y|x) =
- \sum_z q_{t|s}(y|z)Q_s(z, x).
    \label{eq:backward-kolmogorov}
\end{equation}
To summarize so far,
under the assumption that
$q_{t|s}$ is continuous for $s,t$ with $0\le s\le t \le T$
and $Q_t$ in \eqref{eq:transition-forward} is continuous
over $[0, T]$,
we have the two Kolmogorov equations given by \eqref{eq:forward-kolmogorov} and \eqref{eq:backward-kolmogorov}.
Note that all the $\sum_z$ are finite sums
because of the finiteness of $\X$.

\subsection{Kolmogorov equations
for factorized forward processes}
\label{sec:kolmogorov-factorized}
Let us now consider the case where $\X=\S^D$
and $\bm{x}_t = (x_t^d)_{d=1}^D$ follows
a dimensionally independent forward process
with transition rate $Q_t^d$.
Namely, suppose
\begin{equation}
    q^d_{t+\ve|t}(y^d|x^d)
    = \delta_{y^d,x^d}+\ve Q_t^d(y^d, x^d) + o(\ve)
    \label{eq:dimwise-transition-rate}
\end{equation}
for each $d=1,\ldots, D$ and $t<t+\ve$.
In this case,
we have
\begin{equation}
    q_{t+\ve|t}(\bm{y}|\bm{x})
    = \prod_{d=1}^D q_{t+\ve|t}^d(y^d|x^d)
    = \delta_{\bm{y},\bm{x}}
    + \ve
    \sum_{d=1}^D
    Q_t^d(y^d, x^d)\delta_{\bm{y}^\sm{d},\bm{x}^\sm{d}}
     + o(\ve)
    \label{eq:transition-expansion}
\end{equation}
by simply expanding the product,
where $\bm{x}^\sm{d}\in\S^{D-1}$ is given by omitting the $d$-th entry of $\bm{x}$.
From \eqref{eq:transition-expansion},
the transition rate for $\bm{x}_t$ is given by
\begin{equation}
    Q_t(\bm{y}, \bm{x}) =
    \sum_{d=1}^D
    Q_t^d(y^d, x^d)\delta_{\bm{y}^\sm{d},\bm{x}^\sm{d}}
    \label{eq:transition-dim-sum}
\end{equation}
as in \citet[Proposition~3]{campbell2022continuous}.
Let us assume continuity regarding
the forward process in each dimension:
\begin{assp}
    \label{assp:dimwise-conti}
    For each $d=1,\ldots, D$,
    there exists a function $Q_t^d:\S\times\S\to\R$ indexed by
    $t\in[0, T]$
    satisfying \eqref{eq:dimwise-transition-rate}.
    Moreover, for any fixed $x,y\in\S$,
    $q_{t|s}^d(y|x)$ is continuous in
    $\{(s, t)\in [0, T]^2\mid s\le t\}$
    and $Q_t^d(y,x)$ is continuous in $[0, T]$.
\end{assp}

This can be satisfied by broad range of forward diffusion designs,
including uniform, absorbing (masked),
and discretized Gaussian diffusion~\citep[see,][Section~E]{campbell2022continuous}.

Under Assumption~\ref{assp:dimwise-conti},
$q_{t|s}$ and $Q_t$ for the original process $\bm{x}_t$ are also continuous since $q_{t|s}(\bm{y}|\bm{x})=
\prod_{d=1}^Dq_{t|s}^d(y^d|x^d)$ and \eqref{eq:transition-dim-sum}.
Thus, we can apply the argument in Section~\ref{sec:kolmogorov-general}
to obtain Kolmogorov equations
\eqref{eq:forward-kolmogorov} \& \eqref{eq:backward-kolmogorov}.
Moreover, we can show a favorable property of
the time-reversal process.
This is just a re-formalization of a well-known fact
(e.g., \citet[Proposition~3]{campbell2022continuous} and \citet[Proposition~3.2]{sunscore}).

\begin{prop}\label{prop:time-reversal}
    Let $\S^D_{t,+}:=\{\bm{x}\in\S^D\mid q_t(\bm{x})>0\}$.
    Under Assumption~\ref{assp:dimwise-conti},
    there exists a function
    $R_t:\S^D\times\S^D_{t,+}\to\R$
    indexed by $t\in(0, T]$
    such that
    \samepage
    \begin{itemize}
        \item[(a)] we have $q_{t-\ve|t}(\bm{y}|\bm{x}) = \delta_{\bm{y},\bm{x}} +
        \ve R_t(\bm{y}, \bm{x}) + o(\ve)$
        for $\bm{y}\in\S^D$, $\bm{x}\in\S^D_{t,+}$
        and
        $\ve>0$ with $t-\ve\ge0$, and
        \samepage
        \item[(b)]
        $R_t(\bm{y}, \bm{x})$ can be nonzero
        only if $\bm{x}$ and $\bm{y}$
        coincide in at least $D-1$ entries.
    \end{itemize}
\end{prop}
We give its proof in Section~\ref{sec:proof-time-reversal}.
As one can see from the proof, the time-reversal transition rate $R_t$
is given concretely by $R_t(\bm{y},\bm{x})=Q_t(\bm{x},\bm{y})
q_t(\bm{y})/q_t(\bm{x})$
when $\bm{x}\ne\bm{y}$ and $q_t(\bm{x})>0$.
Note that the ratio $q_t(\bm{y})/q_t(\bm{x})$
is treated as a discrete counterpart of the score function \citep{sunscore,loudiscrete}.

Let us add another regularity assumption:
\begin{assp}\label{assp:Q-diff}
    For each $d=1,\ldots, D$ and $x,y\in\S$,
    $Q_t^d(y, x)$ is differentiable for $t\in(0, T)$
    and the derivative $\partial_tQ_t^d(y, x)$ can be continuously
    extended to $[0, T]$.
\end{assp}
Note that usual choices of $Q_t^d$ regarding $t$
including the time-homogeneous case $Q_t=Q$ and the noise scheduling
$Q_t=\beta(t)Q$ with a smooth $\beta$ \citep{campbell2022continuous,loudiscrete}
satisfy this assumption.
Finally,
under these two assumptions,
we can formalize Theorem~\ref{thm:composition} as follows.

\begin{thm}\label{thm:comp-formal}
    Suppose
    $(\bm{x}_t)_{0\le t \le T}$ satisfies Assumptions~\ref{assp:dimwise-conti} \& \ref{assp:Q-diff}.
    Let $p_{s|t}$ be a product model with the correct
    marginals, i.e.,
    $p_{s|t}(\bm{x}_s|\bm{x}_t)=\prod_{d=1}^Dq_{s|t}^d(x_s^d|\bm{x}_t)$ for $s<t$.
    Then, there exists a constant $C>0$ such that,
    given timesteps $t_i=iT/N$ for $i=0,\ldots,N$,
    we have
    \begin{equation}
        d_\mathrm{TV}\!\left(q_0, \E[\bm{x}_T\sim q_T]{
            p_{t_0|t_1}\circ p_{t_1|t_2}\circ\cdots\circ
            p_{t_{N-1}|t_N}
            (\cdot|\bm{x}_T)
        }\right)
        \le \frac{C}{N}.
        \label{eq:comp-formal-ineq}
    \end{equation}
    Furthermore,
    there exists an example of $(\bm{x}_t)_{0\le t\le T}$ satisfying $D=\lvert\S\rvert=2$
    and the same assumptions such that
    the left-hand side of \eqref{eq:comp-formal-ineq} is lower-bounded by $c/N$ with some constant $c>0$
    for sufficiently large $N$.
\end{thm}
This theorem basically says the min-max convergence rate of the analytical sampling is $1/N$.
We give the proof of the first half,
i.e., Eq.~\eqref{eq:comp-formal-ineq},
in Section~\ref{sec:proof-comp-formal}.
For the latter half,
we provide the concrete version
in Proposition~\ref{prop:lower-bound} in the following section.

\subsection{Lower bound of Theorem~\ref{thm:comp-formal}}\label{sec:lower-bound}
We shall provide an example that yields an $\Omega(1/N)$ error
between the analytical and true denoisers.
Our example even satisfies the following stronger assumption,
which is often used in theoretical analysis \citep[e.g.,][Assumption~1]{campbell2022continuous}:
\begin{assp}\label{assp:nonzero}
    For any $t\in[0, T]$ and $\bm{x}\in\S^D$,
    $q_t(\bm{x})>0$ holds.
\end{assp}

Consider $\S=\{a,b\}$ and $D=2$, where the state-space is given by $\X=\{aa,ab,ba,bb\}$
by omitting parentheses.
Consider the (forward) Markov process given by the initial distribution
$q_0 = (\delta_{aa}+\delta_{bb})/2$
and the dimension-wise
time-homogeneous
transition rate
$Q^d_t(y, x)=1/2 - \delta_{yx}$ for $d=1,2$ and $x,y\in\S$.
Under this setting, the forward transition probability
is continuous and
satisfies
$\partial_tq^d_{t|s}(\cdot|a) = Q^d_t q^d_{t|s}(\cdot|a)$ as a vector-valued differential equation,
so we have, for $t > s$,
\[
    \partial_t q^d_{t|s}(a|a) = -\frac12q^d_{t|s}(a|a) + \frac12q^d_{t|s}(b|a)
    = \frac12 - q_{t|s}^d(a|a).
\]
By solving this,
we obtain 
$q_{t|s}^d(a|a)=\frac12(1 + e^{-(t-s)})$
for $t\ge s$.
By symmetry, we generally have
\begin{equation}
    q_{t|s}^d(a|a)=q_{t|s}^d(b|b)=\frac12(1 + e^{-(t-s)}),
    \quad
    q_{t|s}^d(b|a)=q_{t|s}^d(a|b)=\frac12(1 - e^{-(t-s)})
    \label{eq:eg-start}
\end{equation}
This is a special case of uniform diffusion and clearly satisfies Assumptions~\ref{assp:dimwise-conti} \& \ref{assp:Q-diff}.
Although the singularity of $q_0$ violates Assumption~\ref{assp:nonzero} at time zero,
we can consider the time interval $[\delta, T]$ for some $\delta>0$
instead of $[0, T]$ to ensure $q_t>0$.
We will, however, work with the singular $q_0$ for simplicity of computations.
The following proposition gives the lower bound discussed in Theorem~\ref{thm:comp-formal}.
If necessary, we can replace $T$ with $T+\delta$ and consider $\bm{x}^\prime_t = \bm{x}_{t+\delta}$
to match the time intervals.
\begin{prop}\label{prop:lower-bound}
    Let $(\bm{x}_t)_{\delta\le t\le T}$ be the Markov process defined above
    and $p_{s|t}$ be the product model
    $p_{s|t}(\bm{x}_s|\bm{x}_t)=\prod_{d=1}^Dq_{s|t}^d(x_s^d|\bm{x}_t)$ for $s<t$.
    If we let $N\ge 2(T-\delta)/\delta$ be an integer
    and $t_i = \delta + i(T-\delta)/N$ for $i=0,\ldots,N$ be timesteps,
    then there is a constant $c>0$ such that
    \begin{equation}
        d_\mathrm{TV}\!\left(q_\delta, \E[\bm{x}_T\sim q_T]{
            p_{t_0|t_1}\circ p_{t_1|t_2}\circ\cdots\circ
            p_{t_{N-1}|t_N}
            (\cdot|\bm{x}_T)
        }\right)
        \ge \frac{c}{N}.
        \label{eq:comp-formal-lower-bound}
    \end{equation}
\end{prop}
The proof is given in Section~\ref{sec:proof-lower-bound}.

\section{Proofs}

\subsection{Proof of Theorem~\ref{thm:main}}
\label{sec:proof-main}
\begin{proof}
    For simplicity of notation, let $\tilde{p}^{\psi}_{t_n|T}$ be the denoiser given by the teacher with timesteps $t_n<t_{n+1}<\cdots<t_N$, i.e,
    \[
        \tilde{p}^{\psi}_{t_n|T} :=
        p^{\psi}_{t_n|t_{n+1}}
        \circ\cdots\circ p^{\psi}_{t_{N-1}|t_N},
    \]
    so that we have $r_{t_n} = \E[\bm{x}_T\sim r_T]{\tilde{p}^{\psi}_{t_n|T}(\cdot|\bm{x}_T)}$.
    Note that we can just set
    $\tilde{p}^{\psi}_{t_N|T}(\cdot|\bm{x}) = \tilde{p}^{\psi}_{T|T}(\cdot|\bm{x}) = \delta_{\bm{x}}$. 
    
    Also, let $p_{0,n} := \E[\bm{x}_T\sim r_T]{
        p^\theta_{0|t_n}\circ \tilde{p}^\psi_{t_n|T}(\cdot|\bm{x}_T)}$
    for $n=1,\ldots, N$,
    where $p_{0,N}$ is just given by $p_{0,N}= \E[\bm{x}_T\sim r_T]{p^\theta_{0|T}(\cdot|\bm{x}_T)}$.
    We first compare $p_{0,n}$ and $p_{0,n+1}$ with the consistency loss.
    
    For each $0<u<t\le T$,
    we have
    \begin{align*}
        \mathcal{L}_\mathrm{consis}
        (\theta;\psi,r_t,0,u,t)
        &= \E[\bm{x}_t\sim r_t]{D_\mathrm{KL}(
        p_{0|u}^\theta\circ p_{u|t}^\psi(\cdot|\bm{x}_t)
        \,\Vert\,
        p_{0|t}^\theta(\cdot|\bm{x}_t)
        )}
        \\
        &\ge D_\mathrm{KL}\!
        \left( \E[\bm{x}_t\sim r_t]{p_{0|u}^\theta\circ
        p_{u|t}^\psi(\cdot|\bm{x}_t)}
        \,\middle\Vert\,
        \E[\bm{x}_t\sim r_t]{p_{0|t}^\theta(\cdot|\bm{x}_t)}
        \right)
    \end{align*}
    from the convexity (Proposition~\ref{prop:kl}).
    If we let $u=t_n$ and $t=t_{n+1}$
    for some $1\le n<N$, we can see
    \begin{equation}
        \E[\bm{x}_t\sim r_t]{p_{0|u}^\theta\circ
        p_{u|t}^\psi(\cdot|\bm{x}_t)}
        = \E[\bm{x}_T]{p^\theta_{0|t_n}\circ p_{t_n|t_{n+1}}^\psi
        \circ \tilde{p}^\psi_{t_{n+1}|T}(\cdot|\bm{x}_T)}
        =  p_{0, n},
        \nonumber 
    \end{equation}
    and $\E[\bm{x}_t\sim r_t]{p_{0|t}^\theta(\cdot|\bm{x}_t)}
        = p_{0, n+1}$ hold.
    By using Pinsker's inequality (Proposition~\ref{prop:pinsker}), we have
    \begin{equation}
        d_\mathrm{TV}(p_{0,n}, p_{0,n+1})
        \le \frac1{\sqrt{2}}D_\mathrm{KL}(p_{0,n}\,\Vert\, p_{0,n+1})^{1/2}
        \le \frac1{\sqrt2}\mathcal{L}_\mathrm{consis}
        (\theta;\psi,r_{t_{n+1}},0,t_n,t_{n+1})^{1/2}.
        \label{eq:consis-bound}
    \end{equation}

    From a similar argument,
    we have
    \begin{align*}
        \mathcal{L}_\mathrm{distil}
        (\theta;\psi,r_{t_1},t_1)
        &= \E[\bm{x}_{t_1}\sim r_{t_1}]{D_\mathrm{KL}(
        p_{0|t_1}^\psi(\cdot|\bm{x}_{t_1})
        \,\Vert\,
        p_{0|t_1}^\theta(\cdot|\bm{x}_{t_1})
        )} \ge D_\mathrm{KL}(r_0\,\Vert\,p_{0,1}),
    \end{align*}
    and thus
    \begin{equation}
        d_\mathrm{TV}(r_0, p_{0,1})
        \le \frac1{\sqrt{2}}D_\mathrm{KL}(r_0\,\Vert\,p_{0,1})^{1/2}
        \le \frac1{\sqrt{2}}\mathcal{L}_\mathrm{distil}
        (\theta;\psi,r_{t_1},t_1)^{1/2}.
        \label{eq:distil-bound}
    \end{equation}

    By using the triangle inequality of the total variation distance,
    we obtain
    \[
        d_\mathrm{TV}(r_0, p_{0, N})
        \le d_\mathrm{TV}(r_0, p_{0, 1})
        + \sum_{n=1}^{N-1} d_\mathrm{TV}(p_{0,n}, p_{0,n+1}).
    \]
    Finally, applying Eqs.~\eqref{eq:consis-bound}
    and \eqref{eq:distil-bound}
    to its right-hand side yields the desired inequality.
\end{proof}

\subsection{Proof of Proposition~\ref{prop:tv-triangle}}
\label{sec:proof-tv-triangle}
\begin{proof}
    Let us first consider the case of $q_1=q_2$.
    Then, we have
    \begin{align}
        &d_\mathrm{TV}(\E[y\sim q_1]{p_1(\cdot|{y})}, \E[{y}\sim q_1]{p_2(\cdot|{y})})
        \nonumber\\
        &= \frac12\sum_x\left\lvert \sum_yp_1(x|y)q_1(y) - \sum_yp_2(x|y)q_1(y)\right\rvert
        = \frac12\sum_x\left\lvert \sum_y (p_1(x|y) - p_2(x|y))q_1(y)\right\rvert
        \nonumber\\
        &\le \frac12\sum_x\sum_y\left\lvert p_1(x|y) - p_2(x|y)\right\rvert q_1(y)
        =\E[y\sim q_1]{d_\mathrm{TV}(p_1(\cdot|y), p_2(\cdot|y))},
        \label{eq:tv-ineq-1}
    \end{align}
    where we have used $q_1\ge0$ in the inequality.
    On the other hand,
    if $p_1=p_2$, we have
    \begin{align}
        &d_\mathrm{TV}(\E[y\sim q_1]{p_2(\cdot|{y})}, \E[{y}\sim q_2]{p_2(\cdot|{y})})
        \nonumber\\
        &=\frac12\sum_x\left\lvert \sum_yp_2(x|y)q_1(y) - \sum_yp_2(x|y)q_2(y)\right\rvert
        =\frac12\sum_x \left\lvert \sum_y p_2(x|y) (q_1(y) - q_2(y)) \right\rvert
        \nonumber\\
        &\le \frac12\sum_x\sum_y  p_2(x|y)\lvert q_1(y) - q_2(y)\rvert
        = \frac12 \sum_y \lvert q_1(y) - q_2(y) \rvert
        = d_\mathrm{TV}(q_1,q_2),
        \label{eq:tv-ineq-2}
    \end{align}
    where we have used $p_2\ge0$ in the inequality and $\sum_xp_2(x|y)=1$
    in the last equality.

    By utilizing the usual triangle inequality of $d_\mathrm{TV}$
    and inequalities \eqref{eq:tv-ineq-1} \& \eqref{eq:tv-ineq-2},
    we obtain
    \begin{align*}
        &d_\mathrm{TV}(\E[y\sim q_1]{p_1(\cdot|{y})}, \E[{y}\sim q_2]{p_2(\cdot|{y})})\\
        &\le 
        d_\mathrm{TV}(\E[y\sim q_1]{p_1(\cdot|{y})}, \E[{y}\sim q_1]{p_2(\cdot|{y})})
        + d_\mathrm{TV}(\E[y\sim q_1]{p_2(\cdot|{y})}, \E[{y}\sim q_2]{p_2(\cdot|{y})})\\
        &\le
        \E[y\sim q_1]{d_\mathrm{TV}(p_1(\cdot|y), p_2(\cdot|y))}
        + d_\mathrm{TV}(q_1, q_2),
    \end{align*}
    which is the desired inequality.
\end{proof}

\subsection{Proof of Proposition~\ref{prop:time-reversal}}
\label{sec:proof-time-reversal}
\begin{proof}
    Note that, by Assumption~\ref{assp:dimwise-conti},
    $q_{t|s}$ is continuous over
    $\{(s,t)\in [0,T]^2\mid s\le t\}$,
    and $Q_t$ given by \eqref{eq:transition-dim-sum}
    is continuous over $[0, T]$
    and satisfies Eqs.~\eqref{eq:transition-forward}--\eqref{eq:backward-kolmogorov},
    as mentioned in Section~\ref{sec:kolmogorov-factorized}
    just after Assumption~\ref{assp:dimwise-conti}.
    
    Let us simply write $x\in\X$
    instead of the bold style $\bm{x}\in\S^D$
    in this paragraph.
    We consider only $x\in\X$ such that $q_t(x)>0$.
    We follow the argument in
    \citet[Section~B.2]{sunscore}.
    Let us consider the conditional probability
    (namely, the true denoiser)
    $q_{s|t}(\cdot|x)$ for $s\le t$,
    which is uniquely determined since $q_t(x)>0$.
    Then, we have
    \begin{align}
        \partial_s q_{s|t}(y|x)
        &=\partial_s \frac{q_s(y)q_{t|s}(x|y)}{q_t(x)}
        =\frac{(\partial_s q_s)(y) q_{t|s}(x|y) + q_s(y)(\partial_s q_{t|s})(x|y)}{q_t(x)}
        \nonumber
        \\
        &=\frac1{q_t(x)}
        \left(
            q_{t|s}(x|y)\sum_z Q_s(y,z)q_s(z)
            - q_s(y)\sum_w q_{t|s}(x|w)Q_s(w,y)
        \right),
        \label{eq:reversal-kolmogorov}
    \end{align}
    where we have used the forward Kolmogorov equation of $q_t$ given as
    \[
        \partial_tq_t(x)=\sum_w\partial_tq_{t|0}(x|w)q_0(w)
        =\sum_w\sum_zQ_t(x,z)q_{t|0}(z|w)q_0(w)
        =\sum_zQ_t(x, z)q_t(z)
    \]
    for computing $\partial_sq_s$
    and the backward Kolmogorov equation for computing $\partial_sq_{t|s}$.    
    By taking the limit $s\to t-0$ in \eqref{eq:reversal-kolmogorov},
    we obtain
    $\lim_{s\to t-0}\partial_sq_{s|t}(y|x) = - 
    \frac{q_t(y)}{q_t(x)}Q_t(x,y)$
    if $y\ne x$,
    given the continuity of $q_{t|s}$ and $Q_s$.
    Then, from Taylor's theorem,
    we obtain a backward counterpart
    of \eqref{eq:transition-forward} for $y\ne x$ as
    \begin{equation}
        q_{t-\ve|t}(y|x) =
        \ve \frac{q_t(y)}{q_t(x)}Q_t(x,y) + o(\ve),
        \qquad \ve>0.
    \end{equation}
    Since $\sum_{y}q_{t-\ve|t}(y|x)=1$ holds always,
    we also have that
    $q_{t-\ve|t}(x|x)=1 + \ve R_{t,x} + o(\ve)$
    for the coefficient $R_{t,x}=-\sum_{y\ne x}\frac{q_t(y)}{q_t(x)}Q_t(x,y)$.
    Therefore, we can prove (a)
    by letting $R_t(y, x)=\frac{q_t(y)}{q_t(x)}Q_t(x,y)$ for $y\ne x$
    and $R_t(x, x) = R_{t,x}$.

    We can see (b) from \eqref{eq:transition-dim-sum} and the concrete form of $R_t$.
\end{proof}

\subsection{Proof of first half of Theorem~\ref{thm:comp-formal}}
\label{sec:proof-comp-formal}
We first prove the following auxiliary lemma
replacing the $o(\ve)$ term in the backward transition by $O(\ve^2)$.
\begin{lem}\label{lem:lem-comp-formal}
    Under the same setting as in Theorem~\ref{thm:comp-formal},
    there is a constant $C>0$ such that,
    for any $t\in(0, T]$, $\ve\in(0, t]$, and $\bm{x}\in\X$ with $q_t(\bm{x})>0$, we have
    \begin{equation}
        d_\mathrm{TV}(q_{t-\ve|t}(\cdot|\bm{x}), p_{t-\ve|t}(\cdot|\bm{x}))
        \le \frac{C\ve^2}{q_t(\bm{x})}.
        \label{eq:lem-comp-formal}
    \end{equation}
\end{lem}
\begin{proof}
    From \eqref{eq:reversal-kolmogorov} and Assumption~\ref{assp:Q-diff},
    $q_{s|t}(\bm{y}|\bm{x})$
    for $s<t$ is twice-differentiable with regard to $s$,
    and $q_t(\bm{x})\partial_sq_{s|t}(\bm{y}|\bm{x})$
    can be represented as a polynomial of the function values of
    $q_s$, $Q_s$, $q_{t|s}$, and $\partial_sQ_s$.
    Thus, there is a constant $C_1$ depending on $\lvert \S\rvert$,
    $D$,
    $\sup_{s,\bm{z},\bm{w}}Q_s(\bm{z},\bm{w})$
    and $\sup_{s,\bm{z},\bm{w}}\partial_s(\bm{z},\bm{w})$
    such that
    $q_t(\bm{x})\partial_s^2 q_{s|t}(\bm{y}|\bm{x})\le C_1$ for any $s,t,\bm{y},\bm{x}$
    (note that $q_{t|s}$ and $q_s$ are within $[0, 1]$).

    Now that $\partial_sq_{s|t}$ can be continuously extended to $s\in [0, t]$
    from \eqref{eq:reversal-kolmogorov},
    for each $t\in (0, T]$, $\ve\in (0, t]$ and $\bm{x}, \bm{y}\in \S^D$
    with $q_t(\bm{x})>0$,
    Taylor's theorem yields that
    \begin{equation}
        \left\lvert q_{t-\ve|t}(\bm{y}|\bm{x}) - \delta_{\bm{y},\bm{x}}
            - \ve R_t(\bm{y},\bm{x})\right\rvert
        = \left\lvert \frac{(\partial_s^2q_{s|t})(\bm{y}|\bm{x})\vert_{s=\theta}}2\ve^2\right\rvert
        \le \frac{C_1}{2q_t(\bm{x})}\ve^2,
        \label{eq:second-order}
    \end{equation}
    for a certain $\theta\in(t-\ve, t)$.

    Let us next consider the marginal-matching product model $p_{t-\ve|t}$.
    For each $d$, if $y^d\ne x^d$, we have
    \begin{align}
        \left\lvert p_{t-\ve|t}^d(y^d|\bm{x}) - 
        \ve R_t((y^d, \bm{x}^\sm{d}), \bm{x})
        \right\rvert
        &= \left\lvert
            \sum_{\bm{y}^\sm{d}\in\S^{D-1}}q_{t-\ve|t}((y^d, \bm{y}^\sm{d})|\bm{x})
            - \ve R_t((y^d, \bm{x}^\sm{d}), \bm{x})
        \right\rvert \nonumber \\
        &=  \left\lvert
            \sum_{\bm{y}^\sm{d}}\left(q_{t-\ve|t}((y^d, \bm{y}^\sm{d})|\bm{x})
            - \ve R_t((y^d, \bm{y}^\sm{d}), \bm{x})\right)
        \right\rvert\nonumber\\
        &\le \frac{\lvert\S\rvert^{D-1}C_1}{2q_t(\bm{x})}\ve^2,
        \label{eq:marginal-second}
    \end{align}
    where the second equality comes from Proposition~\ref{prop:time-reversal}(b)
    and the inequality is from \eqref{eq:second-order}.
    If $y^d=x^d$, since $p_{t-\ve|t}^d(x^d|\bm{x})=1-\sum_{y^d\ne x^d}\lvert p_{t-\ve|t}^d(y^d|\bm{x})$
    we can use \eqref{eq:marginal-second} to obtain
    \begin{align}
        \left\lvert
        p_{t-\ve}^d(x^d|\bm{x}) - 1 + \ve \sum_{y^d\ne x^d}R_t((y^d, \bm{x}^\sm{d}), \bm{x})
        \right\rvert
        &\le \sum_{y^d\ne x^d}\lvert p_{t-\ve|t}^d(y^d|\bm{x}) - 
        \ve R_t((y^d, \bm{x}^\sm{d}), \bm{x})\rvert \nonumber \\
        &\le \frac{\lvert\S\rvert^DC_1}{2q_t(\bm{x})}\ve^2.
        \nonumber
    \end{align}
    From \eqref{eq:marginal-second} and this,
    by defining $R_t^d:\S\to\R$ as
    $R_t^d(y^d) = R_t((y^d, \bm{x}^\sm{d}), \bm{x})$ for $y^d\ne x^d$
    and $R_t^d(x^d)=-\sum_{y^d\ne x^d}R_t^d(y^d)$,
    there exists a constant $C_2>0$
    and a function $A^d:\S\to\R$ (for fixed $t$ and $\bm{x}$)
    such that
    \begin{equation}
        p_{t-\ve|t}^d(y^d|\bm{x}) = \delta_{y^d,x^d} - \ve R_t^d(y^d) + \frac{\ve^2}{q_t(\bm{x})} A^d(y^d, \ve),
       \qquad
       \sup_{y^d\in\S,\ \ve}\left\lvert A^d(y^d, \ve)\right\rvert \le C_2.
       \label{eq:marginal-summary}
    \end{equation}
    Therefore,
    we have
    \begin{align*}
        p_{t-\ve|t}(\bm{y}|\bm{x})
        &=\prod_{d=1}^D \left(\delta_{y^d,x^d} - \ve R_t^d(y^d) + \frac{\ve^2}{q_t(\bm{x})} A^d(y^d, \ve)\right)\\
        &=\delta_{\bm{y},\bm{x}} + \ve\sum_{d=1}^D R_t^d(y^d)\delta_{\bm{y}^\sm{d},\bm{x}^\sm{d}}
        + \underbrace{\sum_{k=1}^DP_k((\delta_{y^d,x^d},\ R_t^d(y^d),\ A^d(y^d, \ve))_{d=1}^D)
        \left(\frac{\ve^2}{q_t(\bm{x})}\right)^k}_{
            \text{Remainder term}
        },
    \end{align*}
    where $P_k$ is a certain polynomial of $3D$ variables for each $k$.
    Note that, if $\bm{y}\ne\bm{x}$,
    $R_t^d(y^d)\delta_{\bm{y}^\sm{d},\bm{x}^\sm{d}}$ can be
    nonzero only if $y^d\ne x^d$ and $\bm{y}^\sm{d}=\bm{x}^\sm{d}$.
    In that case, from the definition of $R_t^d(y^d)$,
    we have
    \begin{equation}
        p_{t-\ve|t}(\bm{y}|\bm{x})
        = \ve R_t^d(y^d) + (\text{Remainder term})
        = \ve R_t(\bm{y}, \bm{x}) + (\text{Remainder term})
        \label{eq:marg-eq-1}.
    \end{equation}
    This equality also holds when $\bm{y}$ and $\bm{x}$ differ in more than one entry,
    since the coefficient of $\ve$ becomes zero in such a case,
    and $R_t(\bm{y},\bm{x})=0$ from Proposition~\ref{prop:time-reversal}(b).
    Since the inputs for each $P_k$ are all bounded,
    we have
    \begin{equation}
        (\text{Remainder term})
        \le C_3 \sum_{k=1}^D\left(\frac{\ve^2}{q_t(\bm{x})}\right)^k
        \le C_3D\left(\frac{\ve^2}{q_t(\bm{x})} + \frac{\ve^{2D}}{q_t(\bm{x})^D}\right)
    \end{equation}
    for a constant $C_3>0$.
    By combining it with \eqref{eq:second-order},
    for $\bm{y}\ne\bm{x}$,
    we have
    \begin{align*}
        \lvert q_{t-\ve|t}(\bm{y}|\bm{x}) - p_{t-\ve|t}(\bm{y}|\bm{x})\rvert
        &\le \lvert q_{t-\ve|t}(\bm{y}|\bm{x})
            - \ve R_t(\bm{y},\bm{x}) \rvert + \lvert \ve R_t(\bm{y},\bm{x}) - p_{t-\ve|t}(\bm{y}|\bm{x}) \rvert \\
        &\le
        \frac{C_1}{2q_t(\bm{x})}\ve^2 + C_3D\left(\frac{\ve^2}{q_t(\bm{x})} + \frac{\ve^{2D}}{q_t(\bm{x})^D}\right)
        \le C_4 \left(\frac{\ve^2}{q_t(\bm{x})} + \frac{\ve^{2D}}{q_t(\bm{x})^D}\right)
    \end{align*}
    for a constant $C_4>0$.
    In particular,
    we have
    \begin{align*}
        &d_\mathrm{TV}(q_{t-\ve|t}(\cdot|\bm{x}), p_{t-\ve|t}(\cdot|\bm{x}))
        = \frac12\sum_{\bm{y}}\lvert q_{t-\ve|t}(\bm{y}|\bm{x}) - p_{t-\ve|t}(\bm{y}|\bm{x})\rvert
        \nonumber\\
        &= \frac12\left(\sum_{\bm{y}\ne\bm{x}}\lvert q_{t-\ve|t}(\bm{y}|\bm{x}) - p_{t-\ve|t}(\bm{y}|\bm{x})\rvert
        + \left\lvert 1 - \sum_{\bm{y}\ne\bm{x}} q_{t-\ve|t}(\bm{y}|\bm{x}) 
        - 1 + \sum_{\bm{y}\ne\bm{x}} p_{t-\ve|t}(\bm{y}|\bm{x})\right\rvert \right)
        \nonumber\\
        &\le \sum_{\bm{y}\ne\bm{x}}\lvert q_{t-\ve|t}(\bm{y}|\bm{x}) - p_{t-\ve|t}(\bm{y}|\bm{x})\rvert
        \le \lvert \S\rvert^DC_4 \left(\frac{\ve^2}{q_t(\bm{x})} + \frac{\ve^{2D}}{q_t(\bm{x})^D}\right)
        = C_5\left(\frac{\ve^2}{q_t(\bm{x})} + \frac{\ve^{2D}}{q_t(\bm{x})^D}\right),
    \end{align*}
    for a constant $C_5>0$.
    Now, we can assume that $C_5\ge1/2$, by adding a positive number if necessary.
    Since $d_\mathrm{TV}$ is bounded above by $1$ in general,
    we consider two cases:
    \begin{itemize}
        \item[(a)] If $\frac{\ve^2}{q_t(\bm{x})}\ge1$, we have
        $d_\mathrm{TV}(q_{t-\ve|t}(\cdot|\bm{x}), p_{t-\ve|t}(\cdot|\bm{x})) \le 1 \le 2C_5\frac{\ve^2}{q_t(\bm{x})}$
        since $2C_5\ge1$.
        \item[(b)] If $\frac{\ve^2}{q_t(\bm{x})}<1$, we have
        $d_\mathrm{TV}(q_{t-\ve|t}(\cdot|\bm{x}), p_{t-\ve|t}(\cdot|\bm{x}))
        \le C_5\left(\frac{\ve^2}{q_t(\bm{x})} + \frac{\ve^{2D}}{q_t(\bm{x})^D}\right)
        \le 2C_5\frac{\ve^2}{q_t(\bm{x})}$
        since $\frac{\ve^2}{q_t(\bm{x})} \ge \frac{\ve^{2D}}{q_t(\bm{x})^D}$.
    \end{itemize}
    Therefore, we finally obtain~\eqref{eq:lem-comp-formal}.
\end{proof}

By using the lemma and Proposition~\ref{prop:tv-triangle},
we can prove the theorem.
\begin{proof}[Proof of Theorem~\ref{thm:comp-formal}]
    For each $i=0,\ldots,N$,
    let us define the compositions
    \[
        \tilde{p}_{0|t_0}(\cdot|\bm{x})=\delta_{\bm{x}},
        \qquad
        \tilde{p}_{0|t_i}:=p_{t_0|t_1}\circ\cdots\circ p_{t_{i-1}|t_i},
        \quad i=1,\ldots,N.
    \]
    Note also that, for $\bm{x}$ with $q_T(\bm{x})>0$, we have
    $q_{t_i|T}(\cdot|\bm{x})=q_{t_i|t_{i+1}}\circ\cdots\circ q_{t_{N-1}|t_N}(\cdot|\bm{x})$
    from the Markov property of the reverse process.
    Indeed, for $s<t<u$, we have $q_{u|t}(\bm{z}|\bm{y})=q_{u|s,t}(\bm{z}|\bm{x},\bm{y})$
    from the Markov property of the forward process,
    so, for $\bm{z}$ with $q_u(\bm{z})>0$,
    \begin{align*}
        \sum_{\bm{y}}q_{s|t}(\bm{x}|\bm{y})q_{t|u}(\bm{y}|\bm{z})
        &= \sum_{\bm{y}}\frac{q_{s,t}(\bm{x},\bm{y})}{q_t(\bm{y})}
        \frac{q_{t,u}(\bm{y},\bm{z})}{q_u(\bm{z})}\\
        &= \sum_{\bm{y}}
        \frac{q_{s,t}(\bm{x},\bm{y})q_{u|t}(\bm{z}|\bm{y})}{q_u(\bm{z})}
        =\sum_{\bm{y}}
        \frac{q_{s,t}(\bm{x},\bm{y})q_{u|s,t}(\bm{z}|\bm{x},\bm{y})}{q_u(\bm{z})}\\
        &=\frac{\sum_{\bm{y}}q_{s,t,u}(\bm{x},\bm{y},\bm{z})}{q_u(\bm{z})}
        =\frac{q_{s,u}(\bm{x},\bm{z})}{q_u(\bm{z})}
        =q_{s|u}(\bm{x}|\bm{z}),
    \end{align*}
    where we have implicitly used that $q_t(\bm{y})>0$ holds for $\bm{y}$ satisfying
    $q_{t|u}(\bm{y}|\bm{z})>0$ (given $q_u(\bm{z})>0$).
    By using the inequality recursively, we can prove the aforementioned identity.

    We prove the desired estimate by exploiting the compositions.
    Recall $q_0=\E[\bm{x}_T\sim q_T]{q_{0|t_N}(\cdot|\bm{x}_T)}$.
    What we want to estimate is
    $d_\mathrm{TV}(\E[\bm{x}_T\sim q_T]{q_{0|t_N}(\cdot|\bm{x}_T)},
    \E[\bm{x}_T\sim q_T]{\tilde{p}_{0|t_N}(\cdot|\bm{x}_T)})$.
    We bound the distance with the following triangle inequality:
    \begin{align}
        &d_\mathrm{TV}(\E[\bm{x}_T\sim q_T]{q_{0|t_N}(\cdot|\bm{x}_T)},
        \E[\bm{x}_T\sim q_T]{\tilde{p}_{0|t_N}(\cdot|\bm{x}_T)})
        \nonumber\\
        &\le
        \sum_{i=0}^{N-1}
        d_\mathrm{TV}(\E[\bm{x}_T\sim q_T]{\tilde{p}_{0|t_i}\circ q_{t_i|t_N}(\cdot|\bm{x}_T)},
        \E[\bm{x}_T\sim q_T]{\tilde{p}_{0|t_{i+1}}\circ q_{t_{i+1}|t_N}(\cdot|\bm{x}_T)}).
        \label{eq:tv-triangle-final}
    \end{align}
    
    Let us bound each term inside the summation by using Lemma~\ref{lem:lem-comp-formal}
    and Proposition~\ref{prop:tv-triangle}.
    First, since $\tilde{p}_{0|t_{i+1}}=\tilde{p}_{0|t_i}\circ p_{t_i|t_{i+1}}$,
    by letting $p_1=p_2=\tilde{p}_{0|t_i}$ in Proposition~\ref{prop:tv-triangle},
    we have
    \begin{align}
        &d_\mathrm{TV}(\E[\bm{x}_T\sim q_T]{\tilde{p}_{0|t_i}\circ q_{t_i|t_N}(\cdot|\bm{x}_T)},
        \E[\bm{x}_T\sim q_T]{\tilde{p}_{0|t_{i+1}}\circ q_{t_{i+1}|t_N}(\cdot|\bm{x}_T)})
        \nonumber\\
        &\le d_\mathrm{TV}(\E[\bm{x}_T\sim q_T]{q_{t_i|t_N}(\cdot|\bm{x}_T)},
        \E[\bm{x}_T\sim q_T]{p_{t_i|t_{i+1}}\circ q_{t_{i+1}|t_N}(\cdot|\bm{x}_T)}).
    \end{align}
    Second, since $q_{t_{i+1}|t_N}=q_{t_i|t_{i+1}}\circ q_{t_{i+1}|t_N}$,
    by letting $q_1=q_2=q_{t_{i+1}}=\E[\bm{x}_T\sim q_T]{q_{t_{i+1}|t_N}(\cdot|\bm{x}_T)}$
    in Proposition~\ref{prop:tv-triangle} (note that the indices of $q_1,q_2$ here
    are different from time),
    we have
    \begin{align}
        &d_\mathrm{TV}(\E[\bm{x}_T\sim q_T]{q_{t_i|t_N}(\cdot|\bm{x}_T)},
        \E[\bm{x}_T\sim q_T]{p_{t_i|t_{i+1}}\circ q_{t_{i+1}|t_N}(\cdot|\bm{x}_T)})
        \nonumber\\
        &\le \E[\bm{x}\sim q_{t_{i+1}}]{d_\mathrm{TV}
        (q_{t_i|t_{i+1}}(\cdot|\bm{x}), p_{t_i|t_{i+1}}(\cdot|\bm{x}))}\nonumber\\
        &\le \sum_{q_{t_{i+1}}(\bm{x})>0}q_{t_{i+1}}(\bm{x})\cdot \frac{C(t_{i+1}-t_i)^2}{q_{t_{i+1}}(\bm{x})}
        \le \frac{C\lvert\S\rvert^DT^2}{N^2},
        \label{eq:tv-compose-concrete}
    \end{align}
    where we have used \eqref{eq:lem-comp-formal}
    and $t_{i+1}-t_i=T/N$ in the last inequality.
    By combining estimates \eqref{eq:tv-triangle-final}--\eqref{eq:tv-compose-concrete},
    we obtain
    \[
        d_\mathrm{TV}(\E[\bm{x}_T\sim q_T]{q_{0|t_N}(\cdot|\bm{x}_T)},
        \E[\bm{x}_T\sim q_T]{\tilde{p}_{0|t_N}(\cdot|\bm{x}_T)})
        \le \sum_{i=0}^{N-1}\frac{C\lvert\S\rvert^DT^2}{N^2}
        =\frac{C\lvert\S\rvert^DT^2}{N},
    \]
    which completes the proof with a replacement of the constant factor.
\end{proof}

\subsection{Proof of Proposition~\ref{prop:lower-bound}}
\label{sec:proof-lower-bound}
\begin{proof}
Consider the analytical sampler
$p_{s|t}(zw|xy)=q_{s|t}^1(z|x)q_{s|t}^2(w|y)$
for $s<t$.
Note that, because of the symmetry between $a$ and $b$ in $q_0$ and the forward transition,
the distributions $q_t$ or those given by the composition of $p_{s|t}$
are also symmetric.
Thus, the probability of $aa$ recovers all the information
of the distributions we consider over $\X$.

Let us compute several probabilities regarding $q_{s|t}$ and the analytical sampler
through \eqref{eq:eg-start}.
First, note that $q_{0|t}(ab|\cdot)=q_{0|t}(ba|\cdot)=0$.
Therefore, we have
\begin{align}
    q_{0|t}(aa|aa) &= \frac{q_{t|0}(aa|aa)q_0(aa)}{q_t(aa)}\nonumber\\
    &= \frac{q_{t|0}(aa|aa)q_0(aa)}{q_{t|0}(aa|aa)q_0(aa) + q_{t|0}(aa|bb)q_0(bb)}
    = \frac{\frac14(1+e^{-t})^2}{\frac14(1+e^{-t})^2 + \frac14(1-e^{-t})^2} = \frac{(1+e^{-t})^2}{2(1+e^{-2t})},
    \label{eq:aa|aa}\\
    q_{0|t}(bb|aa) &= 1-q_{0|t}(aa|aa) = \frac{(1-e^{-t})^2}{2(1+e^{-2t})},
    \label{eq:bb|aa}\\
    q_{0|t}(aa|ab) &= q_{0|t}(bb|ab) = \frac12,
    \label{eq:aa|ab}
\end{align}
where \eqref{eq:aa|ab} is derived from symmetry.

By using \eqref{eq:aa|aa}--\eqref{eq:aa|ab}
and the general fact (for Markov processes)
\begin{equation}
    q_{s|0,t}(\bm{x}_s|\bm{x}_0, \bm{x}_t)
    =\frac{q_{0,s,t}(\bm{x}_0, \bm{x}_s, \bm{x}_t)}{q_{0,t}(\bm{x}_0, \bm{x}_t)}
    =\frac{q_{s|0}(\bm{x}_s|\bm{x}_0)q_{t|0,s}(\bm{x}_t|\bm{x}_0,\bm{x}_s)}{q_{t|0}(\bm{x}_t|\bm{x}_0)}
    =\frac{q_{s|0}(\bm{x}_s|\bm{x}_0)q_{t|s}(\bm{x}_t|\bm{x}_s)}{q_{t|0}(\bm{x}_t|\bm{x}_0)}
    \label{eq:useful-fact}
\end{equation}
for $0\le s\le t$,
we can compute $q_{s|t}(\cdot|aa)$ for any $s\in[0, t]$ as follows:
\begin{align}
    q_{s|t}(aa|aa) &= q_{0|t}(aa|aa)q_{s|0,t}(aa|aa, aa) + q_{0|t}(bb|aa)q_{s|0,t}(aa|bb,aa)\nonumber\\
    &=\frac{(1+e^{-t})^2}{2(1+e^{-2t})}\frac{\frac14(1+e^{-s})^2\frac14(1+e^{-(t-s)})^2}{\frac14(1+e^{-t})^2}
    +\frac{(1-e^{-t})^2}{2(1+e^{-2t})}\frac{\frac14(1-e^{-s})^2\frac14(1+e^{-(t-s)})^2}{\frac14(1-e^{-t})^2}\nonumber\\
    &=\frac{((1+e^{-s})^2 + (1-e^{-s})^2)(1+e^{-(t-s)})^2}{8(1+e^{-2t})}
    =\frac{(1+e^{-2s})(1+e^{-(t-s)})^2}{4(1+e^{-2t})},\label{eq:st-aa|aa}\\
    q_{s|t}(bb|aa) &= q_{0|t}(aa|aa)q_{s|0,t}(bb|aa, aa) + q_{0|t}(bb|aa)q_{s|0,t}(bb|bb,aa)\nonumber\\
    &=\frac{(1+e^{-t})^2}{2(1+e^{-2t})}\frac{\frac14(1-e^{-s})^2\frac14(1-e^{-(t-s)})^2}{\frac14(1+e^{-t})^2}
    +\frac{(1-e^{-t})^2}{2(1+e^{-2t})}\frac{\frac14(1+e^{-s})^2\frac14(1-e^{-(t-s)})^2}{\frac14(1-e^{-t})^2}\nonumber\\
    &=\frac{((1-e^{-s})^2 + (1+e^{-s})^2)(1-e^{-(t-s)})^2}{8(1+e^{-2t})}
    =\frac{(1+e^{-2s})(1-e^{-(t-s)})^2}{4(1+e^{-2t})}\label{eq:st-bb|aa},\\
    q_{s|t}(ab|aa) &= q_{s|t}(ba|aa) = \frac12(1-q_{s|t}(aa|aa)-q_{s|t}(bb|aa)) \label{eq:st-aa+bb|aa}\\
    &=\frac12 - \frac{(1+e^{-2s})((1+e^{-(t-s)})^2+(1-e^{-(t-s)})^2)}{8(1+e^{-2t})}\nonumber\\
    &=\frac12 - \frac{(1+e^{-2s})(1+e^{-2(t-s)})}{4(1+e^{-2t})}
    =\frac14 - \frac{e^{-2s}+e^{-2(t-s)}}{4(1+e^{-2t})}.\label{eq:st-ab|aa}
\end{align}
We can also compute $q_{s|t}(aa|ab) = q_{s|t}(bb|ab)$ as
\begin{align}
    q_{s|t}(aa|ab) &= q_{0|t}(aa|ab)q_{s|0,t}(aa|aa, ab) + q_{0|t}(bb|ab)q_{s|0,t}(aa|bb,ab)\nonumber\\
    &=\frac12\frac{\frac14(1+e^{-s})^2\frac14(1+e^{-(t-s)})(1-e^{-(t-s)})}{\frac14(1+e^{-t})(1-e^{-t})}
    +\frac12\frac{\frac14(1-e^{-s})^2\frac14(1+e^{-(t-s)})(1-e^{-(t-s)})}{\frac14(1-e^{-t})(1+e^{-t})}\nonumber\\
    &=\frac{((1+e^{-s})^2+(1-e^{-s})^2)(1-e^{-2(t-s)})}{8(1-e^{-2t})}
    =\frac{(1+e^{-2s})(1-e^{-2(t-s)})}{4(1-e^{-2t})}\nonumber\\
    &=\frac14 + \frac{e^{-2s}-e^{-2(t-s)}}{4(1-e^{-2t})}.\label{eq:st-aa|ab}
\end{align}

Let us now compute the probabilities regarding the analytical sampler.
To make it simple, let $q_{s|t}(x*|\cdot):=q_{s|t}(xa|\cdot)+q_{s|t}(xb|\cdot)$
represent marginals; $q_{s|t}(*y|\cdot)$ is defined similarly.
By using this notation and \eqref{eq:st-aa|aa}--\eqref{eq:st-aa|ab},
we have
\begin{align}
    p_{s|t}(aa|aa)&=q_{s|t}(a*|aa)q_{s|t}(*a|aa) = q_{s|t}(a*|aa)^2 = (q_{s|t}(aa|aa) + q_{s|t}(ab|aa))^2\nonumber\\
    &=\left(\frac{(1+e^{-2s})(1+e^{-(t-s)})^2}{4(1+e^{-2t})} + \frac12 - \frac{(1+e^{-2s})(1+e^{-2(t-s)})}{4(1+e^{-2t})}\right)^2\nonumber\\
    &=\left(\frac{2(1+e^{-2t}) + (1+e^{-2s})((1+e^{-(t-s)})^2 - (1+e^{-2(t-s)}))}{4(1+e^{-2t})}\right)^2\nonumber\\
    &=\left(\frac{(1+e^{-2t}) + (1+e^{-2s})e^{-(t-s)}}{2(1+e^{-2t})}\right)^2
    =\left(\frac{(1+e^{-(t+s)})(1+e^{-(t-s)})}{2(1+e^{-2t})}\right)^2\label{eq:pst-aa|aa}\\
    p_{s|t}(bb|aa)&=q_{s|t}(b*|aa)q_{s|t}(*b|aa) = q_{s|t}(b*|aa)^2 = (q_{s|t}(bb|aa)+q_{s|t}(ba|aa))^2\nonumber\\
    &=\left(\frac{(1+e^{-2s})(1-e^{-(t-s)})^2}{4(1+e^{-2t})} + \frac12 - \frac{(1+e^{-2s})(1+e^{-2(t-s)})}{4(1+e^{-2t})}\right)^2\nonumber\\
    &=\left(\frac{2(1+e^{-2t}) + (1+e^{-2s})((1-e^{-(t-s)})^2 - (1+e^{-2(t-s)}))}{4(1+e^{-2t})}\right)^2\nonumber\\
    &=\left(\frac{(1+e^{-2t}) - (1+e^{-2s})e^{-(t-s)}}{2(1+e^{-2t})}\right)^2
    =\left(\frac{(1-e^{-(t+s)})(1-e^{-(t-s)})}{2(1+e^{-2t})}\right)^2\label{eq:pst-bb|aa}
\end{align}
Let us compute the sum of \eqref{eq:pst-aa|aa} and \eqref{eq:pst-bb|aa} as we use it later:
\begin{align}
    \icml{&}{}p_{s|t}(aa|aa) + p_{s|t}(bb|aa)\icml{\nonumber\\}{}
    &=\left(\frac{(1+e^{-(t+s)})(1+e^{-(t-s)})}{2(1+e^{-2t})}\right)^2
    + \left(\frac{(1-e^{-(t+s)})(1-e^{-(t-s)})}{2(1+e^{-2t})}\right)^2\nonumber\\
    &=\frac{((1+e^{-(t+s)})(1+e^{-(t-s)}))^2 + ((1-e^{-(t+s)})(1-e^{-(t-s)}))^2}{4(1+e^{-2t})^2}\nonumber\\
    &=\frac{(1+e^{-2t}+e^{-(t+s)}+e^{-(t-s)})^2 + (1+e^{-2t}-e^{-(t+s)}-e^{-(t-s)})^2}{4(1+e^{-2t})^2}\nonumber\\
    &=\frac{(1+e^{-2t})^2+(e^{-(t+s)}+e^{-(t-s)})^2}{2(1+e^{-2t})^2}
    =\frac12 + \frac{(e^{-(t+s)}+e^{-(t-s)})^2}{2(1+e^{-2t})^2}.
    \label{eq:pst-aa+bb|aa}
\end{align}
Next,
$p_{s|t}(aa|ab)$ is the product of two marginals --- $q_{s|t}(a*|ab)$ and $q_{s|t}(*a|ab)$,
which can be computed as follows:
\begin{align*}
    p_{s|t}(a*|ab)&=q_{0|t}(aa|ab)q^1_{s|0,t}(a|a,a) + q_{0|t}(bb|ab)q_{s|0,t}^1(a|b,a)\\
    &=\frac12\frac{\frac12(1+e^{-s})\frac12(1+e^{-(t-s)})}{\frac12(1+e^{-t})}
    +\frac12\frac{\frac12(1-e^{-s})\frac12(1+e^{-(t-s)})}{\frac12(1-e^{-t})}\\
    &=\frac{((1+e^{-s})(1-e^{-t})+(1-e^{-s})(1+e^{-t}))(1+e^{-(t-s)})}{4(1-e^{-2t})}\\
    &=\frac{(1-e^{-(t+s)})(1+e^{-(t-s)})}{2(1-e^{-2t})}
    =\frac12 + \frac{e^{-(t-s)} - e^{-(t+s)}}{2(1-e^{-2t})}, \\
    p_{s|t}(*a|ab)&= p_{s|t}(a*|ba) = p_{s|t}(b*|ab)= 1 - p_{s|t}(a*|ab)=\frac12 - \frac{e^{-(t-s)} - e^{-(t+s)}}{2(1-e^{-2t})},
\end{align*}
where the latter derivation is from the symmetries of the two dimensions and two characters.
By using these, we have
\begin{align}
    p_{s|t}(aa|ab)&=p_{s|t}(a*|ab)p_{s|t}(*a|ab)
    =\frac14 - \left(\frac{e^{-(t-s)} - e^{-(t+s)}}{2(1-e^{-2t})}\right)^2
    \label{eq:pst-aa|ab}.
\end{align}

Let us consider iteratively denoising from $q_T$ by using $p_{s|t}$.
For an $\ve>0$ and nonnegative integers $n\le T/\ve - 1$, define
\[
    p^\ve_T := p_T,\qquad
    p^\ve_{T-(n+1)\ve}:= \E[\bm{x}\sim p^\ve_{T-n\ve}]{p_{T-(n+1)\ve | T-n\ve}(\cdot|\bm{x})},
    \quad n=0,1,\ldots.
\]
Our goal is to estimate the difference between $p_{T-n\ve}^\ve$ and $q_{T-n\ve}$
for each $n$.
Let us fix $n$ and set $t=T-n\ve$ when computing $p^\ve_{t-\ve}$ in terms of $p^\ve_t$.
Because of the symmetry, $p_t^\ve(aa)=p_t^\ve(bb)$ and $p_t^\ve(ab)=p_t^\ve(ba) = \frac12 - p_t^\ve(aa)$ hold in general.
Therefore, by using \eqref{eq:pst-aa+bb|aa} and \eqref{eq:pst-aa|ab}, we have
\begin{align}
    p_{t-\ve}^\ve(aa) &= p_{t-\ve|t}(aa|aa)p_t^\ve(aa) + p_{t-\ve|t}(aa|bb)p_t^\ve(bb)
    + p_{t-\ve|t}(aa|ab)p_t^\ve(ab) + p_{t-\ve|t}(aa|ba)p_t^\ve(ba)\nonumber\\
    &= p_{t-\ve|t}(aa|aa)p_t^\ve(aa) + p_{t-\ve|t}(bb|aa)p_t^\ve(aa)
    + 2p_{t-\ve|t}(aa|ab)\left(\frac12 - p_t^\ve(aa)\right)\nonumber\\
    & = p_{t-\ve|t}(aa|ab) + (p_{t-\ve|t}(aa|aa) + p_{t-\ve|t}(bb|aa) - 2p_{t-\ve|t}(aa|ab))p_t^\ve(aa)\nonumber\\
    & = \frac14 - \frac{(e^{-\ve} - e^{-(2t-\ve)})^2}{4(1-e^{-2t})^2}
    + \left(
    \frac{(e^{-\ve}+e^{-(2t-\ve)})^2}{2(1+e^{-2t})^2}
    +
    \frac{(e^{-\ve} - e^{-(2t-\ve)})^2}{2(1-e^{-2t})^2}\right)p_t^\ve(aa).
    \label{eq:p-recc}
\end{align}
To compare it with $q_{t-\ve}$,
we also compute a similar recurrence equation by replacing $p$'s with $q$'s
and using
\eqref{eq:st-aa+bb|aa}--\eqref{eq:st-aa|ab}:
\begin{align}
    q_{t-\ve}(aa)
    & = q_{t-\ve|t}(aa|ab) + (q_{t-\ve|t}(aa|aa) + q_{t-\ve|t}(bb|aa) - 2q_{t-\ve|t}(aa|ab))q_t(aa)\nonumber\\
    &=\frac14 - \frac{e^{-2\ve}-e^{-2(t-\ve)}}{4(1-e^{-2t})} + \left(
       \frac{e^{-2\ve}+e^{-2(t-\ve)}}{2(1+e^{-2t})} +\frac{e^{-2\ve}-e^{-2(t-\ve)}}{2(1-e^{-2t})}\right)q_t(aa)
    \label{eq:q-recc}
\end{align}
Let us now compute quantities regarding the coefficients in \eqref{eq:p-recc} and \eqref{eq:q-recc}.
\begin{align}
    \icml{&}{}\frac{e^{-2\ve}-e^{-2(t-\ve)}}{1-e^{-2t}} - \frac{(e^{-\ve} - e^{-(2t-\ve)})^2}{(1-e^{-2t})^2}\icml{\nonumber\\}{}
    &= \frac{(e^{-2\ve}-e^{-2(t-\ve)})(1-e^{-2t}) - (e^{-\ve} - e^{-(2t-\ve)})^2}{(1-e^{-2t})^2}\nonumber\\
    &= \frac{(e^{-2\ve}-e^{-2(t-\ve)}-e^{-2(t+\ve)}+e^{-2(2t-\ve)}) - (e^{-\ve} - e^{-(2t-\ve)})^2}{(1-e^{-2t})^2}\nonumber\\
    &= - \frac{(e^{-(t-\ve)}-e^{-(t+\ve)})^2}{(1-e^{-2t})^2}
    = - \frac{e^{-2t}}{(1-e^{-2t})^2}(e^{\ve}-e^{-\ve})^2,
\end{align}
\begin{align}
    \icml{&}{}\frac{e^{-2\ve}+e^{-2(t-\ve)}}{1+e^{-2t}} - \frac{(e^{-\ve} + e^{-(2t-\ve)})^2}{(1+e^{-2t})^2}\icml{\nonumber\\}{}
    &= \frac{(e^{-2\ve}+e^{-2(t-\ve)})(1+e^{-2t}) - (e^{-\ve} + e^{-(2t-\ve)})^2}{(1+e^{-2t})^2}\nonumber\\
    &= \frac{(e^{-2\ve}+e^{-2(t-\ve)}+e^{-2(t+\ve)}+e^{-2(2t-\ve)}) - (e^{-\ve} + e^{-(2t-\ve)})^2}{(1+e^{-2t})^2}\nonumber\\
    &= \frac{(e^{-(t-\ve)}-e^{-(t+\ve)})^2}{(1+e^{-2t})^2}
    = \frac{e^{-2t}}{(1+e^{-2t})^2}(e^{\ve}-e^{-\ve})^2,
\end{align}
\begin{align}
    \icml{&}{}\frac{e^{-2\ve}+e^{-2(t-\ve)}}{1+e^{-2t}} +\frac{e^{-2\ve}-e^{-2(t-\ve)}}{1-e^{-2t}}\icml{\nonumber\\}{}
    &= \frac{(e^{-2\ve}+e^{-2(t-\ve)})(1-e^{-2t})+(e^{-2\ve}-e^{-2(t-\ve)})(1+e^{-2t})}{1-e^{-4t}}\nonumber\\
    &= 2+\frac{2(e^{-2\ve} - e^{-2(2t-\ve)}) - 2(1-e^{-4t})}{1-e^{-4t}}\nonumber\\
    &= 2+\frac{2(1+e^{2(2t-\ve)})}{1-e^{-4t}}(e^{-2\ve}-1).\label{eq:pq-diff-last}
\end{align}
We shall evaluate the difference $\Delta_{t}^\ve := q_{t}(aa) - p^\ve_{t}(aa)$ by using \eqref{eq:p-recc}--\eqref{eq:pq-diff-last}
as follows:
\begin{align}
    \Delta_{t-\ve}^\ve &=
    - \left(\frac{e^{-2\ve}-e^{-2(t-\ve)}}{4(1-e^{-2t})} - \frac{(e^{-\ve} - e^{-(2t-\ve)})^2}{4(1-e^{-2t})^2}\right)
    \nonumber\\
    &\quad + \left(
       \frac{e^{-2\ve}+e^{-2(t-\ve)}}{2(1+e^{-2t})} +\frac{e^{-2\ve}-e^{-2(t-\ve)}}{2(1-e^{-2t})}\right)(p_t^\ve(aa)+\Delta_t^\ve) \nonumber\\
    &\quad - \left(
    \frac{(e^{-\ve}+e^{-(2t-\ve)})^2}{2(1+e^{-2t})^2} 
        +
    \frac{(e^{-\ve} - e^{-(2t-\ve)})^2}{2(1-e^{-2t})^2}\right)p_t^\ve(aa) \nonumber\\
    &= \frac{e^{-2t}}{4(1-e^{-2t})^2}(e^{\ve}-e^{-\ve})^2 +
    \left(
       \frac{e^{-2\ve}+e^{-2(t-\ve)}}{2(1+e^{-2t})} +\frac{e^{-2\ve}-e^{-2(t-\ve)}}{2(1-e^{-2t})}\right)\Delta_t^\ve \nonumber\\
    &\quad+
    \left(
        \frac{e^{-2\ve}+e^{-2(t-\ve)}}{2(1+e^{-2t})}-\frac{(e^{-\ve}+e^{-(2t-\ve)})^2}{2(1+e^{-2t})^2}
        +\frac{e^{-2\ve}-e^{-2(t-\ve)}}{2(1-e^{-2t})}-\frac{(e^{-\ve} - e^{-(2t-\ve)})^2}{2(1-e^{-2t})^2}
    \right)p_t^\ve(aa)\nonumber\\
    &=\frac{e^{-2t}}{4(1-e^{-2t})^2}(e^{\ve}-e^{-\ve})^2
    + \left(1+\frac{1+e^{2(2t-\ve)}}{1-e^{-4t}}(e^{-2\ve}-1)\right)\Delta_t^\ve \nonumber\\
    &\quad + \left(\frac{e^{-2t}}{2(1+e^{-2t})^2} - \frac{e^{-2t}}{2(1-e^{-2t})^2} \right)(e^{\ve}-e^{-\ve})^2 p_t^\ve(aa) \nonumber\\
    &= \left(\frac{e^{-2t}}{2(1+e^{-2t})^2}p_t^\ve(aa) + \frac{e^{-2t}}{2(1-e^{-2t})^2}\left(\frac12-p_t^\ve(aa)\right) \right)(e^{\ve}-e^{-\ve})^2
    \nonumber\\
    &\quad + \left(1+\frac{1+e^{2(2t-\ve)}}{1-e^{-4t}}(e^{-2\ve}-1)\right)\Delta_t^\ve.
    \label{eq:delta-recc}
\end{align}
Since $p_t^\ve(aa)=p_t^\ve(bb)\le 1/2$, we have
\begin{align*}
    \icml{&}{}\frac{e^{-2t}}{2(1+e^{-2t})^2}p_t^\ve(aa) + \frac{e^{-2t}}{2(1-e^{-2t})^2}\left(\frac12-p_t^\ve(aa)\right)\icml{\\}{}
    &\ge \frac12\min\left\{ \frac{e^{-2t}}{2(1+e^{-2t})^2},\ \frac{e^{-2t}}{2(1-e^{-2t})^2} \right\}
    = \frac{e^{-2t}}{4(1-e^{-2t})^2}.
\end{align*}
Additionally, as the Taylor series of $(e^\ve - e^{-\ve})^2 = e^{2\ve} + e^{-2\ve} - 2$
is given by $\sum_{k=1}^\infty \frac2{(2k)!}(2\ve)^{2k}$,
we especially have $(e^\ve - e^{-\ve})^2 \ge 4\ve^2$.
Thus, we obtain
\begin{align}
    \icml{&}{}\left(\frac{e^{-2t}}{2(1+e^{-2t})^2}p_t^\ve(aa) + \frac{e^{-2t}}{2(1-e^{-2t})^2}\left(\frac12-p_t^\ve(aa)\right) \right)(e^{\ve}-e^{-\ve})^2
    \icml{\nonumber\\}{}
    &\ge \frac{e^{-2t}}{4(1-e^{-2t})^2}\cdot 4\ve^2 = \frac{e^{-2t}}{(1-e^{-2t})^2}\ve^2.
    \label{eq:coeff-1-0}
\end{align}
Also, since $e^{-2\ve}\ge 1 - 2\ve$,
we have
\begin{equation}
    1+\frac{1+e^{2(2t-\ve)}}{1-e^{-4t}}(e^{-2\ve}-1)
    \ge 1 - \frac{2(1+e^{2(2t-\ve)})}{1-e^{-4t}}\ve
    \ge 1 - \frac{4}{1-e^{-4t}}\ve.
\end{equation}
Suppose we are working on the time interval $[\delta, T]$ for some $\delta, T>0$.
Let us take $\ve \le \delta/2$; then we have
\begin{equation}
    1 - \frac{4}{1-e^{-4t}}\ve \ge 1 - \frac{4}{4t}\ve \ge 1 - \frac\ve\delta >0.
    \label{eq:coeff-2}
\end{equation}
For \eqref{eq:coeff-1-0},
we have
\begin{equation}
    \frac{e^{-2t}}{(1-e^{-2t})^2}\ve^2 \ge e^{-2t}\ve^2 \ge e^{-2T}\ve^2.
    \label{eq:coeff-1}
\end{equation}
By combining \eqref{eq:delta-recc}--\eqref{eq:coeff-1},
we first see that $\Delta_t^\ve$ is nonnegative for all $t = T - n\ve$ by induction on $n=0,1,\ldots$
(assuming $\ve\le\delta/2$ and $t\in[\delta, T]$).
Then, we obtain the following simple inequality:
\begin{equation*}
    \Delta_{t-\ve}^\ve \ge \left(1 - \frac\ve\delta\right)\Delta_t^\ve + e^{-2T}\ve^2
\end{equation*}
By recalling that $t = T-n\ve$,
we can rewrite it as
\begin{equation*}
    \left(1-\frac\ve\delta\right)^{-(n+1)}\Delta_{T-(n+1)\ve}^\ve \ge \left(1 - \frac\ve\delta\right)^{-n}\Delta_{T-n}^\ve
    +\left(1-\frac\ve\delta\right)^{-(n+1)}e^{-2T}\ve^2.
\end{equation*}
Since $\Delta_T^\ve = 0$, we have
\begin{align}
    \Delta_{T-n\ve}^\ve \ge
    \left(1-\frac\ve\delta\right)^n \sum_{k=1}^n \left(1-\frac\ve\delta\right)^{-k}e^{-2T}\ve^2
    =\sum_{k=0}^{n-1} \left(1-\frac\ve\delta\right)^{k}e^{-2T}\ve^2.
    \label{eq:lower-bound-sum}
\end{align}
Since $n\le T/\ve$
and $(1-1/x)^x$ is increasing over $x>1$, for $k=0,\ldots,n-1$, we have
\[
    \left(1-\frac\ve\delta\right)^{k}
    \ge \left(1-\frac\ve\delta\right)^{n}
    \ge \left(1-\frac\ve\delta\right)^{T/\ve}
    =\left(\left(1-\frac\ve\delta\right)^{\delta/\ve}\right)^{T/\delta}
    \ge \left(\left(1-\frac12\right)^2\right)^{T/\delta}
    = 2^{-2T/\delta},
\]
where we have exploited the assumption $\ve\le \delta/2$ (so that $\delta/\ve\ge2$).
By applying this to \eqref{eq:lower-bound-sum},
we obtain
\[
    \Delta_{T-n\ve}^\ve \ge (2^{1/\delta}e)^{-2T} n\ve^2.
\]
Now, let $\ve = (T-\delta)/N$ for the given $N$.
Since $N\ge \frac{2(T-\delta)}\delta$ and thus $\ve\le\delta/2$,
we have
\[
    \Delta_\delta^\ve = \Delta_{T-N\ve}^\ve \ge (2^{1/\delta}e)^{-2T} N\ve^2
    = (2^{1/\delta}e)^{-2T} \frac{(T-\delta)^2}N.
\]
Finally, as $d_\mathrm{TV}(q_\delta, p^{(T-\delta)/N}_\delta) \ge \Delta_\delta^\ve$,
the constant $c=(2^{1/\delta}e)^{-2T}(T-\delta)^2$
satisfies \eqref{eq:comp-formal-lower-bound}.
\end{proof}

\section{Control variates}\label{sec:cv}
When we want to compute an expectation $\E{f(\bm{x})}$,
instead of directly doing the Monte Carlo estimate
$\frac1N\sum_{i=1}^N f(\bm{x}_i)\approx \E{f(\bm{x})}$,
we can find a function $g\approx f$ such that $\E{g(\bm{x})}$ is tractable
and then do the Monte Carlo estimate for the remainder term:
\begin{equation}
    \frac1N\sum_{i=1}^N (f(\bm{x}_i) - g(\bm{x}_i)) + \E{g(\bm{x})}
    \approx \E{f(\bm{x})}.
    \label{eq:control-variate}
\end{equation}
This left-hand side is still an unbiased estimator of $\E{f(\bm{x})}$,
and ideally has a lower variance than the vanilla Monte Carlo estimator $\frac1N\sum_{i=1}^N f(\bm{x}_i)$
if $g\approx f$ is a good function approximation.
The role of $g$ in \eqref{eq:control-variate} is called a
{\it control variate} \citep{glasserman2004monte,oates2017control}.

\subsection{Marginal-matching product model
as control variate}\label{app:cv-explanation}
We briefly discuss how the product model $\overline{p}^\theta$
given in \eqref{eq:marginal-product}
satisfies the following favorable properties (already shown in Section~\ref{sec:consis-loss})
for being a control variate:
\begin{itemize}
    \item[(i)] it reasonably approximates $p_{s|t}^\theta(\cdot|\bm{x}_t)$, and
    \item[(ii)] $\E[\bm{x}\sim q]{g(\bm{x})}$ is easy to compute/approximate.
\end{itemize}

For point (i),
note that $\overline{p}^{\theta}$
is defined as a product model having the same marginal as $p^\theta$.
Since dimensionally independent modeling (when combined with multi-step sampling)
works as in Theorem~\ref{thm:composition},
$\overline{p}^\theta$ should approximate $p^\theta$
to a certain degree; see also Lemma~\ref{lem:lem-comp-formal} for a quantitative understanding.
The remainder $p^\theta - \overline{p}^\theta$ can then be regarded as
a dimensional correlation captured by $p^\theta$,
with which we conduct a usual Monte Carlo integration.

Regarding (ii), given a product distribution $\overline{p}(\bm{x})=\prod_{d=1}^D \overline{p}^d(x^d)$ over $\X=\S^D$,
we can indeed compute $H(q, \overline{p})$ with a Monte Carlo integral using samples of 
$\eta$ as
\begin{align}
    H(q,\overline{p})&=
    \E[\bm{x}_s\sim q]{-\log \overline{p}(\bm{x}_s)}
    =\mathbb{E}_{\eta}
    \E[\bm{x}_s\sim q^\eta]{
        -\log \overline{p}(\bm{x}_s)
    }
    \nonumber\\
    &
    =\E[\eta]{H(q^\eta, \overline{p})}
    =\E[\eta]{
        -\sum_{d=1}^D\sum_{x_s^d\in\S}q^\eta(x_s^d)\log \overline{p}^d(x_s^d)
    }.\label{eq:cv-computation}
\end{align}
While it still requires Monte Carlo with $\eta$ to estimate this,
it utilizes the product structure of each $q^\eta$ and $\overline{p}$
to exactly compute $H(q^\eta, \overline{p})$.
Thus, we heuristically expect it to be more accurate
than the Monte Carlo estimate using samples from $q$.

\subsection{Derivations of dimension-wise computable control variates
for mixture model}
\label{sec:cv-derivation}
\paragraph{Convex upper bound as control variate.}
To simplify the notation and situation,
suppose we are given probability distributions
$q = \E[\eta]{q^\eta}$ and $p^\theta = \E[\lambda]{p^{\theta, \lambda}}$,
where
$q^\eta$ and $p^{\theta,\lambda}$ are product distributions,
i.e.,
we have
\[
    q^{\eta}(\bm{x}) = \prod_{d=1}^D q^{\eta,d}(x^d),
    \qquad
    p^{\theta,\lambda}(\bm{x}) = \prod_{d=1}^D p^{\theta,\lambda,d}(x^d).
\]
By letting $H$ be the (cross) entropy,
we want to minimize
\[
    D_\mathrm{KL}(q \Vert p^\theta)
    = {H(q, p^\theta)} - H(q)
    = \E[\bm{x}\sim q]{-\log p^\theta(\bm{x})}
    -\E[\bm{x}\sim q]{-\log q(\bm{x})}.
\]
Since $q$ is fixed,
we simply want to minimize
\begin{align*}
     H(q, p^\theta) = \E[\bm{x}\sim q]{-\log p^\theta(\bm{x})}
     = \mathbb{E}_\eta\E[\bm{x}\sim q^\eta]{-\log p^\theta(\bm{x})}
\end{align*}
with regard to $\theta$.
However, it might have a high variance when we only sample
$\bm{x}\sim q$ and execute Monte Carlo.
One option is using the following upper bound like a negative ELBO
given by Jensen's inequality (convex inequality)
as a control variate:
\[
    -\log p^\theta(\bm{x})
    = -\log \E[\lambda]{p^{\theta,\lambda}(\bm{x})}
    \le \E[\lambda]{-\log p^{\theta, \lambda}(\bm{x})}.
\]
Indeed, its expectation regarding $\bm{x}\sim q$
is dimension-wise computable as
\begin{align*}
    \icml{&}{}\mathbb{E}_{\bm{x}\sim q}\E[\lambda]{-\log p^{\theta, \lambda}(\bm{x})}\icml{\\}{}
    &=\mathbb{E}_\eta\mathbb{E}_{\bm{x}\sim q^\eta}
    \E[\lambda]{-\log p^{\theta, \lambda}(\bm{x})}
    =\mathbb{E}_\eta\mathbb{E}_{\lambda}
    \E[\bm{x}\sim q^\eta]{-\log p^{\theta, \lambda}(\bm{x})}\\
    &= 
    \mathbb{E}_\eta\mathbb{E}_{\lambda}\sum_{d=1}^D\E[x^d\sim q^{\eta,d}]{
        -\log p^{\theta,\lambda, d}(x^d)
    }
    =\mathbb{E}_\eta\mathbb{E}_{\lambda}\!
    \left[-\sum_{d=1}^D
    \sum_{x^d}q^{\eta,d}(x^d)\log p^{\theta,\lambda, d}(x^d)
    \right],
\end{align*}
which does not require Monte Carlo sampling of $\bm{x}$.
Overall, we can decompose
the computation as
\begin{align*}
    \hspace{-3mm}
    H(q, p^\theta)
    =
    \underbrace{\E[\bm{x}\sim q]{-\log p^{\theta}(\bm{x})
        + \E[\lambda]{\log p^{\theta,\lambda}(\bm{x})}
    }}_{\text{Monte Carlo approximation}}
    +\underbrace{\mathbb{E}_{\bm{x}\sim q}\E[\lambda]{-\log p^{\theta,\lambda}(\bm{x})}}_{\text{dim-wise computable}}.
\end{align*}

\paragraph{Marginal control variate.}
The previous convex upper bound seems good, but
since
\[
    \mathbb{E}_{\bm{x}\sim q}\E[\lambda]{-\log p^{\theta,\lambda}(\bm{x})}
    = \E[\lambda]{H(q, p^{\theta,\lambda})}
    \ge \inf_\lambda H(q,p^{\theta,\lambda}),
\]
it might be a very loose bound
(we want the mixture to outperform the best product distribution $p^{\theta,\lambda}$).
To make it more practical,
we can consider its
dimension-wise tractable lower bound as follows:
\begin{align*}
    \mathbb{E}_{\bm{x}\sim q}\E[\lambda]{-\log p^{\theta,\lambda}(\bm{x})}
    & = \mathbb{E}_\eta\sum_{d=1}^D
    \mathbb{E}_{x^d\sim q^{\eta,d}}\E[\lambda]{
        -
        \log p^{\theta,\lambda, d}(x_d)}
    \ge -\mathbb{E}_\eta\sum_{d=1}^D
    \mathbb{E}_{x^d\sim q^{\eta,d}}\log \E[\lambda]{
        p^{\theta,\lambda, d}(x_d)},
\end{align*}
which is given by Jensen's inequality as well.
Therefore, if we define the product distribution
\[
    \overline{p}^{\theta}(\bm{x})
= \prod_{d=1}^D \overline{p}^{\theta,d}(x_d),
\qquad \overline{p}^{\theta,d}(x_d)
= \E[\lambda]{\overline{p}^{\theta,d}(x_d)},
\]
we have
$
    \mathbb{E}_{\bm{x}\sim q}\E[\lambda]{-\log p^{\theta,\lambda}(\bm{x})}
    \le \E[\bm{x}\sim q]{-\log \overline{p}^{\theta}(\bm{x})}
$
and this alternative is also dimension-wise computable.
Since $p^\theta$ and $\overline{p}^{\theta}$
coincide in each one-dimensional marginal, the difference between these two
can be regarded as the result of dimensional correlation.

Therefore, we propose the following decomposition,
which is also discussed in Section~\ref{sec:consis-loss}:
\[
    H(q, p^\theta) =
    \underbrace{\E[\bm{x}\sim q]{-\log p^{\theta}(\bm{x})
        + \log \overline{p}^{\theta}(\bm{x})
    }}_{\text{Monte Carlo approximation}}
    +\underbrace{\E[\bm{x}\sim q]{-\log \overline{p}^{\theta}(\bm{x})}}_{\text{dim-wise computable}}.
\]

\subsection{Product teacher model
as control variate}
\label{app:derivation-marginal}
For two models with the same marginals,
we have the following proposition:
\begin{prop}
    Let $q$, $\tilde{q}$ be probability distributions
    on $\X = \S^D$
    with the same marginals $q^d = \tilde{q}^d$.
    Then, for a product distribution $p(\bm{x}) = \prod_d p^d(x^d)$ over $\X$,
    we have $H(q,p) = H(\tilde{q}, p)$.
\end{prop}
\begin{proof}
    It suffices to prove that $H(q, p)$ can be computed only
    by using the marginals $q^d$.
    Indeed, we have
    \begin{align*}
        \E[\bm{x}\sim q]{\log p(\bm{x})}
        =\E[\bm{x}\sim q]{\sum_{d=1}^D \log p^d(x^d)}
        =\sum_{d=1}^D\E[\bm{x}\sim q]{\log p^d(x^d)}
        =\sum_{d=1}^D\sum_{x^d}q^d(x^d)\log p^d(x^d),
    \end{align*}
    and it yields the desired conclusion.
\end{proof}
From this proposition,
under $p_{0|t}^{\psi,d}\approx q_{0|t}^d$
and the fact that $\overline{p}^\theta$ is a product model,
we have
\[
    \E[\bm{x}_t\sim q_t]{H(q_{0|t}(\cdot|\bm{x}_t),
        \overline{p}_{0|t}^\theta(\cdot|\bm{x}_t))}
    \approx
    \E[\bm{x}_t\sim q_t]{H(p_{0|t}^\psi(\cdot|\bm{x}_t),
        \overline{p}_{0|t}^\theta(\cdot|\bm{x}_t))}.
\]
Since $H(p_1, p_2)=D_\mathrm{KL}(p_1\,\Vert\,p_2) + H(p_2, p_2)$
this right-hand side can be rewritten as
\[
    \E[\bm{x}_t\sim q_t]{H(p_{0|t}^\psi(\cdot|\bm{x}_t),
        \overline{p}_{0|t}^\theta(\cdot|\bm{x}_t))}
    = \E[\bm{x}_t\sim q_t]{
        D_\mathrm{KL}(p_{0|t}^\psi(\cdot|\bm{x}_t)\,\Vert\,
        \overline{p}_{0|t}^\theta(\cdot|\bm{x}_t)))
    } + const.,
\]
where the constant term is independent of $\theta$.
Since the KL divergence between two product distributions
decomposes into the sum of the KL divergence between each marginal,
we obtain approximation~\eqref{eq:cv-data}.

\section{Experimental details}
\label{sec:ex-details}

\subsection{Discretized Gaussian diffusions}\label{sec:expelimental-details}
\subsubsection{Sampling schemes}\label{sec:sampling}
In the experiments,
we used the following two sampling schemes
when evaluating the already trained product teacher model.
\paragraph{$\boldsymbol\tau$-leaping.}
In \citet{campbell2022continuous},
the authors first approximate the infinitesimal transition rate
by using each marginal $p^{\psi,d}_{0|t}$.
Indeed, the transition rate can be represented only with
$q_{0|t}^d$ and does not require a joint conditional distribution \citep[Proposition~3]{campbell2022continuous}.
After estimating the transition rate, they
apply a dimensionally parallel sampling method called $\tau$-leaping \citep{gillespie2001approximate} coming from computational chemistry.
Simply put,
$\tau$-leaping is a sort of generalization of the Euler method
for solving the backward SDE,
exploiting the ordinal structure of $\S$.
We omit the corrector steps; the $\tau$-leaping in Table~\ref{tab:experiment}
corresponds to $\tau$LDR-0 in \citet{campbell2022continuous}.

\paragraph{Analytical sampling.}
Although the $\tau$-leaping (or Euler method) is efficient with a large number of sampling steps,
we find that it deteriorates when we reduce the number of steps
seemingly due to discretization error.
Analytical sampling \citep{sunscore} (a.k.a. Tweedie $\tau$-leaping; \citealp{loudiscrete}),
which is simply a parallel exact sampling of each dimension
given as
\begin{equation}
    q_{s|t}^d(x_s^d|\bm{x}_t)
    = \sum_{x_0^d}q^d_{s|0,t}(x_s^d|x_0^d, x_t^d)
    q_{0|t}^d(x_0^d|\bm{x}_t)
    \approx \sum_{x_0^d}q^d_{s|0,t}(x_s^d|x_0^d, x_t^d)
    p_{0|t}^{\psi,d}(x_0^d|\bm{x}_t),
    \label{eq:dimiwise-analytical}
\end{equation}
does not suffer so much from the discretization.
This is also mentioned in \citet{gu2022vector}
as a fast inference strategy,
though they do not discuss dimensional correlations.
See also \eqref{eq:deduce-analytical} for the derivation
of a dimensionally independent denoiser based on the product model $p_{0|t}$.

Note that these schemes are both dimensionally independent
in the sense of \eqref{eq:product-model}
while not explicitly modeling $p_{s|t}$.
Indeed, the dimensional independence is ubiquitous even when modeling
$p_{s|t}$ implicitly.
First,
the reparametrization $p_{s|t}(\bm{x}_s|\bm{x}_t) = \sum_{\bm{x}_0}p_{0|t}(\bm{x}_0|\bm{x}_t)q_{s|0,t}(\bm{x}_s|\bm{x}_0,\bm{x}_t)$
    \citep{austin2021structured,gu2022vector},
    also used in analytical sampling,
    is dimensionally independent,
    provided that $p_{0|t}(\cdot|\bm{x}_t)$ is given by a product model
    and the forward diffusion is dimensionally independent.
Second, we can apparently avoid the heuristic in the above modeling through
    the estimation of the transition rate
    in the continuous-time discrete diffusion \citep[Proposition~3]{campbell2022continuous},
    but the existing sampling schemes of $\bm{x}_s$ given $\bm{x}_t$ in
    continuous-time settings
    including $\tau$-leaping \citep{campbell2022continuous} and the Euler-based method \citep{sunscore,loudiscrete}
    are still dimensionally independent.

$N$-step sampling in the actual experiment is given as follows.
We first set the timesteps $0=t_0<t_1 < \cdots < t_N=1$,
with $t_i=0.01 + 0.99\times \frac{i-1}{N-1}$ for $i\ge1$.
Given a terminal noise $\bm{x}_{t_N}$,
we sample $\bm{x}_{t_i}$ with our $p_{t_i|t_{i+1}}$
iteratively for $i=N-1,N-2,\ldots, 1$.
Finally,
we sample $\bm{x}_0\in\mathop\mathrm{argmax}p^\psi_{0|t_1}(\cdot|\bm{x}_{t_1})$
when using the teacher product model
and $\bm{x}_0\in\mathop\mathrm{argmax}p^\theta_{0|t_1}(\cdot|\bm{x}_{t_1};\lambda)$
with a random $\lambda$ when using the student mixture model.

\subsubsection{Additional experimental results}
To complement the main experimental results presented in Section~\ref{sec:tldr},
we provide additional details and analysis here.

\paragraph{FID/IS results of fewer sampling steps.}
For additional comparison,
we computed FID/IS of the teacher and student modles
using 10K samples (fewer than the 50K samples used in the main body, so the numbers may be slightly worse) for 2-20 steps in \Cref{tab:more-tldr}.
In terms of FID, our method achieves approximately 1.4 times acceleration in the 10–20 steps range.
However, it does not perform well in very few steps (e.g., 2–4 steps).
\begin{table}[h]
    \centering
    \caption{Comparison of models on CIFAR-10 dataset in various sampling steps.
        Same setting as \Cref{tab:tldr} except that 10,000 generated samples were used for computing FID/IS.}
    \label{tab:more-tldr}
    \vspace{2mm}
    \begin{tabular}{cccccccccccc}
        \toprule
         & \#steps \quad & 2 & 4 & 6 & 8 & 10 & 12 & 14 & 16 & 18 & 20  \\
        \midrule
        \multirow{2}{*}{FID} & teacher & 392.24 & 173.29 & 78.24 & 49.44 & 34.70 & 26.33 & 21.47 & 18.05 & 15.80 & 14.42 \\
        & student & 411.70 & 147.67 & 59.62 & 33.85 & 22.57 & 17.46 & 14.37 & 12.86 & 12.28 & 11.81 \\
        \midrule
        \multirow{2}{*}{IS} & teacher & 1.17 & 2.99 & 5.90 & 7.01 & 7.45 & 7.82 & 7.96 & 8.16 & 8.35 & 8.46 \\
        & student & 1.25 & 3.48 & 6.71 & 7.68 & 8.17 & 8.33 & 8.37 & 8.50 & 8.38 & 8.39 \\
        \bottomrule
    \end{tabular}
\end{table}

\paragraph{More detailed results of \Cref{tab:tldr}.}
We evaluated two different sampling strategies with the teacher model
$p^\psi$:
(1) $\tau$-leaping \citep{campbell2022continuous}
and (2) analytical sampling \citep{sunscore,loudiscrete}.
The complete evaluation results are shown in \Cref{tab:experiment},
which extends the results presented in Section~\ref{sec:tldr}.
A notable observation is that analytical sampling significantly outperforms
$\tau$-leaping in terms of sampling efficiency;
40-step analytical sampling
achieves better FID scores than 1000-step $\tau$-leaping.

\begin{table*}[!h]
    \centering
    \caption{Comparison of models on CIFAR-10 dataset.
    Fr\'echet inception distance (FID~$\downarrow$) against training dataset
    and inception score (IS~$\uparrow$)
    are calculated using 50,000 generated samples.
    $^*$: reported values from \citet{campbell2022continuous}.
    } 
    \label{tab:experiment}
    \vspace{2mm}
    \begin{NiceTabular}{ccccccccc} 
    \toprule 
    \Block{2-1}{Method}  & \multicolumn{2}{c}{10 steps}  & \multicolumn{2}{c}{20 steps}
    & \multicolumn{2}{c}{40 steps}
    & \multicolumn{2}{c}{1000 steps}\\ 
    \cmidrule(lr){2-3} \cmidrule(lr){4-5}
    \cmidrule(lr){6-7} \cmidrule(lr){8-9}
     & FID  & IS & FID  & IS & FID & IS & FID & IS \\
    \midrule
    $p^\psi$ + $\tau$-leaping
    & - & - & - & - 
    & 315.75 & 1.66\spm{0.01}
    & 8.10$^*$ & 8.74$^*$ \\
    $p^\psi$ + analytical
    & 32.61 & 7.59\scriptsize$\pm$0.10
    & 12.36 & 8.55\scriptsize$\pm$0.13
    & \bf 8.01 & \bf 8.77\scriptsize$\pm$0.09
    & - & - \\
    \midrule
    $p^\theta$ (student) & \bf 20.64 & \bf 8.29\spm{0.13} & 9.77 & 8.52\spm{0.08} & 9.66 & 8.28\spm{0.10} & - & -
    \\
    $p^\theta$\&$p^\psi$ (hybrid)
    & 25.54 & 8.00\spm{0.11} & \bf 9.47 & \bf 8.56\spm{0.14} & 8.02 & 8.43 \spm{0.11} & - & -
    \\
    \bottomrule
    \end{NiceTabular}
\end{table*}

Regarding distilled models,
as we have highlighted in Section~\ref{sec:tldr},
$p^\theta$ works well in 10 steps,
while it deteriorates as we grow the number of sampling steps.
The hybrid model interestingly beats other models in 20-step FID
and shows almost the same 40-step FID with the teacher,
while using the student solely gets worse in 40 steps.
We hypothesize (elaborating on the description in Section~\ref{sec:tldr}) that this is because
the true denoiser $q_{s|t}$ ($s<t$) becomes more ``dimensionally independent''
as $t-s$ or $t$ is small. The former condition (small $t-s$) explains the worse performance
gain of the mixture model as the number of steps grows,
and the latter partially explains the effectiveness of using the combined model.
However, we should further consider different
forward diffusion and/or
noise schedule to investigate it.

\subsubsection{Implementation and training}\label{sec:implementation}
\paragraph{Diffusion modeling.}
As explained in Section~\ref{sec:experiments},
the state-space has $D=3\times32\times32$ dimensions,
and each dimension has $256$ possibilities of pixel values which corresponds to $\S=\{0,\ldots,255\}$.
The forward diffusion process is defined through
a discretized Gaussian transition rate
with $T=1$ \citep[Section~E]{campbell2022continuous}.

\paragraph{Network architecture.}
All the models are based on the implementation explained in \citet[Section~H.2]{campbell2022continuous},
where $p_{0|t}^\psi$ is parameterized with a U-net \citep{ho2020denoising}
that has feature resolutions from $32\times32$ to $4\times4$.
Since the output of the original U-net architecture \citep{ho2020denoising} is a $D$-dimensional sequence (in $\S^D$)
rather than $D$ marginal distributions,
\citet{campbell2022continuous} adjusted the network so that
it first outputs a Gaussian distribution over the real line for each marginal
and then normalized it to obtain a distribution over $\S$.
The time $t$ in their implementation
is passed to feature map used in \citet{ho2020denoising},
and this embedding is fed to the upsampling/downsampling layers of the U-net
after passing through SiLU-activated linear layers \citep{elfwing2018sigmoid}.
See \citet[Section~H.2]{campbell2022continuous} and
their GitHub repository for more details on the original implementation.
All the models output the estimation of $q_{0|t}$,
and we conduct denoising from time $t$ to time $s$
by using the dimension-wise analytical sampling \eqref{eq:dimiwise-analytical},
except for the $\tau$-leaping benchmark in Table~\ref{tab:experiment}.

The only change we made on the architecture is
the insertion of $\lambda$.
We sample $\lambda$ from the uniform distribution over $[0, 1]$,
so we can basically use the same embedding architecture
as the time $t$.
For the downsampling layers,
the embedding of $\lambda$ is concatenated with the time embedding,
and then fed to the linear layers.
After the linear layers, similarly to the time embedding,
it is added to the latent vector of the image.
For the upsampling layers,
we concatenate the embeddings of $\lambda$, $t$,
and the pixel-wise average of the $4\times4$ resolution latent tensor,
and the remaining process is the same as for the downsampling layers.

\paragraph{Training.}
Since our model is an expansion of the original model for $p^\psi$,
we trained (finetuned) our student model $p^\theta$ from the checkpoint of $p^\psi$.
The bias terms and the final layers concerning the embeddings of $\lambda$ are zero-intialized,
and the rest are randomly intialized following the default setting of the original model.

For the Di4C finetuning, we followed the original setting
in terms of the use of the Adam optimizer and the learning rate $2\times10^{-4}$
as well as other hyperparameters.
The two primary differences in training are loss functions
and the training steps/minibatch size (due to the Monte Carlo for $\lambda$).
For the former point,
we basically used
\begin{equation}
    \L_\mathrm{distil}(\theta;\psi, q_\delta, \delta)
    +
    \L_\mathrm{consis}(\theta;\psi, q_t, 0, t-\Delta t, t)
    +
    \alpha_t\L_\mathrm{corr}(\theta; t)+ \L_\mathrm{marginal}(\theta; \psi,q_t,t),
    \label{eq:loss-cifar10}
\end{equation}
with techniques described in Section~\ref{sec:training-di4c}.
The following are additional details:
\begin{itemize}
    \item Sampling from $q_\delta$ and $q_t$ is based on the same sample of $\bm{x}_0\sim q_0$.
    \item $\delta=0.01$ with probability $1/2$; otherwise, $\delta$ is taken uniformly from
    $[0.01, 0.02]$.
    \item $\Delta t$ is sampled from a log-uniform distribution over $[0.001, 0.01]$;
    $t$ is then sampled uniformly from $[0.01+\Delta t,1]$.
    \item We can use several $\alpha_t$ as in the ablation study in the following section.
    In the main model $p^\theta$ given in Table~\ref{tab:experiment},
    we used the following sigmoid-based function as $\alpha_t$:
    \begin{equation}
        g(t) = \frac1{1+\exp(10-20t)}.
        \label{eq:sigmoid-alpha}
    \end{equation}
\end{itemize}
Regarding the training steps/minibatch details,
the original teacher model checkpoint had been trained for 2M steps,
where each step used 128 images from the CIFAR-10 dataset as a minibatch.
In our finetuning,
we stopped all the trainings in 320K steps (without warm-ups).
Each step used a minibatch of $128/L$ images from the CIFAR-10 dataset,
where $L$ is a batch size for $\lambda$ in the Monte Carlo estimates;
we set $M=N=L$ in \eqref{eq:monte-carlo}. 
$L=16$ is adopted in our model in Table~\ref{tab:experiment},
while the ablation study in the following section compares
various choices of $L$.

\paragraph{Evaluation.}
We measured FID and IS with the PyTorch-based implementation\footnote{\url{https://github.com/w86763777/pytorch-image-generation-metrics},
which got renamed from the original repository ``{\ttfamily pytorch-gan-metrics}'' to ``{\ttfamily pytorch-image-generation-metrics}''.}
following \citet{campbell2022continuous}.

\subsubsection{Ablation study}\label{sec:ablation}
\begin{table}[h]
    \centering
    \caption{Ablation study on $\alpha_t$ and use of control variates.
    } 
    \label{tab:alpha-ab}
    \vspace{2mm}
    \begin{NiceTabular}{ccccccc} 
    \toprule 
    \Block{2-1}{Method}  & \multicolumn{2}{c}{10 steps}  & \multicolumn{2}{c}{20 steps}
    & \multicolumn{2}{c}{40 steps}\\
    \cmidrule(lr){2-3} \cmidrule(lr){4-5}
    \cmidrule(lr){6-7} 
     & FID  & IS & FID  & IS & FID & IS \\ 
    \midrule
    $p^\psi$ + analytical
    & 32.61 & 7.59\scriptsize$\pm$0.10
    & 12.36 & 8.55\scriptsize$\pm$0.13
    & \bf 8.01 & \bf 8.77\scriptsize$\pm$0.09 \\
    \midrule
    $\alpha_t=0$ & 26.23 & 8.02\spm{0.09} & 11.55 & \bf8.59\scriptsize$\pm$0.07 & 9.01 & 8.65\spm{0.14} \\
    $\alpha_t=0$, w/o CV & 44.09 & 6.79\spm{0.10} & 26.16 & 7.54\spm{0.10} & 22.20 & 7.72\spm{0.08} \\
    $\alpha_t=1$ & 24.14 & 7.54\spm{0.08} & 12.30 & 8.06\spm{0.07} & 10.32 & 8.14\spm{0.10} \\
    $\alpha_t=1$, w/o CV & 26.92 & 8.12\spm{0.08} & 13.77 & 8.57\spm{0.14} & 10.59 & 8.66\spm{0.05}\\
    $\alpha_t=t$ & 24.21 & 8.10\spm{0.11} & 10.85 & 8.55\spm{0.08} & 9.27 & 8.51\spm{0.10} \\
    $\alpha_t=g(t)$ (see~\eqref{eq:sigmoid-alpha}) & \bf 22.77 & \bf8.19\spm{0.08} & \bf 10.07 & 8.54\spm{0.12} & 9.01 & 8.42\spm{0.11}
    \\
    \bottomrule
    \end{NiceTabular}
\end{table}
As an ablation study,
we compared several loss functions,
mainly changing $\alpha_t$,
which controls the degree of dimensional correlations we aim to learn from datapoints.
We also investigated whether the use of control variates is effective.
The results are shown in Table~\ref{tab:alpha-ab},
where
``w/o CV'' means that the control variates were not used in training.
The efficiency of control variates was consistent, while $\alpha_t=0$
and $\alpha_t=1$ had pros and cons.
Non-constant functions of $\alpha_t$ worked better,
partially matching the hypothesis discussed at the end of Section~\ref{sec:experiments}.
\begin{table}[h]
    \centering
    \caption{Ablation study on Monte Carlo sample size of $\lambda$.
    } 
    \label{tab:lam-ab}
    \vspace{2mm}
    \begin{NiceTabular}{ccccccc} 
    \toprule 
    \Block{2-1}{Method}  & \multicolumn{2}{c}{10 steps}  & \multicolumn{2}{c}{20 steps}
    & \multicolumn{2}{c}{40 steps}\\
    \cmidrule(lr){2-3} \cmidrule(lr){4-5}
    \cmidrule(lr){6-7} 
     & FID  & IS & FID  & IS & FID & IS \\ 
    \midrule
    $p^\psi$ + analytical
    & 32.61 & 7.59\scriptsize$\pm$0.10
    & 12.36 & 8.55\scriptsize$\pm$0.13
    & \bf8.01 & \bf8.77\scriptsize$\pm$0.09 \\
    \midrule
    $L=2$ & 27.29 & 8.00\spm{0.01} & 11.42 & \bf8.67\spm{0.12} & 8.94 & 8.64\spm{0.09} \\
    $L=4$ & 24.94 & 8.05\spm{0.14} & 10.66 & 8.60\spm{0.11} & 8.90 & 8.59\spm{0.07} \\
    $L=8$ & 22.77 & 8.19\spm{0.08} & 10.07 & 8.54\spm{0.12} & 9.01 & 8.42\spm{0.11} \\
    $L=16$ & 20.64 & \bf8.29\spm{0.13} & \bf9.77 & 8.52\spm{0.08} & 9.66 & 8.28\spm{0.10} \\ 
    $L=32$ & 20.25 & 8.28\spm{0.13} & 9.93 & 8.44\spm{0.10} & 9.91 & 8.26\spm{0.13} \\
    $L=64$ & \bf19.26 & 8.13\spm{0.10} & 10.13 & 8.26\spm{0.11} & 10.59 & 8.02\spm{0.15}\\
    \bottomrule
    \end{NiceTabular}
\end{table}

Additionally, we compared different batch-sizes of $\lambda$
in Table~\ref{tab:lam-ab}
(also see the end of the previous section).
The non-constant $\alpha_t=g(t)$ was used in all the settings. 
$L$ in the table represents the batch size of $\lambda$ in Monte Carlo sampling.
There is a certain tradeoff between FID and IS in 10- or 20-step sampling;
we can expect a better FID with a larger $L$ (smaller data batch),
while a smaller $L$ tends to result in a better IS.

\subsubsection{Generated samples}
\begin{figure}[!h]
    \centering
    \subfigure[\textbf{teacher}, $10$ steps]{
        \includegraphics[width=0.31\textwidth]{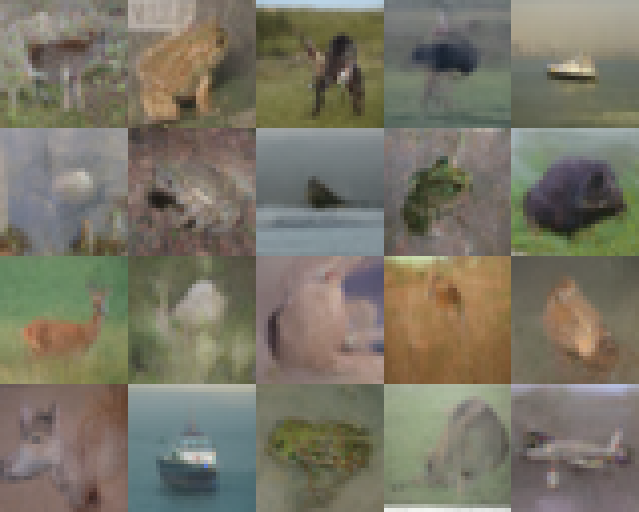}
    }
    \subfigure[\textbf{teacher}, $20$ steps]{
        \includegraphics[width=0.31\textwidth]{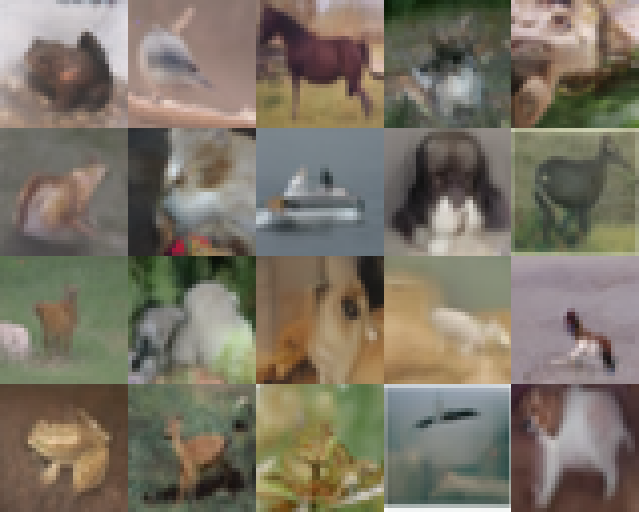}
    }
    \subfigure[\textbf{teacher}, $40$ steps]{
        \includegraphics[width=0.31\textwidth]{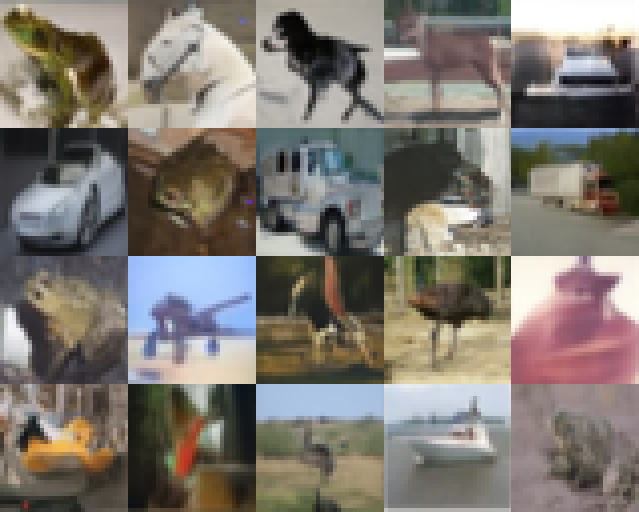}
    }
    \subfigure[\textbf{student}, $10$ steps]{
        \includegraphics[width=0.31\textwidth]{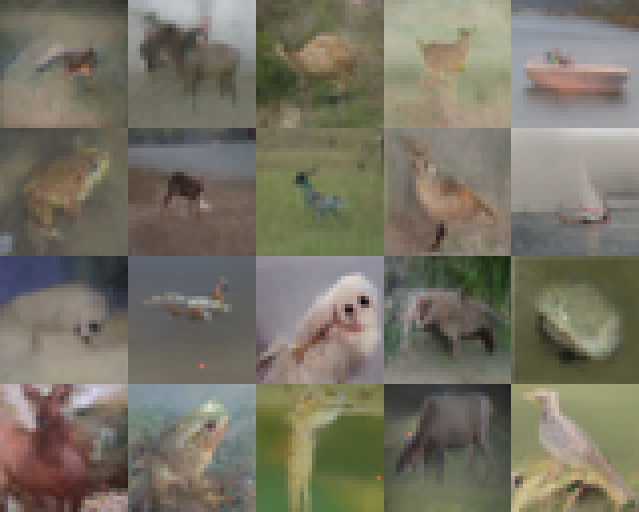}
    }
    \subfigure[\textbf{student}, $20$ steps]{
        \includegraphics[width=0.31\textwidth]{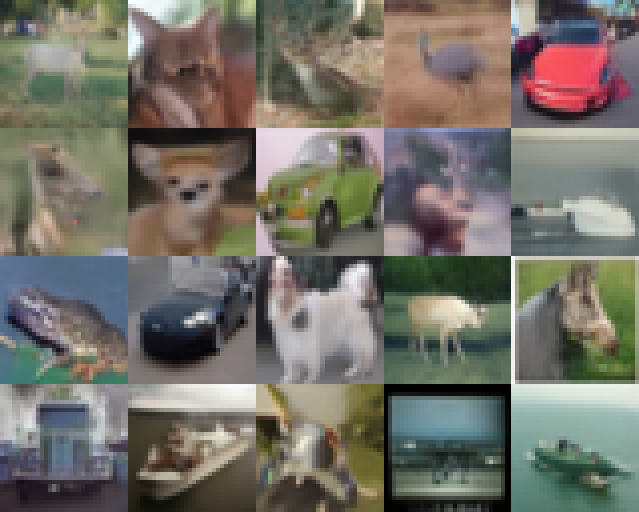}
    }
    \subfigure[\textbf{student}, $40$ steps]{
        \includegraphics[width=0.31\textwidth]{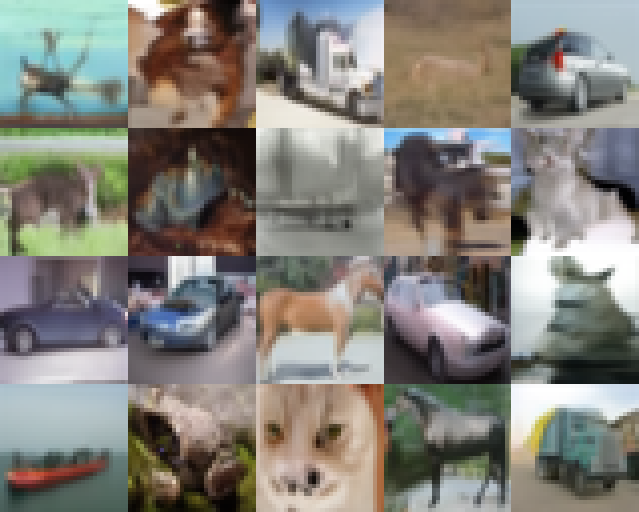}
    }
    \subfigure[\textbf{hybrid}, $10$ steps]{
        \includegraphics[width=0.31\textwidth]{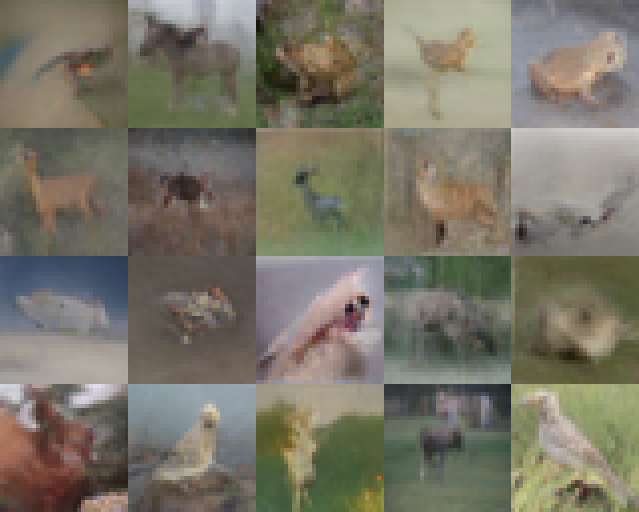}
    }
    \subfigure[\textbf{hybrid}, $20$ steps]{
        \includegraphics[width=0.31\textwidth]{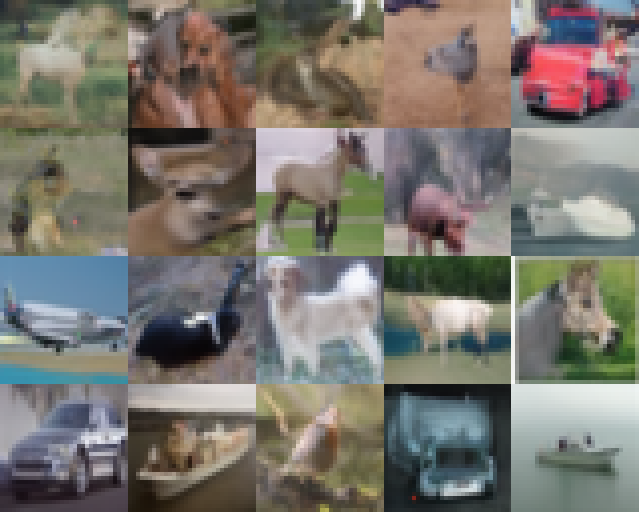}
    }
    \subfigure[\textbf{hybrid}, $40$ steps]{
        \includegraphics[width=0.31\textwidth]{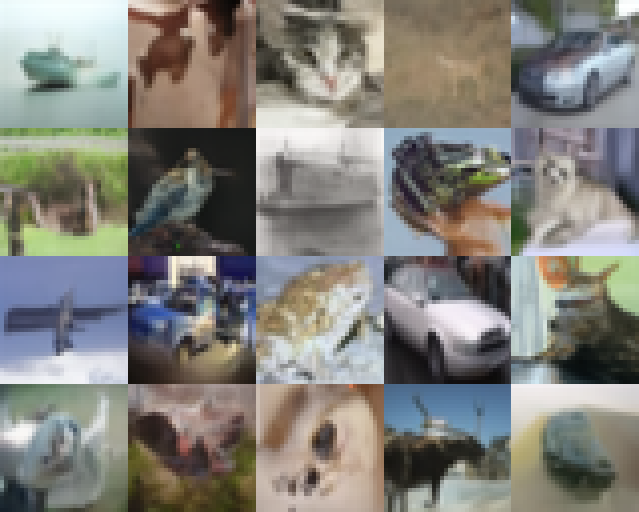}
    }
    \vspace{-2mm}
    \caption{Comparison of generated samples in CIFAR-10 experiment.}
    \label{fig:tldr-samples}
\end{figure}

\Cref{fig:tldr-samples} shows image examples corresponding to \Cref{tab:tldr},
which were all generated with the analytical sampling.

\subsection{Masked generative image modeling}\label{sec:mg}
\subsubsection{Masked diffusion modeling}
As described in \Cref{sec:ex-maskgit},
we use the pretrained VQGAN codebook $\S^*$ with $\lvert\S^*\rvert=1024$
and add one $\!\mask\!$ token to define $\S$.
Also, $D=256$ in this experiment.
The only ingredient we need in masked diffusion is the {\it masking probability} $m_t$:
a monotonically increasing function with $m_0=0$ and $m_1=1$.
Following \citet{chang2022maskgit} and \citet{besnier2023pytorch},
we model the forward process of $\bm{x}_t$
given $\bm{x}_0\in(\S^*)^D$ as
\begin{equation}
    x_t^d = \begin{cases}
        \!\mask\! & \text{with probability $m_t$},\\
        x_0^d & \text{with probability $1-m_t$},
    \end{cases}
    \label{eq:mask-forward}
\end{equation}
independently for each $d\in\{1,\ldots,D\}$.
Note that \eqref{eq:mask-forward} does not necessarily determine a Markov process
(and indeed an explicit Markov formulation is not needed for training).
If one needs a Markov formulation, however,
for $t>s>0$ and $x\in\S^*=\S\setminus\{\!\mask\!\}$, we have
\[
    m_t = q_{t|0}^d(\!\mask\!|x)=
    q_{s|0}^d(\!\mask\!|x) + q_{t|s}^d(\!\mask\!|x)q_{s|0}^d(x|x)
    = m_s + (1-m_s)q_{t|s}^d(\!\mask\!|x)
\]
and thus we have
\begin{equation}
    q_{t|s}^d(\!\mask\!|x) = \frac{m_t-m_s}{1-m_s}.
    \label{eq:masking-t|s}
\end{equation}
In the actual experiment,
we used the arccos scheduler
$m_t = 2\arccos(1-t)/\pi$.

\subsubsection{Confidence-based sampling}\label{sec:mg-sampling}
Given $\bm{x}_t$ at time $t$,
let $M_t:=\left\{d\in\{1,\ldots,D\}\mid x_t^d=\!\mask\!\right\}$.
Suppose we have a {\it product} model
$p_{0|t}(\cdot|\bm{x}_t) = \prod_{d=1}^Dp_{0|t}^d(\cdot|\bm{x}_t)$
such that
$p_{0|t}^d(\!\mask\!|\bm{x}_t)=0$ for all $d\in\{1,\ldots,D\}$
and $p_{0|t}^d(x_t^d|x_t^d)=1$ for all $d\not\in M_t$.
Let us explain how we sample $\bm{x}_s$ with $s<t$ in
one step of confidence-based sampling~\citep{chang2022maskgit,besnier2023pytorch}.
Following the original implementation of MaskGIT-PyTorch~\citep{besnier2023pytorch}
we conduct the sampling as follows:
\begin{enumerate}
    \item Sample $\tilde{\bm{x}}_0 = (\tilde{x}_0^d)_{d=1}^D \sim p_{0|t}(\cdot|\bm{x}_t)$.
        Note that we have $\tilde{x}_0^d\ne\!\mask\!$ for each $d$
        and $\tilde{x}_0^d=x_t^d$ for $d\not\in M_t$.
    \item Calculate the confidence for each chosen $\tilde{x}_0^d$ for $d\not\in M_t$ as
    \[
        \mathrm{conf}(d) = \log p_{0|t}^d(\tilde{x}_0^d|\bm{x}_t) + c_\mathrm{gb}(t)\cdot\ve^d_\mathrm{gb},
    \]
    where the second term is given by constant multiplication of Gumbel noise
    to add stochasticity in confidence-based sampling
    (see, e.g., \citet[Section~3.4.4]{comunita2024specmaskgit} for a concise explanation).
    To be concrete, $\ve^d_\mathrm{gb}$ for each $d\not\in M_t$ is an independent standard Gumbel noise,
    and $c_\mathrm{gb}$ is a scale factor given by $c_\mathrm{gb}(t) = \frac92\frac{t-1/N}{1-1/N}$
    in our experiments, where $N$ is the number of steps in the whole sampling process.
    \item Let $n(t,s)$ be the number of tokens we unmask in this single sampling step from $t$ to $s$.
        Let $d^*_{n(t,s)}\not\in M_t$ be the index with the $n(t,s)$-th largest $\mathrm{conf}(d)$.
        We define $\bm{x}_s=(x_s^d)_{d=1}^D$ as follows:
    \[
        x_s^d = \begin{cases}
            \tilde{x}_0^d & \text{for $d\not\in M_t$ with $\mathrm{conf}(d) \ge \mathrm{conf}(d^*_{n(t,s)})$},\\
            x_t^d & \text{otherwise.}
        \end{cases}
    \]
    Note that $\mathrm{conf}(d)$ coincides with probability zero thanks to the Gumbel noise.
\end{enumerate}
This is for one step of confidence-based denoising.
When we use a mixture model,
we first sample $\lambda$
and conduct the confidence-based sampling for the product model
conditioned by $\lambda$.
In the actual sampling process,
we specify $n$ (the number of unmasked tokens)
as we explain below\footnote{The definition of $n(t_i, t_{i-1})$ described below
is different from the implementation of MaskGIT-PyTorch,
since the original implementation did not correspond to forward processes.}.

Let $N$ be the number of sampling steps
and $t_i=i/N$ for $i=0,\ldots, N$.
We jump from $t_{i}$ to $t_{i-1}$ at each sampling step
for $i=N,N-1,\ldots,1$,
so we set $t=t_i$ and $s=t_{i-1}$.
Let $m_t$ be the masking probability at time $t$,
so that the expected number of $\!\mask\!$ tokens equals $Dm_t$
given the forward process design,
where $m_0 = 0$ and $m_1 = 1$.
By rounding them into integers,
we define $n(t_{i}, t_{i-1}) = \mathrm{round}(Dm_{t_i}) - \mathrm{round}(Dm_{t_{i-1}})$
for each $i$.
In the code, we further add $1$ when $n(t_i, t_{i-1})=0$ and deduct the added ones from $n(t_1, t_0)$
so that at least one token gets unmasked,
following the original implementation.

\subsubsection{Discrete classifier-free guidance}\label{sec:cfg}
Following the original implementation,
we use the discrete classifier-free guidance~\citep[discrete CFG;][]{tang2022improved}.
Given an unconditional denoiser $p_{0|t}(\cdot|\bm{x}_t)$ and
a conditional denoiser $p_{0|t}(\cdot|\bm{x}_t, c)$ with a class label $c$,
we sample from the distribution
\[
    p_{0|t}[w](\cdot|\bm{x}_t, c) \propto p_{0|t}(\cdot|\bm{x}_t,c)^{1+w} p_{0|t}(\cdot|\bm{x}_t)^{-w}
\]
for a CFG guidance scale $w$;
$w=0$ corresponds to sampling from the conditional denoiser.
In our implementation,
given the number of steps $N$,
we let $w = w_\mathrm{cfg}\cdot\frac{(1-t)N}{N-1}$
to linearly increase the guidance scale in the sampling process,
where $w_\mathrm{cfg}$ is a user-selected CFG coefficient,
mentioned in Section~\ref{sec:ex-maskgit}.
When we use a mixture model,
we first sample a single $\lambda$ and sample from the CFG-guided distribution
given by the $\lambda$-conditioned unconditional/conditional denoisers $p_{0|t}(\cdot|\bm{x}_t; \lambda)$
and $p_{0|t}(\cdot|\bm{x}_t, c; \lambda)$ as a heuristic.

\subsubsection{Implementation and training}\label{sec:mg-training}
\paragraph{Network architecture.}
For the teacher model $p^\psi$,
we just used the implementation of \citet[Table~1]{besnier2023pytorch},
which uses a $24$-layer transformer with $16$ attention heads
to compute the logits for each of $D=256$ visual tokens.
As inputs,
$p^\psi$ gets the $D$ visual tokens $\bm{x}_t=(x_t^d)_{d=1}^D$
and a class token $c$ (when unconditional, it is replaced by an ``unconditional token''),
and these $D+1$ tokens are added positional encodings.
Note that time conditioning is not fed to the model.
Each learned embedding (of 1024 VQGAN codebook elements and 1000 ImageNet classes)
is of 768 dimensions.
Since the output of the transformer has $D+1$ vectors in $\R^{768}$,
we discard the vector corresponding to $c$
and compute the logits by using the similarity with the embeddings of codebook elements.

To realize a mixture model $p^\theta(\cdot|\bm{x}_t;\lambda)$ upon this implementation,
we first sample $\lambda\sim\mathrm{Unif}([0,1])$,
pass it to the timestep embedding used in the CIFAR-10 experiment (\Cref{sec:implementation}),
and transform its output into a $768$-dimensional vector using a two-layer MLP with
a single GELU activation after the first layer.
Thus, we obtain $D+2$ vectors in $\R^{768}$ consisting of
$D$ visual embeddings, a class embedding, and an embedding of $\lambda$.
We simply feed them to a transformer having the same architecture as $p^\psi$
and discard the final output vectors concerning $c$ and $\lambda$ to calculate the logits.
The logit calculation is done in the same way as the teacher model.

\paragraph{Training.}
We basically used the same formulation as \eqref{eq:loss-cifar10}
with a slight modification to reduce the computational burden
as follows:
\begin{equation}
    \frac{\bm{1}_{\{t\le \Delta t\}}}{\Delta t}\L_\mathrm{distil}(\theta;\psi, q_t, t)
    +
    \bm{1}_{\{t>\Delta t\}}\L_\mathrm{consis}(\theta;\psi, q_t, 0, t-\Delta t, t)
    +
    \alpha_t\L_\mathrm{corr}(\theta; t)+ \L_\mathrm{marginal}(\theta; \psi,q_t,t),
    \label{eq:loss-maskgit}
\end{equation}
with $\Delta t = 0.05$ and the following details:
\begin{itemize}
    \item We sampled $t \sim \mathrm{Unif}([0,1])$
        and $\bm{x}_t$ according to \eqref{eq:mask-forward} using $\bm{x}_0$ from data.
    \item We used $\alpha_t = 0$ for \textbf{di4c} and
        $\alpha_t = 0.1\cdot g(t)$ for \textbf{di4c-d} in the experiment (\Cref{fig:maskgit}),
        where $g(t)$ is given in \eqref{eq:sigmoid-alpha}.
    \item Control variates were used in both variants of Di4C.
\end{itemize}

For optimization,
we followed the original implementation,
i.e., the AdamW optimizer with a learning rate~$10^{-5}$, $(\beta_1, \beta_2)=(0.9, 0.96)$
and a weight decay $10^{-5}$.
The teacher model was trained for 300 epochs with a minibatch size of $512$
using eight A100 GPUs (adding up to $768$ GPU hours; \citealp{besnier2023pytorch}).
Our finetuning used two A6000 GPUs with a minibatch size $4$ ($2$ for each of two GPUs)
and a $\lambda$-batch size of $32$.
It was trained for 30K iterations (so only 120K out of 1.28M ImageNet training images were used),
which amounts to approximately $50$ GPU hours.
\paragraph{Evaluation.}
In each experiment,
we generated 50,000 samples (50 images for each ImageNet class)
and then computed FID (against test data) and IS
with the original implementation of \citet{besnier2023pytorch}.

\subsubsection{Generated samples}

\begin{figure}
    \centering
    \subfigure[\textbf{teacher}, $8$ steps]{
        \includegraphics[width=0.4\textwidth]{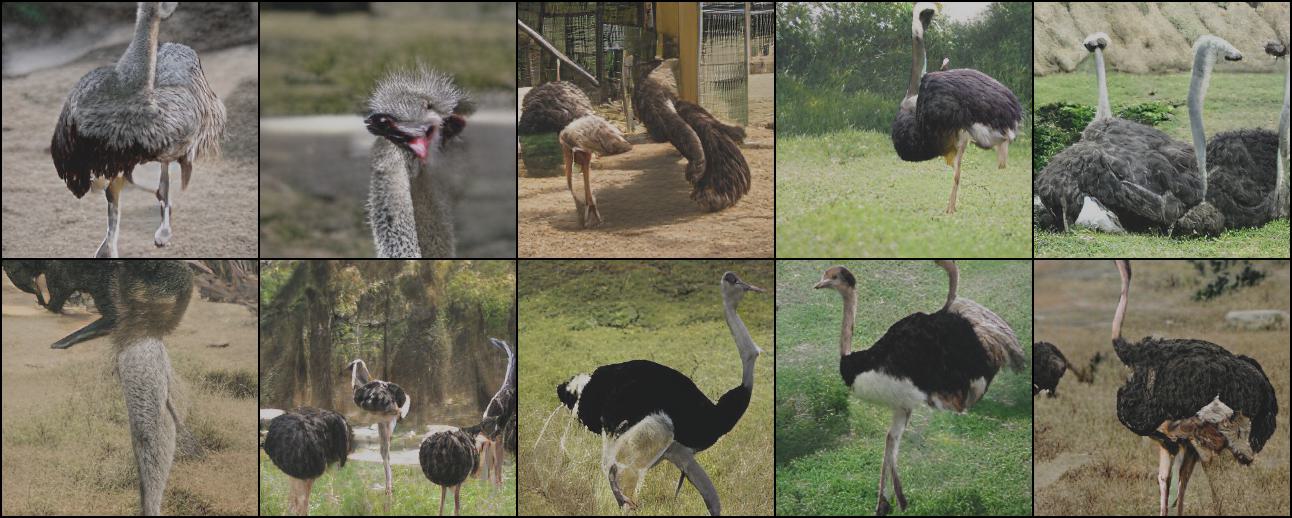}
    }
    \subfigure[\textbf{teacher}, $4$ steps]{
        \includegraphics[width=0.4\textwidth]{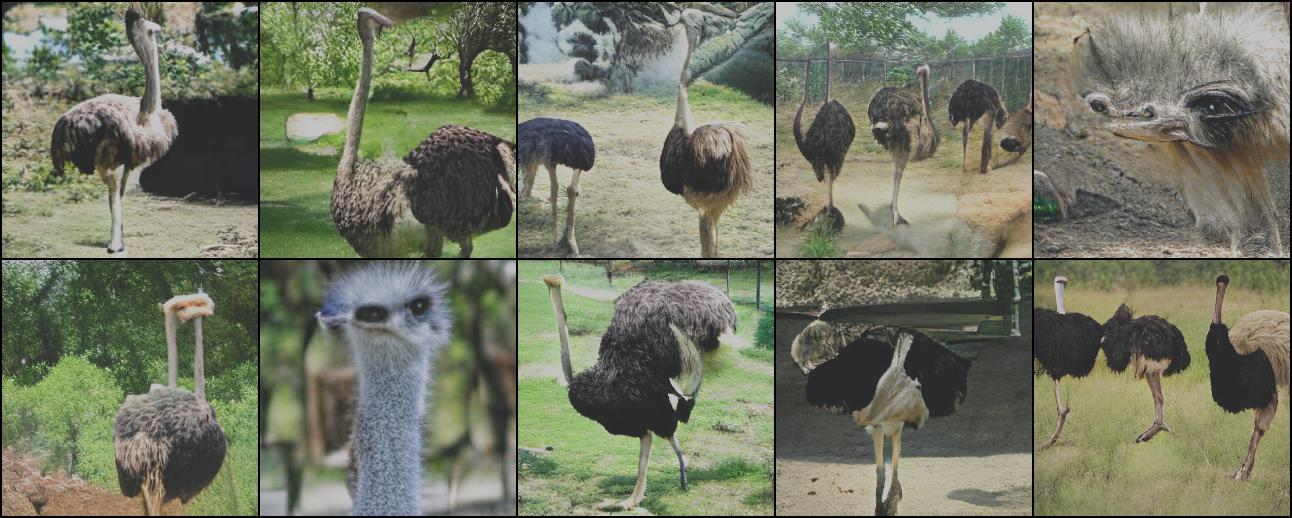}
    }
    \subfigure[\textbf{di4c}, $4$ steps]{
        \includegraphics[width=0.4\textwidth]{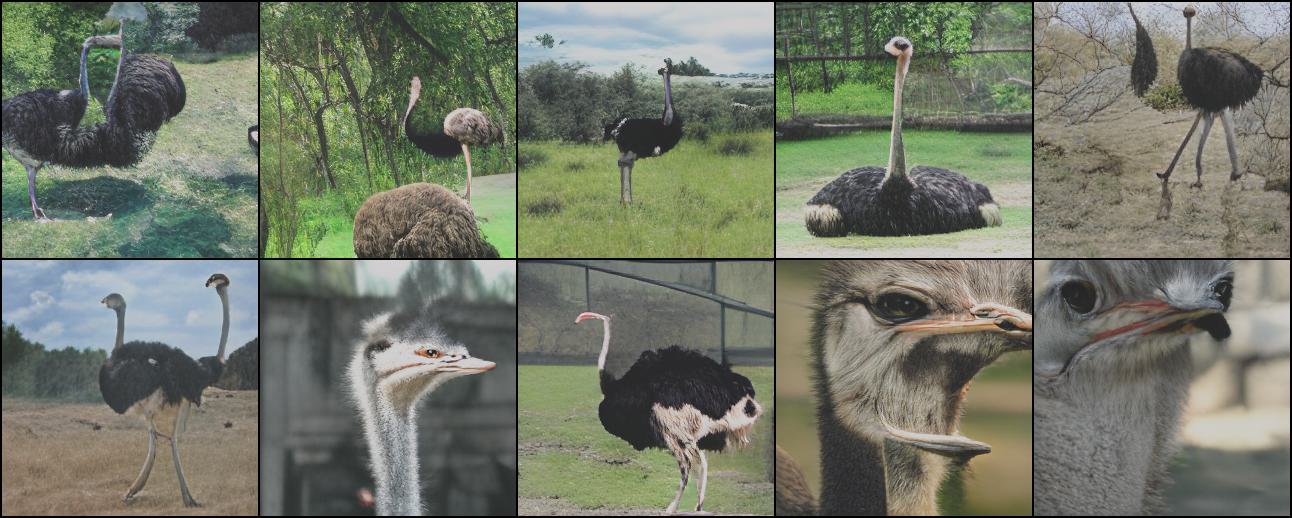}
    }
    \subfigure[\textbf{di4c-d}, $4$ steps]{
        \includegraphics[width=0.4\textwidth]{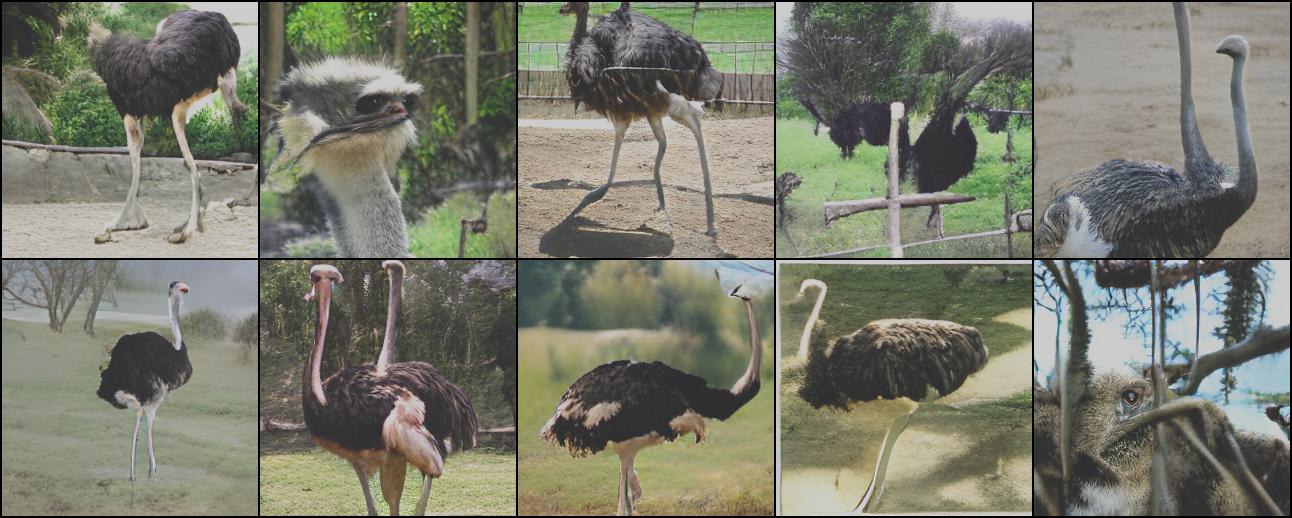}
    }
    \subfigure[\textbf{teacher}, $8$ steps]{
        \includegraphics[width=0.4\textwidth]{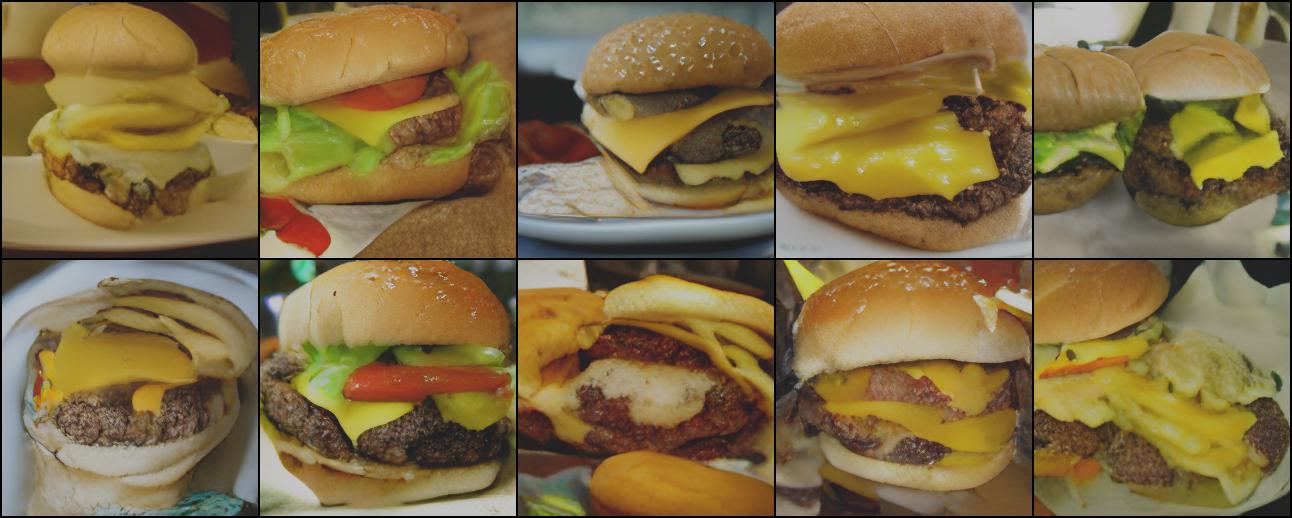}
    }
    \subfigure[\textbf{teacher}, $4$ steps]{
        \includegraphics[width=0.4\textwidth]{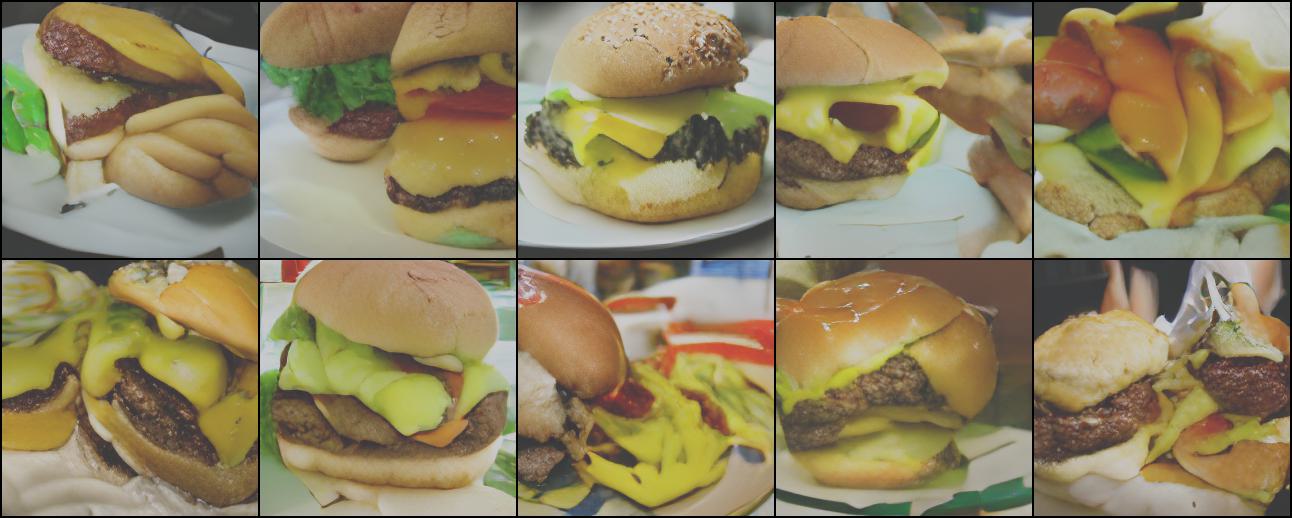}
    }
    \subfigure[\textbf{di4c}, $4$ steps]{
        \includegraphics[width=0.4\textwidth]{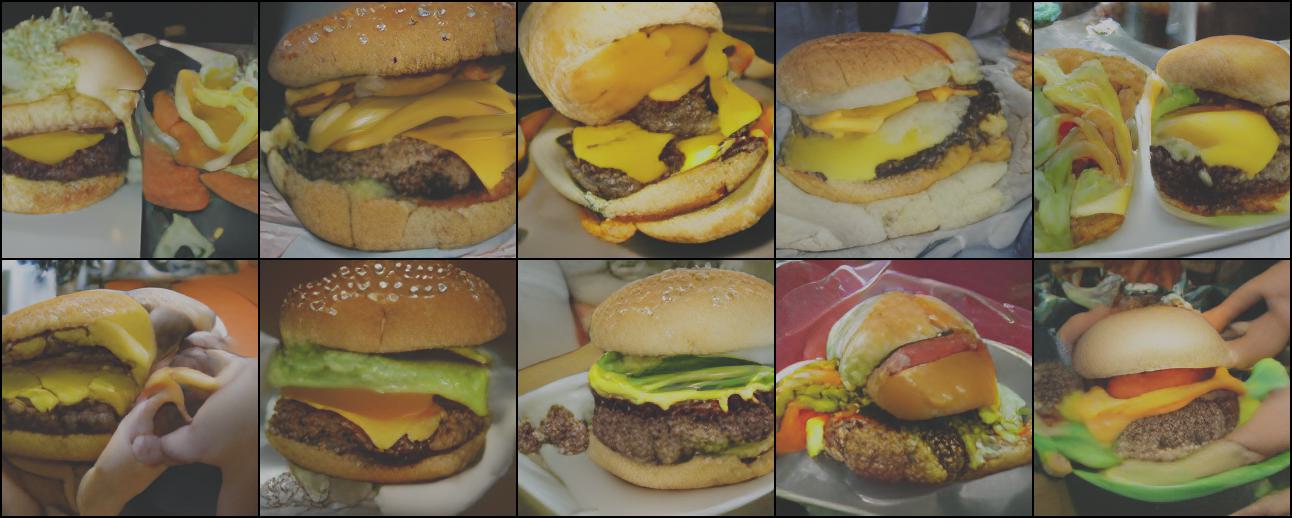}
    }
    \subfigure[\textbf{di4c-d}, $4$ steps]{
        \includegraphics[width=0.4\textwidth]{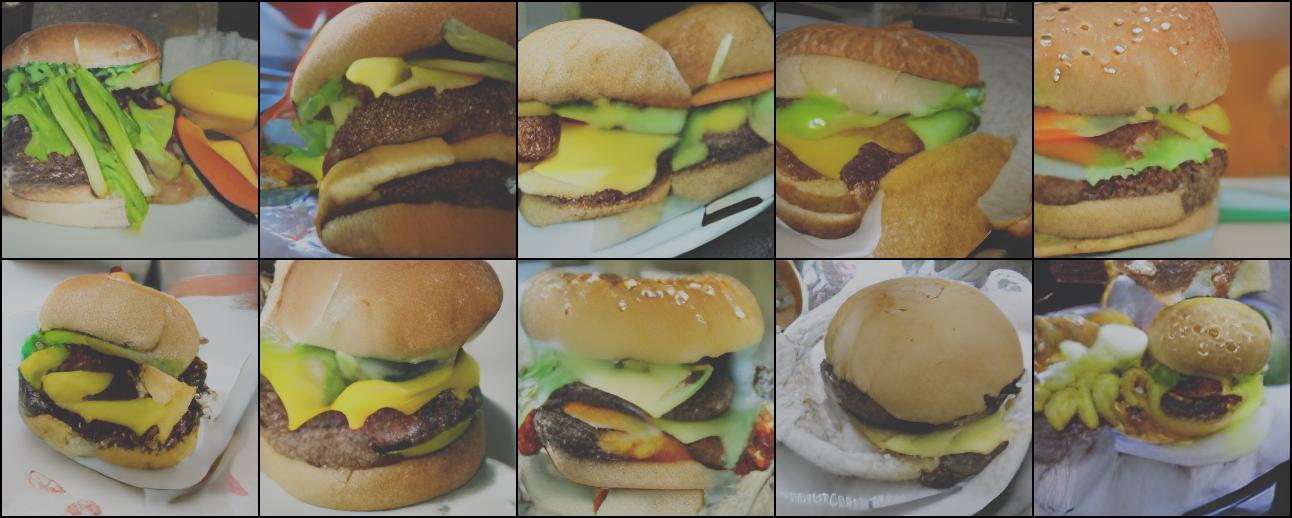}
    }
    \subfigure[\textbf{teacher}, $8$ steps]{
        \includegraphics[width=0.4\textwidth]{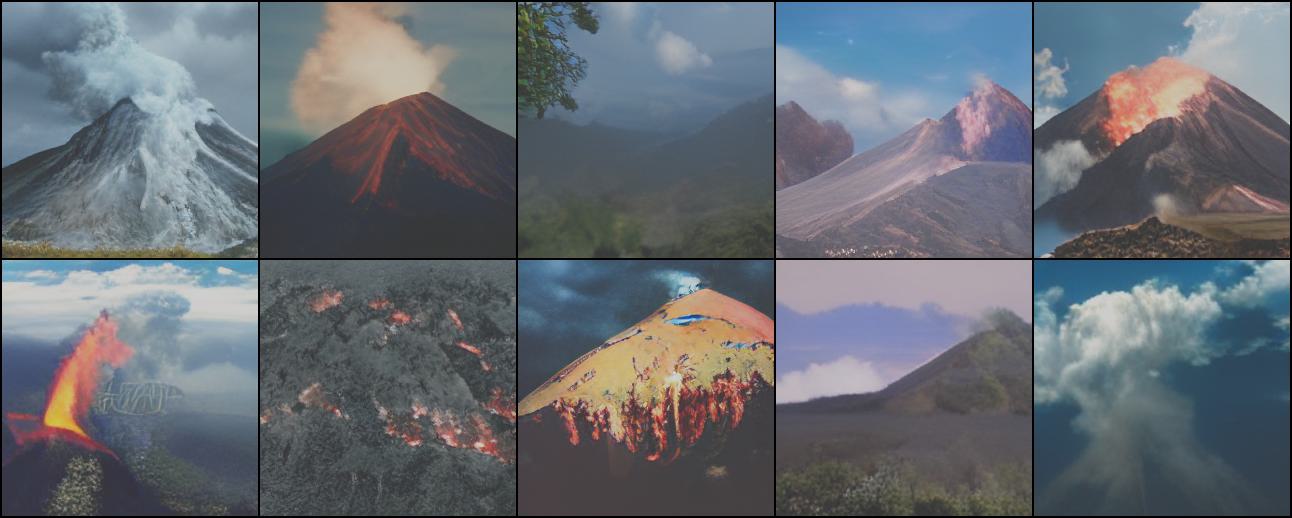}
    }
    \subfigure[\textbf{teacher}, $4$ steps]{
        \includegraphics[width=0.4\textwidth]{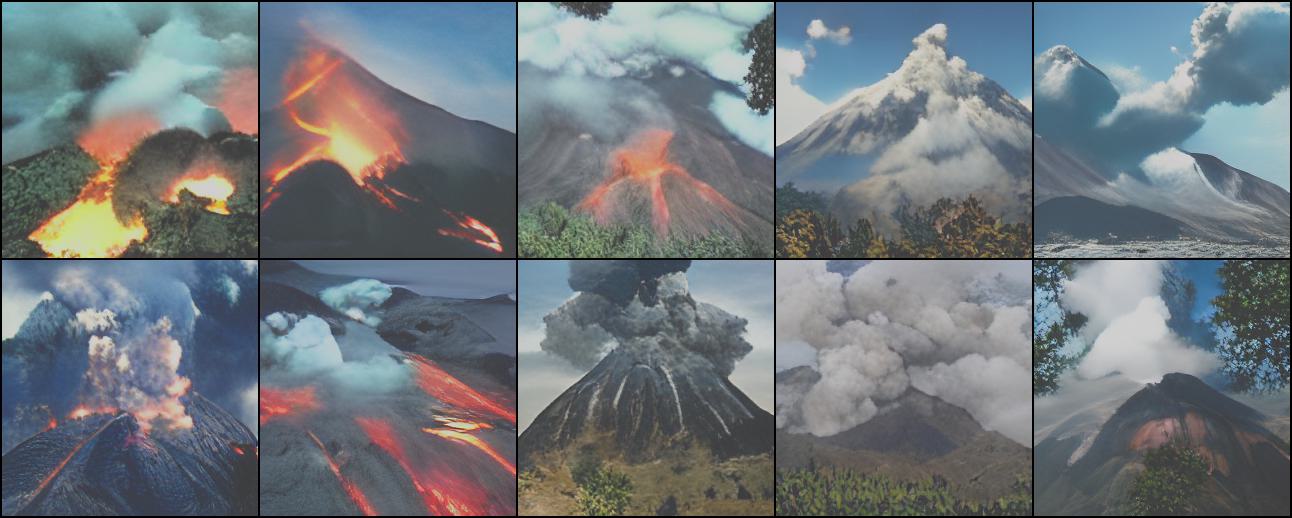}
    }
    \subfigure[\textbf{di4c}, $4$ steps]{
        \includegraphics[width=0.4\textwidth]{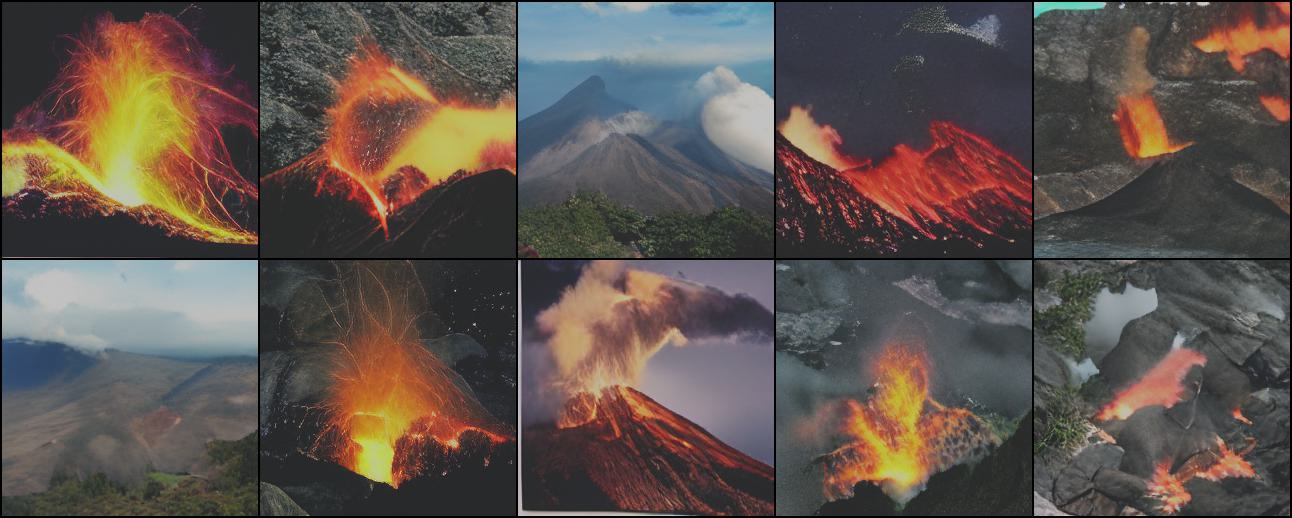}
    }
    \subfigure[\textbf{di4c-d}, $4$ steps]{
        \includegraphics[width=0.4\textwidth]{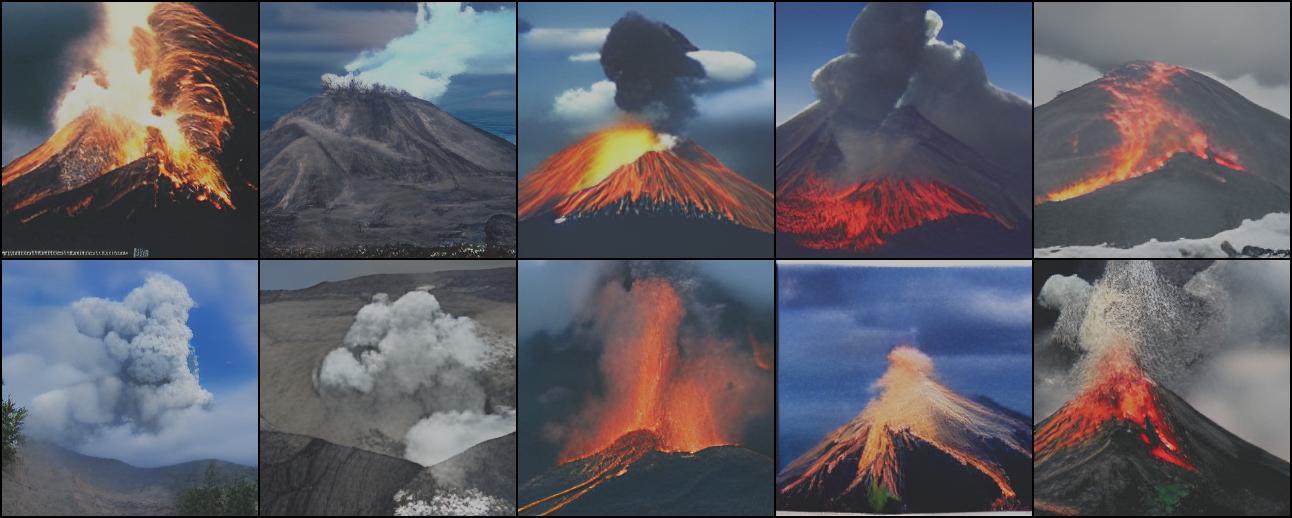}
    }
    \vspace{-2mm}
    \caption{Comparison of generated samples.
    (a)--(d): Conditioned with ImageNet label Ostrich (009).
    (e)--(h): Conditioned with Burger (933).
    (i)--(l): Conditioned with Volcano (980).}
    \label{fig:maskgit-samples}
\end{figure}

Samples generated by each model in Section~\ref{sec:ex-maskgit} are shown in
\Cref{fig:maskgit-samples}.
The best CFG coefficient in \Cref{fig:maskgit} in terms of FID was chosen for each model,
as shown in \Cref{tab:cfg-choice}.
The table also shows the FID/IS performance, together with the Precision/Recall values computed by the code of \citet{besnier2023pytorch}.
Regarding the choice of ImageNet labels (Ostrich, Burger, Volcano),
we followed \citet[][Figure~3]{besnier2023pytorch}.

\begin{table}[h]
    \centering
    \caption{Chosen CFG coefficients and corresponding
    FID/IS values.}
    \label{tab:cfg-choice}
    \vspace{2mm}
    \begin{tabular}{ccccccc}
        \toprule
               Model & \# steps & $w_\mathrm{cfg}$ & FID ($\downarrow$) & IS ($\uparrow$) & Precision ($\uparrow$) & Recall ($\uparrow$) \\
        \midrule
        \textbf{teacher} & 8 & 3.0 & 6.57 & 202.0 & 0.7939 & 0.5499 \\
        \textbf{teacher} & 4 & 11.0 & 7.97 & 216.0 & 0.7737 & 0.5057 \\
        \textbf{di4c} & 4 & 6.0 & 6.79 & 209.2 & 0.7910 & 0.5363 \\
        \textbf{di4c-d} & 4 & 7.0 & 6.57 & 213.6 & 0.7866 & 0.5391 \\
        \bottomrule
    \end{tabular}
\end{table}


\subsection{Masked diffusion language models}\label{sec:app-mdlm}
\subsubsection{Diffusion modeling and sampling}
\paragraph{Diffusion modeling.}
We basically worked under the same setting as in Section~\ref{sec:mg-training}
except for the following:
\begin{itemize}
    \item $D=1024$.
    \item The non-mask codebook $\S^*$ is given by the set of GPT-2 tokenizers,
    which satisfies $\lvert\S^*\rvert = 50257$.
    \item The masking probability is given by $m_t = t$.
\end{itemize}
While the above apparently linear noise scheduling is called {\it log-linear} \citep[Section~E.1]{sahoo2024simple},
the naming comes from their parametrization: $m_t = 1 - e^{-\sigma(t)}$ with $\sigma(t) = - \log(1-t)$.

\paragraph{Analytical sampling in masked diffusion.}
The sampling algorithm used for the MDLMs is essentially
the same as the analytical sampling in Section~\ref{sec:sampling},
tailored for masked diffusions.
What we have is a dimensionally independent denoiser
$p_{0|t}(\bm{x}_0|\bm{x}_t) = \prod_{d=1}^D p_{0|t}^d(x_0^d|\bm{x}_t)$.
By using \eqref{eq:useful-fact},
we can deduce the resulting {\it product} denoiser $p_{s|t}$ based on $p_{0|t}$ as follows:
\begin{align}
    p_{s|t}(\bm{x}_s|\bm{x}_t)
    &= \sum_{\bm{x}_0}q_{s|0,t}(\bm{x}_s|\bm{x}_0,\bm{x}_t)p_{0|t}(\bm{x}_0|\bm{x}_t) 
    = \sum_{\bm{x}_0}
    \frac{q_{s|0}(\bm{x}_s|\bm{x}_0)q_{t|s}(\bm{x}_t|\bm{x}_s)}{q_{t|0}(\bm{x}_t|\bm{x}_0)}p_{0|t}(\bm{x}_0|\bm{x}_t)\nonumber\\
    &= \sum_{\bm{x}_0}\prod_{d=1}^D
    \frac{q^d_{s|0}(x^d_s|x^d_0)q^d_{t|s}(x^d_t|x^d_s)}{q^d_{t|0}(x^d_t|x^d_0)}p^d_{0|t}(x^d_0|\bm{x}_t)\nonumber\\
    &= \prod_{d=1}^D \sum_{x^d_0}
    \frac{q^d_{s|0}(x^d_s|x^d_0)q^d_{t|s}(x^d_t|x^d_s)}{q^d_{t|0}(x^d_t|x^d_0)}p^d_{0|t}(x^d_0|\bm{x}_t)
    =: \prod_{d=1}^D p_{s|t}^d(x_s^d|\bm{x}_t).
    \label{eq:deduce-analytical}
\end{align}
Note that, once a token (dimension) is unmasked,
we do not need to further change that token in the backward process:
this property is incorporated in $p_{0|t}$
as $p_{0|t}^d(x_t^d|\bm{x}_t)=1$ for $x_t^d\ne\!\mask\!$ \citep[Section~3.2.3]{sahoo2024simple}.
Thus, we just need to consider the case $x_t^d=\!\mask\!$.
By using the fact that the masking probability is given by $m_t=t$
and $q_{t|s}^d(\!\mask\!|x) = \frac{m_t-m_s}{1-m_s}$ for $x\ne\!\mask\!$
in \eqref{eq:masking-t|s},
if $x_t^d=\!\mask\!$ and $x\ne\!\mask\!$,
we simply have
\[
    p_{s|t}^d(x|\bm{x}_t)
    = \frac{q^d_{s|0}(x|x)q^d_{t|s}(\!\mask\!|x)}{q^d_{t|0}(\!\mask\!|x)}p^d_{0|t}(x|\bm{x}_t)
    = \frac{(1-m_s)\frac{m_t-m_s}{1-m_s}}{m_t}p^d_{0|t}(x|\bm{x}_t)
    = \frac{m_t-m_s}{m_t}p^d_{0|t}(x|\bm{x}_t),
\]
which corresponds to \citet[Eq.~7]{sahoo2024simple}.

\subsubsection{Implementation and training}\label{sec:app-sdtt-training}
\paragraph{Network architecture.}
For the teacher model,
we used the two ``small" checkpoints (round $6$ and $7$) of \citet{deschenaux2024beyond}\footnote{
It is loaded by {\ttfamily load\_small\_student(loss={\textquotesingle}kld{\textquotesingle}, round=$n$)} with $n=6,7$,
from the library {\ttfamily sdtt} in \url{https://github.com/jdeschena/sdtt}.
},
which uses a transformer architecture to compute the logits of $p_{0|t}(\cdot|\bm{x}_t)$ for each token.
The transformer architecture is originally from \citep{sahoo2024simple},
and has 169M parameters with $12$ layers, the embedding dimension of $768$, and $12$ attention heads.
The model receives $D = 1024$ tokens and does not depend on the timestep.
For adaptation to mixture modeling, we just applied the same modification as given in Section~\ref{sec:mg-training}.

\paragraph{Training.}
While the model accepts a continuous time training,
we followed \citet{deschenaux2024beyond} to digitize the timesteps to
$\mathbb{T}=\{\Delta t\cdot n\mid n=0,1,2,3,\ldots, 1024\}$,
with $\Delta t = 1/1024$.
We used the following loss function:
\begin{equation}
    \bm{1}_{\{t\le \delta\}}\L_\mathrm{distil}(\theta;\psi, q_t, t)
    +
    \bm{1}_{\{t>\delta\}}\L_\mathrm{consis}(\theta;\psi, q_t, 0, t-\Delta t, t)
    +
    \alpha_t\L_\mathrm{corr}(\theta; t)+ \L_\mathrm{marginal}(\theta; \psi,q_t,t),
    \label{eq:loss-sdtt}
\end{equation}
where the details are as follows:
\begin{itemize}
    \item We set $\delta=0.02$.
    \item We sample $r\sim \mathrm{Unif}([0,1])$ and let
    \[
        t = 2^{-10}(1 + \lfloor2^{10}t^*(r)\rfloor),
        \qquad \text{where} \quad 
        t^*(r) = \begin{cases}
            \delta r & \text{with probability $0.3$}, \\
            \delta + (1-\delta)r & \text{with probability $0.7$}.
        \end{cases}
    \]
    Therefore, $t\in\mathbb{T}\setminus\{0\}$ almost surely.
    \item We sampled $\bm{x}_t$ according to \eqref{eq:mask-forward} using $\bm{x}_0$ from data (OpenWebText).
    \item We used $\alpha_t = 0$ in the experimental results shown in Section~\ref{sec:mdlm}.
        In Section~\ref{sec:mdlm-data-loss},
        we also report the result of setting $\alpha_t = 0.1\cdot g(t)$ with $g(t)$ from \eqref{eq:sigmoid-alpha}.
    \item Control variates were used in all the experiments.
\end{itemize}
For training, we mostly followed the original setting of \citet{deschenaux2024beyond}:
We used the Adam optimizer (but with a learning rate of $3\times10^{-5}$) with EMA (weight decay 0.9999)
and did a constant warm-up (increasing the learning rate linearly for the first 500 iterations and setting it constant after that).
For each experiment, the Di4C training (one round) was run for 100K iterations over 2x A6000 GPUs,
where the minibatch size was $2$ ($1$ for each device) and $\lambda$-batch size was $16$.

\paragraph{Self-BLEU computation.}
As described in Section~\ref{sec:mdlm},
in the conditional generation experiment,
we generated $M=5$ continuations $C^{(i)}=\{X_1^{(i)},\ldots, X_5^{(i)}\}$
conditioned on the first $50$ tokens (prompt) of each WebText datapoint $X^{(i)}$.
Each continuation was of $100$ tokens including the prompt,
and we used $M=256$ prompts from the WebText dataset in total.
To quantify the diversity of continuations, we followed \citet{deschenaux2024beyond}
and computed the Self-BLEU score as
\[
    \frac1N\sum_{i=1}^N \frac1M\sum_{j=1}^M \mathrm{BLEU}(X_j^{(i)}; C^{(i)}\setminus X_j^{(i)}),
\]
where $\mathrm{BLEU}(X; C)$ is the BLEU score of a sentence $X$ against the set of reference sentences $C$.
To actually compute this,
we tokenized the inputs with the GPT-2 tokenizer,
and then utilized the implementation of \citet{zhu2018texygen}\footnote{\url{https://github.com/geek-ai/Texygen/blob/master/utils/metrics/SelfBleu.py}.} with $\texttt{ngram}=4$,
which internally calls the {\ttfamily sentence\_bleu} function of the NLTK library \citep{bird2009natural}\footnote{\url{https://github.com/nltk/nltk/blob/3.7/nltk/translate/bleu_score.py}.}
with equal weighting and the {\ttfamily method1} smoothing function.

\paragraph{Evaluation.}
Except for the Self-BLEU computation,
all the evaluations were done using the code of SDTT~\citep{deschenaux2024beyond}\footnote{\url{https://github.com/jdeschena/sdtt/tree/main}.}.

\subsubsection{Additional experimental results}\label{sec:mdlm-data-loss}
\begin{figure*}[!t]
    \centering
    \subfigure[Gen.~PPL vs Num.~Steps in unconditional generation]{
        \includegraphics[width=0.45\textwidth]{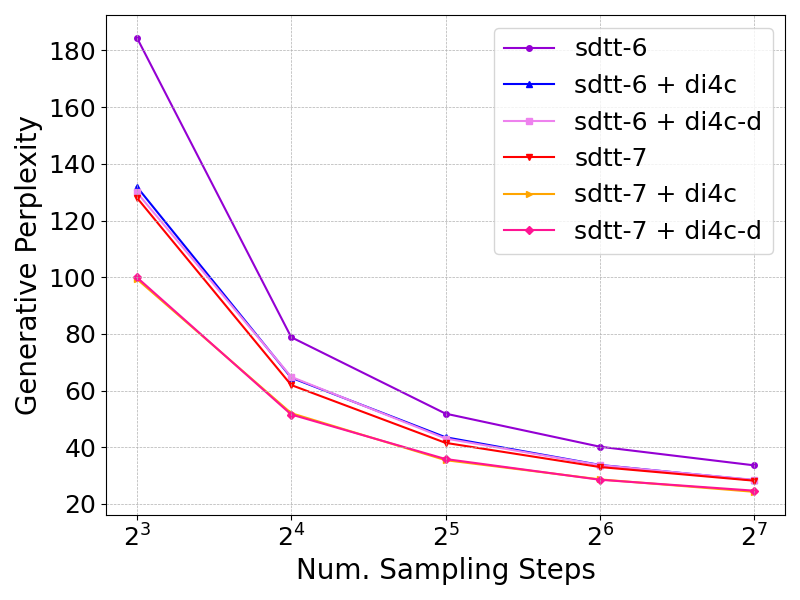}
        \label{fig:language-app-c}
    }
    \subfigure[Gen.~PPL vs Self-BLEU in conditional generation]{
        \includegraphics[width=0.45\textwidth]{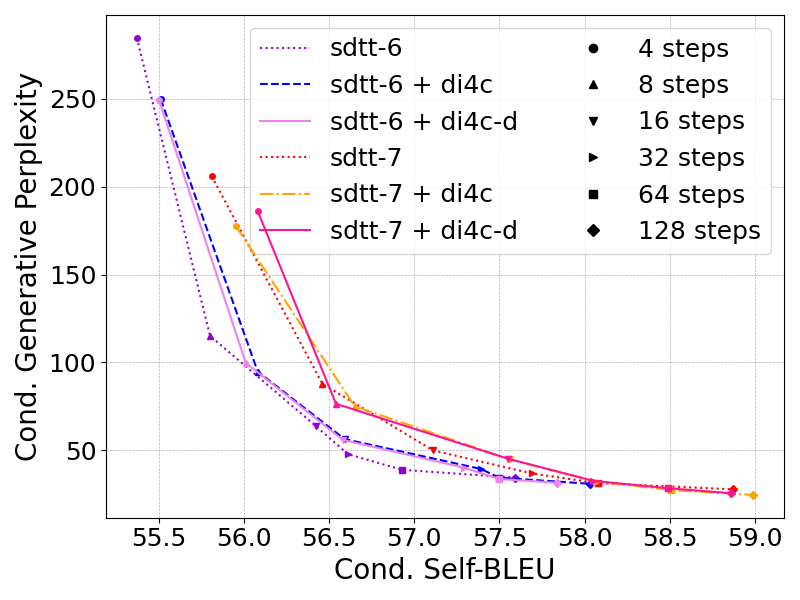}
        \label{fig:language-app-d}
    }
    \vspace{-2mm}
    \caption{Comparison of Di4C distillations of SDTT checkpoints with and without data loss.}
    \label{fig:language-app}
\end{figure*}

\paragraph{Data loss ablation.}
In Figure~\ref{fig:language-app},
we show the result for \textbf{sdtt-6/7 + di4c-d},
which used loss \eqref{eq:loss-sdtt} with $\alpha_t=0.1 g(t)$
as mentioned in Section~\ref{sec:app-sdtt-training}.
The results are almost the same as those without data loss
(especially in Figure~\ref{fig:language-app}, where \textbf{sdtt-7 + di4c} is hidden behind the curve of
\textbf{sdtt-7 + di4c-d}).
Since the ones without data loss showed slightly better generative perplexities,
we presented them as the main model in Section~\ref{sec:mdlm}.

\begin{figure*}[!t]
    \centering
    \subfigure[MAUVE results with additional Di4C iteration]{
        \includegraphics[width=0.45\textwidth]{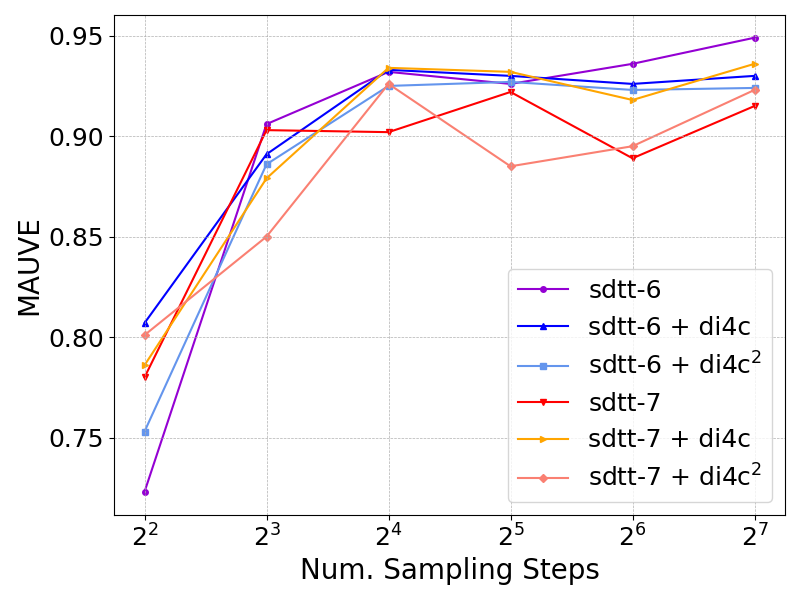}
        \label{fig:mauve-c}
    }
    \subfigure[MAUVE results with and without data loss]{
        \includegraphics[width=0.45\textwidth]{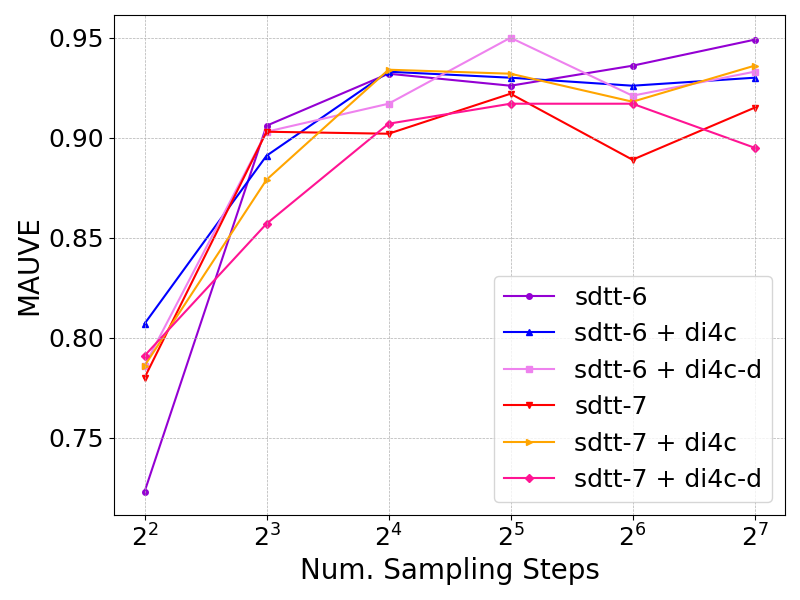}
        \label{fig:mauve-d}
    }
    \vspace{-2mm}
    \caption{Comparison of Di4C distillations of SDTT checkpoints with and without data loss.}
    \label{fig:mauve}
\end{figure*}

\paragraph{MAUVE results.}
We also tested our models with the MAUVE score~\citep{pillutla2021mauve}.
The setting is the same as the unconditional generation in Section~\ref{sec:mdlm},
and the MAUVE computation is done by using the code of SDTT~\citep{deschenaux2024beyond}.
As shown in Figure~\ref{fig:mauve}, no significant performance decay from the teacher model was observed.

\subsubsection{Generated samples}
Let us qualitatively compare our best model (\textbf{sdtt-7 + di4c$^{\boldsymbol{2}}$}) with
the SDTT checkpoint which our model is based on (\textbf{sdtt-7}),
conditioned on the first 50 tokens from this paper's abstract (old version; highlighted in {\color{blue}blue}):
\begin{itemize}
    \item (\textbf{sdtt-7}, 4 steps)
    {\color{blue}Diffusion models have demonstrated exceptional performances in various fields of generative modeling. While they often outperform competitors including VAEs and GANs in sample quality and diversity, they suffer from slow sampling speed due to their iterative nature. Recently, dist}ricting the image spl Image\texttt{\string\n\string\n}distriction the size spl Image size The stable image spl consists of thin--arr 2D-frame data that supports band-- construction data and generates the physical representation of image presentation. Although much is known
    \item (\textbf{sdtt-7 + di4c$^{\boldsymbol{2}}$}, 4 steps)
    {\color{blue}Diffusion models have demonstrated exceptional performances in various fields of generative modeling. While they often outperform competitors including VAEs and GANs in sample quality and diversity, they suffer from slow sampling speed due to their iterative nature. Recently, dist}illing the high spl Image Image\texttt{\string\n\string\n}distriction the whole spl Image Image generating an image Image is a thin-narr 2D-layer architecture that exhibits banding-resolving effects in the lowizing of a low resolution. Little is known
    \item (\textbf{sdtt-7}, 16 steps)
    {\color{blue}Diffusion models have demonstrated exceptional performances in various fields of generative modeling. While they often outperform competitors including VAEs and GANs in sample quality and diversity, they suffer from slow sampling speed due to their iterative nature. Recently, dist}illilusion models, which use a dataset from the sample and using inferred model data, have become standard. self-arrained three-block sampling that utilizes a subset of the data and combining Bayesianesian and inferred model data that combines Bayesian
    \item (\textbf{sdtt-7 + di4c$^{\boldsymbol{2}}$}, 16 steps)
    {\color{blue}Diffusion models have demonstrated exceptional performances in various fields of generative modeling. While they often outperform competitors including VAEs and GANs in sample quality and diversity, they suffer from slow sampling speed due to their iterative nature. Recently, dist}orting Diffusion models, nested within a conventional model architecture, and using different model architectures, has become a flexible self-arrative three-model architecture that supports intensive problem-solving and mature Bayesianesian model and model development. This approach has
    \item (\textbf{sdtt-7}, 64 steps)
    {\color{blue}Diffusion models have demonstrated exceptional performances in various fields of generative modeling. While they often outperform competitors including VAEs and GANs in sample quality and diversity, they suffer from slow sampling speed due to their iterative nature. Recently, dist}ancing from the Diffusion model has allowed developers to construct new models using rapid-processing, supervised learning supervised (GCl supervised) software-drawing that improves the ability to identify discriminant parameters, functional model depth, and is the process of rapidly
    \item (\textbf{sdtt-7 + di4c$^{\boldsymbol{2}}$}, 64 steps)
    {\color{blue}Diffusion models have demonstrated exceptional performances in various fields of generative modeling. While they often outperform competitors including VAEs and GANs in sample quality and diversity, they suffer from slow sampling speed due to their iterative nature. Recently, dist}ancing from the Diffusion model has allowed scientists to analyze the model using discriml, a software that utilizes to discriml. This software can generate images that offset the time to be discrimlative, and reduce the time to be discrimlative.
    \item (\textbf{sdtt-7}, 256 steps)
    {\color{blue}Diffusion models have demonstrated exceptional performances in various fields of generative modeling. While they often outperform competitors including VAEs and GANs in sample quality and diversity, they suffer from slow sampling speed due to their iterative nature. Recently, dist}ancing from the traditional resource management process has led to a change in the type of user experience, producing significant advances in the types of software-developing operations, while facilitating the adoption of parallel programming and functional programming approaches. However,, it is increasingly
    \item (\textbf{sdtt-7 + di4c$^{\boldsymbol{2}}$}, 256 steps)
    {\color{blue}Diffusion models have demonstrated exceptional performances in various fields of generative modeling. While they often outperform competitors including VAEs and GANs in sample quality and diversity, they suffer from slow sampling speed due to their iterative nature. Recently, dist}orting cloud images are in the process of processing to high standards in both sample quality and storage, producing important advances in the development of software-developing operations, while facilitating the development of parallel programming and functional efficiency. GPUs are in the process of processing
\end{itemize}

\end{document}